\documentclass[10pt,journal,compsoc]{IEEEtran}
\usepackage{graphicx,comment}
\usepackage{amsmath,amsfonts,times,epsfig,color,bm}
\usepackage{soul,xargs}

\usepackage{wrapfig}
\usepackage{varwidth}
\usepackage{latexsym}
\usepackage{mathtools,nccmath}
\usepackage{caption}
\usepackage{subcaption}
\usepackage{algorithm,algorithmicx,setspace}
\usepackage{algpseudocode}
\usepackage{bm}
\usepackage{fancyhdr}
\usepackage{blindtext}
\ifCLASSOPTIONcompsoc
  \usepackage[nocompress]{cite}
\else
  \usepackage{cite}
\fi

\ifCLASSINFOpdf
\else
\fi
\def\bbtheta{{\mbox{\boldmath $\theta$}}}
\def\bbTheta{{\mbox{\boldmath $\Theta$}}}
\def\bbepsilon{{\mbox{\boldmath $\epsilon$}}}
\def\bbSig{{\mbox{\boldmath $\Sigma$}}}
\def\bbphi{{\mbox{\boldmath $\phi$}}}
\def\bbPhi{{\mbox{\boldmath $\Phi$}}}

\def \nrf {n_{\text{RF}} }
\def \noisevar {\sigma_{n}^2}
\def \esym {m}
\def \Esym {M}

\def \myspa {}
\def \myspace {}

\newcommand{\mb}[1]{\boldsymbol{\mathbf{#1}}}
\newcommand{\mc}[1]{{\mathcal{#1}}}
\newcommandx{\thiswillnotshow}[2][1=]{\todo[disable,#1]{#2}}

\newcommand{\ysub}[1]{#1:}

\algdef{SE}[SUBALG]{Indent}{EndIndent}{}{\algorithmicend\ }%
\algtext*{Indent}
\algtext*{EndIndent}

\renewcommand{\thesection}{\arabic{section}}

\begin{document}
\title{Incremental Ensemble Gaussian Processes }
%
%
%
%

\author{Qin~Lu,~\IEEEmembership{Member,~IEEE,}
        Georgios V. ~Karanikolas,~\IEEEmembership{Student Member,~IEEE,}
        and~Georgios~B.~Giannakis,~\IEEEmembership{Fellow,~IEEE}
\IEEEcompsocitemizethanks{\IEEEcompsocthanksitem The authors are with Dept. of Electrical and Computer Engineering and Digital Technology Center,
	University of Minnesota
	Minneapolis, MN 55455. 
E-mails: qlu@umn.edu; karan029@umn.edu; georgios@umn.edu
\IEEEcompsocthanksitem The first two authors are equally contributed.
}
}


\IEEEtitleabstractindextext{%
\begin{abstract}
Belonging to the family of Bayesian nonparametrics, Gaussian process (GP) based approaches have well-documented merits not only in learning over a rich class of nonlinear functions, but also in quantifying the associated uncertainty. However, most GP methods rely on a single preselected kernel function, which may fall short in characterizing data samples that arrive sequentially in time-critical applications. To enable {\it online} kernel adaptation, the present work advocates an incremental ensemble (IE-) GP framework, where an EGP meta-learner employs an {\it ensemble} of GP learners, each having a unique kernel belonging to a prescribed kernel dictionary. With each GP expert leveraging the random feature-based approximation to perform online prediction and model update with {\it scalability}, the EGP meta-learner capitalizes on data-adaptive weights to synthesize the per-expert predictions. 
Further, the novel IE-GP is generalized to accommodate time-varying functions by modeling structured dynamics at the EGP meta-learner and within each GP learner. To benchmark the performance of IE-GP and its dynamic variant in the adversarial setting where the modeling assumptions are violated, rigorous performance analysis has been conducted via the notion of regret, as the norm in online convex optimization. Last but not the least, online unsupervised learning for dimensionality reduction is explored under the novel IE-GP framework. Synthetic and real data tests demonstrate the effectiveness of the proposed schemes.
\end{abstract}

\begin{IEEEkeywords}
Gaussian processes,  ensemble learning, online prediction, random features, regret analysis
\end{IEEEkeywords}}

\maketitle

\IEEEdisplaynontitleabstractindextext

\IEEEpeerreviewmaketitle

\IEEEraisesectionheading{\section{Introduction}\label{sec:introduction}}
\IEEEPARstart{G}{aussian} processes (GPs) cross-fertilize merits of kernel methods and Bayesian models to benefit several learning tasks, including regression, classification,  ranking, and dimensionality reduction \cite{williams2006gaussian}. In GP-based approaches, a Gaussian \emph{prior} is assumed over a learning function $f(\cdot)$ with covariance (kernel) capturing similarities among $\{f(\mathbf{x}_t)\}$ dependent on inputs $\{\mathbf{x}_t\}$. Given observed outputs  $\{y_t\}$ linked to the latent function $f(\cdot)$ via the conditionally independent per-datum likelihood $p(y_t|f(\mathbf{x}_t))$, Bayes rule produces the \emph{posterior} distribution of $f(\cdot)$, based on which task-specific inference can be effected on the unseen data. Besides learning functions with rich expressiveness, the Bayesian framework of GP-based approaches further quantifies uncertainty of the function estimate, which is of utmost importance in safety-critical applications. For instance in medical diagnosis~\cite{kononenko2001machine}, human intervention would be called for when machine operated decisions are accompanied by high uncertainty.

In spite of the intriguing performance, applicability of plain-vanilla GPs in the big data regime is discouraged by the cubic computational complexity in the number of training samples \cite{williams2006gaussian}. To relieve the scalability issue, various attempts have been made, including efficient numerical operation~\cite{cutajar2016preconditioning, wang2019exact}, and structured approximants of the kernel matrix~\cite{snelson2006sparse, titsias2009variational}. Of special interest to us is the random feature (RF) based approach, which, leveraging the spectral properties of stationary kernels, converts the nonparametric GP paradigm to a parametric one \cite{quia2010sparse, rahimi2008random}.
Such a parametric approach readily accommodates online processing of data samples \cite{gijsberts2013real},  which is necessitated in time-critical applications. For instance, the detection of spam emails is performed on an email-by-email basis in real time. Albeit accommodating online operation, the performance of existing RF-based GP methods hinges on the single {\it preselected} kernel, that may fall short in characterizing upcoming data. Henceforth, online kernel adaptation is essential to real-time decision making.

On the theoretical horizon, to benchmark performance of online approaches, analysis is usually conducted via the notion of regret, the norm in online convex optimization~\cite{hazan2016introduction} and online learning with experts~\cite{cesa2006prediction}, to combat with the adversarial setting where the generative assumptions are violated. Although several scalable GP approaches have been developed for the online operation~\cite{cheng2016incremental, bui2017streaming}, regret analysis has not been touch upon except for the plain-vanilla GP~\cite{kakade2006worst}.

In accordance with the aforementioned desiderata, the goal of the current work is to pursue algorithmic developments of scalable GPs that could enable kernel adaptation to cope with function dynamics in the online scenario, as well as benchmark performance of the resultant approaches via regret-based analysis.

\subsection{Related works}
To contextualize the current contribution, the following existing works will be outlined.
\vspace{0.1cm}

\noindent {\bf (Online) Scalable GPs}. 
Approaches to effect scalability in GPs rely on advanced numerical methods~\cite{cutajar2016preconditioning, wang2019exact}, special kernel functions~\cite{gilboa2013scaling, cunningham2008fast}, or low rank approximants of the kernel matrix~\cite{snelson2006sparse, quinonero2005unifying, titsias2009variational, quia2010sparse}. A well-known low-rank scheme summarizes the $T$ training samples via $q (\ll T)$ pseudo data with inducing inputs that are employed for inference in the testing 
phase~\cite{snelson2006sparse, quinonero2005unifying, titsias2009variational}. 
This global summary amounts to approximating the original GP prior with a kernel matrix having low rank $q$, thus reducing the complexity of batch computations  to $\mathcal{O}(Tq^2)$.
Rather than the spatial sampling,  another less explored low-rank approach  leverages spectral components of shift-invariant kernels to yield the random feature (RF) based kernel approximation~\cite{rahimi2008random}. Converting the nonparametric GP prior to a parametric one, the resultant RF-based GP approaches can afford complexity comparable to the inducing points-based 
approximants~\cite{quia2010sparse,gal2015improving}. To accommodate time-critical applications, online scalable GP approaches have been developed relying on stochastic optimization or online variational inference; see, e.g., \cite{cheng2016incremental, bui2017streaming, hoang2015unifying}. Albeit ensuring scalability, these approaches rely on a {\it single} GP kernel, which may limit expressiveness of the sought function. Also, theoretical analyses that quantify the robustness of the online solvers to the adversarial setting are largely unexplored.

\vspace{0.1cm}
\noindent {\bf Expert-based GPs}. 
An {\it ensemble} of ({\it local} or distributed) GP experts, each relying on a unique kernel to summarize a subset of the training samples, has been leveraged to lower computational complexity, or/and account for nonstationarity of the learning function. Depending on how data samples are distributed and how predictions over experts are aggregated, well-known examples include the naive-local-experts~\cite{kim2005analyzing}, product-of-experts~\cite{tresp2000bayesian, deisenroth2015distributed}, mixture-of-experts~\cite{rasmussen2002infinite, meeds2006alternative}, and most recently sum-product networks~\cite{trapp2020deep} based approaches. In spite of the advantages enjoyed in different lines of work, existing expert-based GP approaches operate in \emph{batch} mode, thus falling short in dealing with time-critical applications that welcome online decision-making. 

\vspace{0.1cm}
\noindent {\bf (Online) Multi-kernel learning}. Parallel to the probabilistic GP paradigm, kernel-based learning has been pursued also in the \emph{deterministic} reproducing kernel Hilbert space (RKHS).
Faced with inscalability arising from abudance of training data, kernel-based approaches also resort to low-rank approximants of the kernel matrix, including the RF-based approximation~\cite{rahimi2008random}. Bypassing kernel selection via cross validation, data-driven multi-kernel learning enjoys well-documented performances; see, e.g., \cite{micchelli2005learning,alvarez2012kernels}. To further accommodating online operation, scalable kernel-based learning has been investigated for a single ~\cite{lu2016large} as well as for an ensemble of learners~\cite{jin2010online,shen2019random}. Most recently, online RF-based  approaches based on an ensemble of RKHS learners have been reported along with their regret-based performance for static and dynamic settings~\cite{shen2019random}. 

\vspace{0.1cm}
\noindent {\bf GP latent variable model (LVM).} Leveraging GPs to model the mapping from the hidden low-dimensional input space to high-dimensional observations, GPLVMs are established probabilistic approaches to {\it nonlinear} dimensionality reduction~\cite{Lawrence05}. Scalable GPLVMs have been devised by relying on
inducing points-based approximations in the batch setting~\cite{Lawrence07,damianou2016variational}, as well as the variational and online variants~\cite{damianou2016variational, yao2011learning,yali2014bayesian,qin2019scalable}. However, to the best of our knowledge, RF-based counterparts, in spite of the application in kernel principle component analysis (PCA) (see, e.g.~\cite{ghashami2016streaming}), have not been touched upon in the realm of GPLVMs. Regarding ensemble learning, a GPLVM scheme which can be broadly categorized in this area is~\cite{yali2014bayesian}, where different from the proposed approach the goal is to track the latent state of a dynamical system. Ensemble methods are, nonetheless,  more commonplace in the context of probabilistic PCA for \emph{linear} dimensionality reduction; see~\cite{tipping1999mixtures} for the seminal work and~\cite{bellas2013model} for an online variant.

\subsection{ Contributions}
Relative to the aforementioned past works, the present paper aims at bringing together the fields of scalable GPs and online learning with expert advice \cite{cesa2006prediction}. 
The pursuit lies in algorithmic development, as well as performance analysis via the measure of regret to account for the violations of the generative models. The detailed contributions are highlighted as follows.
\begin{itemize}
	\item[\textbf{c1)}] 
	Towards online kernel adaptation, the present work advocates an incremental (I) approach based on a weighted ensemble (E) of GP learners with scalable RF-based kernel approximations. The novel IE-GP learns the unknown function and jointly adapts to the appropriate EGP kernel on-the-fly. 
	\item[\textbf{c2)}] To cope with learning nonstationary functions, dynamic IE-GP variants have been devised to capture structured dynamics at the EGP meta-learner and individual GP learners via a hidden Markov model and the state-space models, respectively.
	\item[\textbf{c3)}] To account for data being adversarially chosen in the online setting, the performances of IE-GP and its dynamic variant are compared with some benchmark functions with data in hindsight via static and switching regret analyses. In both cases, the cumulative regrets over $T$ slots are of order $\mathcal{O} (\log T)$, implying no regret on average.
	\item[\textbf{c4)}] Complementary to the supervised function learning task, online unsupervised learning for dimensionality reduction (a.k.a. latent variable model) is investigated under the proposed IE-GP paradigm. 
	\item[\textbf{c5)}] Extensive experimental results are provided to validate the merits of the proposed methods in regression, classification and dimensionality reduction tasks.
\end{itemize}
Relative to the conference precursor \cite{lu2020ensemble}, the novelty lies in the following four aspects: 1) A switching (S) IE-GP approach is devised by modeling dynamics at the EGP meta-learner via a first-order Markov chain;  2) The performance of the proposed SIE-GP is analysed via the notion of switching regret; 3) EGP-based  online unsupervised learning with RFs for scalability is further explored; 4) Experimental section has been significantly expanded via the inclusion of classification and dimensionality reduction tests.\\

\noindent {\bf Notation}. Scalars are denoted by lowercase, column vectors by bold lowercase, and matrices  by bold uppercase fonts. Superscripts $~^\top$ and $~^{-1}$  denote transpose, and matrix inverse, respectively; while $\mathbf{0}_N$ stands for the $N\times1$ all-zero vector; and 
$\mathcal{N}(\mathbf{x}; \boldsymbol{\mu}, \mathbf{K})$ for the probability density function (pdf) of a Gaussian random vector $\mathbf{x}$ with mean $\boldsymbol{\mu}$, and covariance matrix $\bf K$. Subscript ``${t+1|\mathbf{t}}$" signifies that prediction for slot $t+1$ relies on the \emph{batch} of samples up to and including $t$, while ``${t+1|t}$" stands for a single-step predictor. $I(x)$ represents the indicator function, which is $1$ if $x$ is true, and $0$ otherwise.

\section{Preliminaries and background} 
As a prelude to our online EGP approach that will also introduce context and notation, this section deals with batch and scalable learning based on a single GP.  

\subsection{Non-scalable batch GP-based learning}
Given data $\{\mathbf{x}_\tau,y_\tau\}$, the goal is to learn a function $f(\cdot)$ that links the $d\times 1$ input $\mathbf{x}_\tau$ with the scalar output $y_\tau$ as $\mathbf{x}_\tau \rightarrow f(\mathbf{x}_\tau) \rightarrow y_\tau$. Postulating $f$ with a GP prior as $f\sim \mathcal{GP}(0, \kappa(\mathbf{x},\mathbf{x}'))$, where $\kappa(\cdot,\cdot)$ is a kernel function measuring pairwise similarity of any two inputs,  the joint prior pdf of function evaluations $\mathbf{f}_t := [f(\mathbf{x}_1),\ldots, f(\mathbf{x}_t)]^\top$ at any inputs $\mathbf{X}_t := \left[\mathbf{x}_1, \ldots, \mathbf{x}_t\right]^\top$ is Gaussian distributed as \cite{williams2006gaussian}
\begin{equation}
p(\mathbf{f}_t| \mathbf{X}_t) = \mathcal{N} (\mathbf{f}_t ; {\bf 0}_t, {\bf K}_t )\  \  \ \forall t
\label{eq:gp_prior}
\end{equation}\vspace*{-0.5cm}\\
where ${\bf K}_t$ is a $t\times t$ covariance matrix with $(\tau,\tau')$th entry
$[{\bf K}_t]_{\tau,\tau'} = {\rm cov} (f(\mathbf{x}_\tau), f(\mathbf{x}_{\tau'})):=\kappa(\mathbf{x}_\tau, \mathbf{x}_{\tau'})$.

To estimate $f$, we rely on the observed outputs $\mathbf{y}_t := [y_1, \ldots, y_t]^\top$ that are linked with $\mathbf{f}_t$ via the conditional likelihood $p (\mathbf{y}_t| \mathbf{f}_t, \mathbf{X}_t)  = \prod_{\tau = 1}^{t} p(y_{\tau}| f(\mathbf{x}_{\tau}))$ that is assumed known. Through Bayes' rule, the latter will yield the posterior $p( \mathbf{f}_{t}| \mathbf{y}_t, \mathbf{X}_{t}) \propto p( \mathbf{f}_{t}| \mathbf{X}_{t}) p (\mathbf{y}_t| \mathbf{f}_t, \mathbf{X}_t)$. 
For Gaussian process regression (GPR) the conditional likelihood is assumed normal with mean $\mathbf{f}_t$ and covariance matrix $\noisevar {\bf I}_t$, that is, $p (\mathbf{y}_t |\mathbf{f}_t, \mathbf{X}_t) = \mathcal{N}(\mathbf{y}_t; \mathbf{f}_t, \noisevar {\bf I}_t)$, which along with the GP prior in \eqref{eq:gp_prior} yields the Gaussian posterior $p( \mathbf{f}_{t}| \mathbf{y}_t, \mathbf{X}_{t})$. For non-Gaussian likelihoods, sampling or approximate inference techniques will be called for to carry out the analytically intractable posterior $p( \mathbf{f}_{t}| \mathbf{y}_t, \mathbf{X}_{t})$~\cite{williams2006gaussian}. \\

\noindent
{\bf Prediction with a single GP}. Given training data $\{\mathbf{X}_{t}, \mathbf{y}_t\}$ and a new test input $\mathbf{x}_{t+1}$, we have from \eqref{eq:gp_prior} that $p(f(\mathbf{x}_{t+1})| \mathbf{f}_{t}, \mathbf{X}_{t})$ is Gaussian with known mean and covariance. Together with the known posterior $p( \mathbf{f}_{t}| \mathbf{y}_t, \mathbf{X}_{t})$, the so-termed predictive pdf of $f(\mathbf{x}_{t+1})$ can be obtained as~\cite{williams2006gaussian}
\vspace*{-0.2cm}
\begin{align}
	\hspace*{-0.25cm} p(f(\mathbf{x}_{t+1})| \mathbf{y}_t, \mathbf{X}_{t}) \!= \!\!\int\!\! p(f(\mathbf{x}_{t+1})| \mathbf{f}_{t}, \mathbf{X}_{t}) p( \mathbf{f}_{t}| \mathbf{y}_t, \mathbf{X}_{t}) d  \mathbf{f}_{t} \label{eq:p_f_pre}
\end{align}\vspace*{-0.4cm}\\
which is generally non-Gaussian if $p( \mathbf{f}_{t}| \mathbf{y}_t, \mathbf{X}_{t})$ is non-Gaussian, 
and thus necessitates Monte Carlo (MC) sampling to estimate it. Alternatively, $p( \mathbf{f}_{t}| \mathbf{y}_t, \mathbf{X}_{t})$ can be approximated by a Gaussian, yielding a Gaussian approximation for \eqref{eq:p_f_pre} as well.
Of course, $p(f(\mathbf{x}_{t+1})| \mathbf{y}_t, \mathbf{X}_{t})$ is Gaussian for GPR, with its mean and covariance matrix available in closed form. 

Using the pdf in \eqref{eq:p_f_pre} and the known $p(y_{t+1}|f(\mathbf{x}_{t+1}))$, it is also possible to find the predictive pdf of $y_{t+1}$ as 
\begin{align}
	&p(y_{t+1}| \mathbf{y}_t, \mathbf{X}_{t+1})  = \label{pdf4pred}\\
&\qquad\quad	\int\!\! p(y_{t+1}|f(\mathbf{x}_{t+1})) p(f(\mathbf{x}_{t+1})| \mathbf{y}_t, \mathbf{X}_{t}) df(\mathbf{x}_{t+1}) \nonumber
\end{align}
which generally requires MC sampling or $p(f_{t+1}|\mathbf{y}_t;$ $\mathbf{X}_{t+1})$ to be 
(at least approximately) Gaussian. Either way, \eqref{pdf4pred} yields the data predictive pdf
that fully quantifies the uncertainty of $y_{t+1}$. 

Specifically for GPR, we have 
\begin{align}
	p(y_{t+1}|\mathbf{y}_t,\mathbf{X}_{t+1}) = \mathcal{N}(y_{t+1}; \hat{y}_{t+1|\mathbf{t}}, \sigma_{t+1|\mathbf{t}}^2)
\end{align} \vspace*{-0.4cm}\\
where the mean and the variance of the predictor are given by ~\cite{williams2006gaussian}\vspace*{-0.1cm}
\begin{subequations}
	\begin{align}	
		\hat{y}_{t+1|\mathbf{t}} & = \mathbf{k}_{t+1}^{\top} (\mathbf{K}_t + \noisevar
		\mathbf{I}_t)^{-1} \mathbf{y}_t \\
		\sigma_{t+1|\mathbf{t}}^2& = \!\kappa(\mathbf{x}_{t+1},\mathbf{x}_{t+1}\!)\! -\! \mathbf{k}_{t+1}^{\top}\! (\mathbf{K}_t\! +\! \noisevar\mathbf{I}_t)^{-1} \mathbf{k}_{t+1}\! +\! \noisevar
	\end{align}\label{eq:plain_gpp}
\end{subequations}\vspace*{-0.4cm}\\
with $\mathbf{k}_{t+1} := [\kappa(\mathbf{x}_1, \mathbf{x}_{t+1}), \ldots, \kappa(\mathbf{x}_t,  \mathbf{x}_{t+1})]^\top$.
Clearly, this GP predictor is not scalable, since the complexity $\mathcal{O} (t^3)$ for inverting the $t\times t$ matrix in (5) will become prohibitively high as $t$ grows. 

\subsection{Scalable RF learning with a single GP}
Various attempts have been made to effect scalability in GP-based learning; see, e.g., \cite{quinonero2005unifying, titsias2009variational, quia2010sparse}. Most existing approaches amount to summarizing the training data via a much smaller number of pseudo data with inducing inputs, thereby obtaining a low-rank approximant of $\mathbf{K}_t$~\cite{quinonero2005unifying}. However, finding the locations of these inducing inputs entails involved optimization procedure. Targeting a low-rank approximant that bypasses such intricate training practice, we rely here on a standardized shift-invariant $\bar{\kappa}(\mathbf{x}, \mathbf{x}') = \bar{\kappa}(\mathbf{x}-\mathbf{x}') $, whose inverse Fourier transform is  \vspace*{-0.1cm}
\begin{align}
\hspace*{-0.13cm}
 \bar{\kappa}(\mathbf{x}-\mathbf{x}') = \int \pi_{\bar{\kappa}} (\mathbf{v}) e^{j\mathbf{v}^\top (\mathbf{x} - \mathbf{x}')} d\mathbf{v}:= \mathbb{E}_{\pi_{\bar{\kappa}}} \left[e^{j\mathbf{v}^\top (\mathbf{x} - \mathbf{x}')} \right] \label{eq:kernel_psd}
\end{align}\vspace*{-0.5cm}\\
where $\pi_{\bar{\kappa}}$ is the power spectral density (PSD), and the last equality follows after normalizing so that 
$\pi_{\bar{\kappa}}(\mathbf{v})$ integrates to $1$, thus allowing one to view it as a pdf.   

Since $\bar{\kappa}$ is real, the expectation in \eqref{eq:kernel_psd} is given by $\mathbb{E}_{\pi_{\bar{\kappa}}} \left[\cos(\mathbf{v}^\top (\mathbf{x} - \mathbf{x}'))\right]$, which,
upon drawing a sufficient number, say $\nrf$, of  independent and identically distributed (i.i.d.) samples $\{\mathbf{v}_j \}_{j = 1}^{\nrf}$ from $\pi_{\bar{\kappa}} (\mathbf{v})$, can be approximated by \footnote{Quantities with $\check{}$ involve RF approximations.}\vspace*{-0.15cm}
\begin{align}
	\check{\bar{\kappa}} (\mathbf{x}, \mathbf{x}'):= \frac{1}{\nrf} \sum_{j = 1}^{\nrf} \cos  \left(\mathbf{v}_j^\top (\mathbf{x} - \mathbf{x}') \right) \;. \label{kern_est}
\end{align}
Define the $2\nrf\times 1$ random feature (RF) vector as~\cite{quia2010sparse}
\vspace*{-0.15cm} 
\begin{align}
	&\bbphi_{\mathbf{v}} (\mathbf{x})  \label{eq:phi_x}\\
	&:=\! \frac{1}{\sqrt{\nrf}}\!\left[\sin(\mathbf{v}_1^\top \mathbf{x}), \cos(\mathbf{v}_1^\top \mathbf{x}), \ldots, \sin(\mathbf{v}_{\nrf}^\top \mathbf{x}), \cos(\mathbf{v}_{\nrf}^\top \mathbf{x})\right]^{\top} \nonumber
\end{align}
which allows us to rewrite $\check{\bar{\kappa}}$ in \eqref{kern_est} with
$\check{\bar{\kappa}}(\mathbf{x}, \mathbf{x}') = \bbphi_{\mathbf{v}}^{\top} (\mathbf{x})\bbphi_{\mathbf{v}}(\mathbf{x}')
$; and thus, the parametric approximant  
\begin{align}
	{\check f} (\mathbf{x}) =  \bbphi_{\mathbf{v}}^\top (\mathbf{x}) \bbtheta, \quad \bbtheta\sim \mathcal{N}(\bbtheta; \mathbf{0}_{2\nrf}, \sigma_\theta^2\mathbf{I}_{2\nrf}) \label{eq:f_check}
\end{align}
can be viewed as coming from a realization of the Gaussian $\bbtheta$ combined with $\bbphi_{\mathbf{v}}$ to yield the GP prior in \eqref{eq:gp_prior} with $\kappa = \sigma_\theta^2\bar{\kappa}$, where $\sigma_\theta^2$ is the magnitude of $\kappa$. 
Clearly, for any $\mathbf{X}_t$, the prior pdf of $\check{\mathbf f}_t$ is then
\begin{align}
	p(\check{\mathbf f}_t| \mathbf{X}_t) = \mathcal{N}(\check{\mathbf f}_t; \mathbf{0}_t, \check{\mathbf{K}}_t), \quad \check{\mathbf{K}}_t = \sigma_\theta^2\bbPhi_t \bbPhi_t^{\top} \label{eq:approx_GPP}
\end{align}
where $\bbPhi_t:=\left[\bbphi_{\mathbf{v}}(\mathbf{x}_1), \ldots, \bbphi_{\mathbf{v}}(\mathbf{x}_t) \right]^\top$, and $\check{\mathbf{K}}_t$ is then a low rank ($2\nrf$) approximant of $\mathbf{K}_t$ in \eqref{eq:gp_prior} for $t>2\nrf$.

With the parametric form of ${\check f} (\mathbf{x})$ in \eqref{eq:f_check}, the likelihood $p(\mathbf{y}_t|\check{ \mathbf{f}}_t, \mathbf{X}_t)$ is also parametrized by $\bbtheta$. This 
together with the Gaussian prior of $\bbtheta$ (cf. \eqref{eq:f_check}), yields the posterior $p(\bbtheta|\mathbf{y}_t, \mathbf{X}_t)$, based on which we can predict $f$ and $y$ at new test input $\mathbf x$. Specifically, upon replacing $p(f(\mathbf{x}_{t+1})|\mathbf{f}_t, \mathbf{X}_{t})$ and
$p(\mathbf{f}_t|\mathbf{y}_t, \mathbf{X}_t)$ in \eqref{eq:p_f_pre} by $p(\check{f}( \mathbf{x}_{t+1})|\bbtheta) = \delta(\check{f}(\mathbf{x}_{t+1}) - \bbphi_{\mathbf{v}}^\top (\mathbf{x}_{t+1}) \bbtheta)$ and $p(\bbtheta|\mathbf{y}_t, \mathbf{X}_t)$, respectively, we obtain the predictive pdf of the RF-based $\check{f}(\mathbf{x}_{t+1})$, which further leads to the predictive pdf of $y_{t+1}$ in \eqref{pdf4pred} after replacing $f(\mathbf{x}_{t+1})$ by $\check{f}(\mathbf{x}_{t+1})$. 
For GPR, the predictive pdf of $y_{t+1}$  is 
\begin{align}
	{p}(y_{t+1}|\mathbf{y}_t,\mathbf{X}_{t+1}) = \mathcal{N}(y_{t+1}; \hat{\check{y}}_{t+1|\mathbf{t}}, \check{\sigma}^2_{t+1|\mathbf{t}})
\end{align}
where
\begin{subequations}
	\vspace*{-0.3cm}
	\begin{align}
		\hat{\check{y}}_{t+1|\mathbf{t}}& = \bbphi_{\mathbf{v}}^\top(\mathbf{x}_{t+1}) \left( \bbPhi_t^{\top} \bbPhi_t+ \frac{\noisevar}{\sigma_\theta^2}\mathbf{I}_{2\nrf} \right)^{-1}\!\!\!\! \bbPhi_t^\top \mathbf{y}_t  \\
		\check{\sigma}^2_{t+1|\mathbf{t}}& = \bbphi_{\mathbf{v}}^\top(\mathbf{x}_{t+1})\!\! \left(\!\!\frac{\bbPhi_t^{\top} \bbPhi_t}{\noisevar} \!+\!  \frac{\mathbf{I}_{2\nrf}}{\sigma_\theta^2} \!\!\right)^{-1}\!\!\!\!\!\!\!\!\bbphi_{\mathbf{v}}(\mathbf{x}_{t+1}) \!+ \!\noisevar.
	\end{align}
\end{subequations}
This \emph{batch} predictor incurs complexity $\mathcal{O}(t(2\nrf)^2+(2\nrf)^3)$, which is dominated by $\mathcal{O}(t(2\nrf)^2)$ for $t\gg 2\nrf$.  This linear (in $t$) complexity is apparently much more affordable than the plain-vanilla GP predictor \eqref{eq:plain_gpp}.

The RF-based function approximant $\check{f}$ easily accommodates \emph{online} operation~\cite{gijsberts2013real}, which is called for in many time-critical applications, including time series prediction \cite{richard2008online}, and robot localization \cite{xu2014gp}. 
While the RF-based online approach for GPR is offered in~\cite{gijsberts2013real}, its performance hinges on a \emph{preselected} kernel for the GP prior, which may fall short in characterizing upcoming data samples.
Next, we will broaden the scope of a single GP prior by an ensemble (E) of GPs to enable real-time kernel adaptation. Besides serving the role of a non-Gaussian prior, EGP will turn out to be scalable too, after adopting once again the RF approximation. 

\section{Online scalable ensemble GPs } \label{sec:OIE-GP}

Towards data-driven kernel selection in the online setting, an EGP meta-learner employs an ensemble of $M$ GP experts (a.k.a. models or learners),
each of which places a unique GP prior on $f$ as $f|m \sim \mathcal{GP}(0, \kappa^m (\mathbf{x}, \mathbf{x}'))$, where $m\in \mathcal{M}:=\{1,\ldots,M\}$ is the expert index and $\kappa^m$ is a shift-invariant kernel selected from a {\it known} kernel dictionary $\mathcal{K}:=\{\kappa^1, \ldots, \kappa^M\}$.
Here, $\mathcal{K}$ should be constructed as large as computational constraints allow, depending on resources and the learning task. Per expert $m$, the prior pdf of function values at $\mathbf{X}_t$ is
\begin{align}
	p(\mathbf{f}_t|i\!=\! m, \mathbf{X}_t) \!=\! \mathcal{N} (\mathbf{f}_t ; {\bf 0}_t, {\bf K}^m_t ), \ [{\bf K}_t^m]_{\tau,\tau'} \!:=\!\kappa^m(\mathbf{x}_\tau, \mathbf{x}_{\tau'}\!)\nonumber
\end{align}
where the hidden random variable $i$ is introduced to denote the expert index.
The ensemble prior pdf of $\mathbf{f}_t$ that accounts for all GP experts is given by the Gaussian mixture (GM)\vspace*{-0.2cm}
\begin{align}
	p(\mathbf{f}_t|\mathbf{X}_t) = \sum_{m = 1}^M w^m \mathcal{N}(\mathbf{f}_t; {\bf 0}_t, \mathbf{K}^m_t)\;,\;\;  \;\;\;\; \sum_{m=1}^M w^m =1  \label{eq:EGP_prior}
\end{align}\\
where the {\it unknown} weights $\{w^m\}_{m=1}^M$, viewed as probabilities of the GP experts to be present in the EGPs, are to be learned from data that arrive sequentially.

Seeking a scalable predictor, each expert $m$ relies on the RF-based function approximant~\eqref{eq:f_check} with the per-expert parameter vector $\bbtheta^m$ and RF vector $\bbphi_{\mathbf{v}}^m(\mathbf{x})$ constructed as in \eqref{eq:phi_x} using $\{\mathbf{v}_j^m\}_{j=1}^{\nrf}$. Vectors $\{\mathbf{v}_j^m\}_{j=1}^{\nrf}$ here are drawn i.i.d. from $\pi_{\bar{\kappa}}^m (\mathbf{v})$, which is the PSD of the standardized kernel $\bar{\kappa}^m$,  relating to $\kappa^m$ through the magnitude $\sigma_{\theta^m}^2$ as $\kappa^m = \sigma_{\theta^m}^2\bar{\kappa}^m$. The per expert $m$ generative model for output $y$ is then\vspace*{-0.1cm}
\begin{subequations}
	\begin{align}
		{p}(\bbtheta^m) &= \mathcal{N} (\bbtheta^m; \mathbf{0}_{2\nrf}, \sigma_{\theta^m}^2\mathbf{I}_{2\nrf}) \label{eq:p_theta}\\
		p(y|\bbtheta^m, \mathbf{x}) &= p(y| \bbphi_{\mathbf{v}}^{m\top} (\mathbf{x})\bbtheta^m)\;.\label{eq:LF}
	\end{align}
\end{subequations}

Focusing on the incremental (I) setting, each expert $m$ interleaves prediction of $y_{t+1}$ based on $p(\bbtheta^m|\mathbf{y}_t, \mathbf{X}_t)$, and update of the parameter posterior upon the arrival of $y_{t+1}$ per slot.
To assess the per-expert contribution, the EGP meta-learner relies on the posterior probability $w_t^m:={\rm Pr}(i=m|\mathbf{y}_{t};\mathbf{X}_{t} )$. As we shall see next, the resultant IE-GP proceeds in two steps, namely prediction and correction, by propagating per-expert weights and posterior pdfs $\{w_t^m, {p}(\bbtheta^m|\mathbf{y}_t,\mathbf{X}_t)\}_{m = 1}^M$ from slot to slot.\\

\noindent {\bf Prediction.}
Upon receiving $\mathbf{x}_{t+1}$, each expert $m$ constructs the RF vector using $\bbphi_{\mathbf{v}}^m (\mathbf{x}_{t+1})$ as in \eqref{eq:phi_x}. With ${p}(\bbtheta^m |\mathbf{y}_t,\mathbf{X}_{t})$ available from slot $t$,  the per-expert predictive pdf of $y_{t+1}$ can be obtained by invoking the sum-product probability rule
\begin{align}
	&{p}(y_{t+1}|\mathbf{y}_{t}, i=m,\mathbf{X}_{t+1})    \!\! =\!\! \int\!\! p(y_{t+1}| \bbtheta^m\! ,\mathbf{x}_t) 
	p(\bbtheta^m|\mathbf{y}_t,\mathbf{X}_{t}) d \bbtheta^m.  \label{eq:p_y_pre_1}
\end{align}
Leveraging again the sum-product rule, the EGP meta-learner seeks the ensemble predictive pdf as \vspace*{-0.2cm}
\begin{align}
	{p}(y_{t+1}|\mathbf{y}_{t},\mathbf{X}_{t+1}) \! &=\!\! \sum_{m = 1}^M\!\! {\rm Pr}(i\!=\! m|\mathbf{y}_{t},\!\mathbf{X}_{t}) {p}(y_{t+1}\!|\mathbf{y}_{t}, \! i\! =\! m,\!\mathbf{X}_{t+1}\! ) \nonumber\\
	& = \!\! \sum_{m = 1}^M w_t^m {p}(y_{t+1}|\mathbf{y}_{t},\! i\! =\! m,\mathbf{X}_{t+1})
	\label{eq: p_y_pred_ens}
\end{align}\vspace*{-0.3cm}\\
which takes an intuitive form as a weighted combination of predictions from the individual GP experts.
Having available the predictive pdf, we are ready to update the posterior pdf of the RF model parameter vector.  \\

\noindent {\bf Correction.}
With the arrival of $y_{t+1}$, each expert $m$ updates the posterior pdf of $\bbtheta^m$ via Bayes' rule as 
\begin{equation}
{p}(\bbtheta^m |\mathbf{y}_{t+1},\mathbf{X}_{t+1})  =  	\frac{{p}(\bbtheta^m |\mathbf{y}_{t},\mathbf{X}_{t}\!) {p}(y_{t+1}|\bbtheta^m,\mathbf{x}_{t+1})}
{{p}(y_{t+1}|\mathbf{y}_{t}, i=m,\mathbf{X}_{t+1})}
\label{eq:p_theta_up_1} 
\end{equation}
where ${p}(y_{t+1}|\bbtheta^m,\mathbf{x}_{t+1})$ is the known likelihood (cf.~\eqref{eq:LF}), and ${p}(\bbtheta^m |\mathbf{y}_{t},\mathbf{X}_{t})$ is available from slot $t$. For later use, the so-termed Bayesian loss incurred by expert $m$ at slot $t+1$ is (cf.  \cite{kakade2005online}) 
\begin{align}
	l_{t+1|t}^m :=-\log {p}(y_{t+1}|\mathbf{y}_{t}, i=m,\mathbf{X}_{t+1}) \label{eq:loss_s}
\end{align}
whose ensemble version is given by 
\begin{align}
	\ell_{t+1|t} &:=  -\log {p}(y_{t+1}|\mathbf{y}_{t},\mathbf{X}_{t+1}) \nonumber\\
	&\ = -\log \!\!\sum_{m = 1}^M\!\! w_{t}^m \exp\!\left(- l_{t+1|t}^m \!\right). \label{eq_loss_update}
\end{align}\\
Simultaneously, the EGP meta-leaner obtains the updated weight $w_{t+1}^m := {\rm Pr}(i=m|\mathbf{y}_{t+1},\mathbf{X}_{t+1})$ as
\begin{align}
	w_{t+1}^m &
	= \frac{{\rm Pr}(i=m|\mathbf{y}_{t},\mathbf{X}_{t}) {p}(y_{t+1}|\mathbf{y}_{t},  i=m,\mathbf{X}_{t+1})}{{p}(y_{t+1}|\mathbf{y}_{t},\mathbf{X}_{t+1})} \nonumber\\ & = w_t^m \exp (\ell_{t+1|t} - l_{t+1|t}^m)   \label{eq:w_update_1}
\end{align}
where $w_t^m$ is available from slot $t$. Intuitively, large $l_{t+1|t}^m$ implies small  $\ell_{t+1|t}-l_{t+1|t}^m$, and thus $w_{t+1}^m$ relative to the rest will be smaller than that at slot $t$.   

Summarizing, our scalable IE-GP algorithm for general likelihoods (and thus posteriors) relies on \eqref{eq:p_y_pre_1}-\eqref{eq:p_theta_up_1} to transition from slot $t$ to slot $t+1$. Next, we specialize our novel IE-GP to GPR that enjoys closed-form pdf and weight updates, as well as non-Gaussian likelihoods that entail Laplace approximation~\cite{williams1998bayesian} to evaluate the (possibly high-dimensional) integrals in the prediction and correction steps.

\begin{algorithm}[t]
	\caption{IE-GP for GPR}\label{Alg: EM_static}
	\begin{algorithmic}[1]
		\State{\textbf{Input:}  $\kappa^m$, $m = 1,\ldots, M$, and number of RFs $\nrf$}.
		\State{\textbf{Initialization:}  }
		\For{$m = 1, 2, \ldots, M$}
		\State Draw $\nrf$ random vectors $\{\mathbf{v}_i^m\}_{i = 1}^{\nrf}$;
		\State  $w_0^m = 1/M$;	$\hat{\bbtheta}_0^m = \mathbf{0}_{2D}$; $\bbSig_{0}^m = \sigma_{\theta^m}^2\mathbf{I}_{2D}$;
		\EndFor
		\newline
		\For{$t = 1, 2, \ldots, T$}
		\State Receive input datum $\mathbf{x}_t$ ;
		\For{$m = 1, 2, \ldots, M$}
		\State Construct RF $\bbphi_\mathbf{v}^m(\mathbf{x}_t)$ via \eqref{eq:phi_x};
		\State  \begin{varwidth}[t]{\linewidth}
			Obtain per-expert pdf of $y_t$ via \eqref{eq:p_y_pre_GPR};
		\end{varwidth}
		\State Update $w_{t}^m$ via \eqref{eq:w_update_2};
		\State  \begin{varwidth}[t]{\linewidth}
			Update per-expert pdf of $\bbtheta^m$ via \eqref{eq:p_theta_up_2};
		\end{varwidth}	 
		\EndFor
		\EndFor
	\end{algorithmic}
\end{algorithm}

\subsection{Closed-form updates for GPR}
For GPR, the likelihood per expert is given by ${p}(y_t|\bbtheta^m, \mathbf{x}_t) = \mathcal{N}(y_t; \bbphi_{\mathbf{v}}^{m\top}(\mathbf{x}_t)\bbtheta^m, \noisevar)$, which together with the per-expert Gaussian prior $p(\bbtheta^m)$ (cf. \eqref{eq:p_theta}), yields the Gaussian posterior at the end of slot $t$ expressed as
\begin{align}
	{p}(\bbtheta^m | \mathbf{y}_t,\mathbf{X}_t) = \mathcal{N}(\bbtheta^m; \hat{\bbtheta}_{t}^m, \bbSig^m_{t})\label{eq:p_theta_post_static}
\end{align}\vspace*{-0.4cm}\\
with mean $\hat{\bbtheta}_{t}^m$ and covariance matrix $\bbSig^m_{t}$ per expert $m$.

Building on \eqref{eq:p_theta_post_static} and \eqref{eq:p_y_pre_1}, the predictive pdf of $y_{t+1}$ from expert $m$ is also Gaussian 
\begin{align}
	{p}(y_{t+1}|\mathbf{y}_{t}, i\! =\! m,\mathbf{X}_{t+1}) \! =\!  \mathcal{N} \left(y_{t+1};  \hat{y}_{t+1|t}^{m}, 	(\sigma_{t+1|t}^{m})^2 \right) \label{eq:p_y_pre_GPR}
\end{align}
where the predicted mean and variance are 
\begin{subequations}
	\label{eq:26}
	\begin{align}
		\hat{y}_{t+1|t}^{m} &=  \bbphi^{m\top}_{\mathbf{v}} (\mathbf{x}_{t+1})\hat{\bbtheta}_t^{m} \\	
		(\sigma_{t+1|t}^{m})^2 & = \bbphi^{m\top}_{\mathbf{v}}  (\mathbf{x}_{t+1}) \bbSig^m_{t} \bbphi^m_{\mathbf{v}} (\mathbf{x}_{t+1})+\noisevar\;.	
	\end{align}
\end{subequations}
Thus, the ensemble predictive pdf of $y_{t+1}$ in \eqref{eq: p_y_pred_ens} specialized to GPR is a GM, based on which the EGP meta-learner obtains the minimum mean-square error (MMSE) predictor of $y_{t+1}$ together with the associated variance as
\begin{subequations}\label{eq:y_pre_static}
	\begin{align}
		\hat{y}_{t+1|t}  & =   \sum_{m = 1}^M w_t^m \hat{y}_{t+1|t}^{m}\label{eq:mean_f_pre} \\	
		\sigma_{t+1|t}^2  & = \sum_{m = 1}^M w_t^m [(\sigma_{t+1|t}^{m})^2+  (\hat{y}_{t+1|t}\! -\! \hat{y}_{t+1|t}^{m})^2]. \label{eq:var_f_pre}
	\end{align}
\end{subequations}
When $y_{t+1}$ becomes available, the EGP meta-learner updates the per-expert weight as (cf.~\eqref{eq:w_update_1} and \eqref{eq:p_y_pre_GPR})  \vspace*{-0.1cm}
\begin{align}
	w_{t+1}^m   = \frac{w_t^m \mathcal{N}\left(y_{t+1};  \hat{y}_{t+1|t}^{m}, (\sigma_{t+1|t}^{m})^2 \right)}{\sum_{m' = 1}^M w_t^{m'} \mathcal{N}\left(y_{t+1};  \!\hat{y}_{t+1|t}^{m'}, (\sigma_{t+1|t}^{m'})^2 \right)}\;. \label{eq:w_update_2}
\end{align}
With the per-expert Gaussian likelihood, the arrival of $y_{t+1}$ also propagates Gaussianity of the posterior pdf of $\bbtheta^m$ from slot $t$ to $t+1$, expressed as
\begin{align}
	{p}(\bbtheta^m |\mathbf{y}_{t+1},\mathbf{X}_{t+1}) 
	& = \mathcal{N}(\bbtheta^m; \hat{\bbtheta}_{t+1}^m, \bbSig^m_{t+1}) \label{eq:p_theta_up_2}
\end{align}
where the per-expert mean $ \hat{\bbtheta}_{t+1}^m$ and covariance matrix $\bbSig^m_{t+1}$ are\hspace*{-0.2cm}
\begin{subequations} \label{eq:posterior_update_1}
	\begin{align}
		\hspace*{-0.2cm}\hat{\bbtheta}_{t+1}^m &= \hat{\bbtheta}_{t}^m \!+\!  (\sigma_{t+1|t}^{m})^{-2}\bbSig^m_{t} \bbphi^m_{\mathbf{v}}(\mathbf{x}_{t+1})(y_{t+1} \!-\! \hat{y}_{t+1|t}^{m}) \\
		\bbSig_{t+1}^m &= \bbSig_{t}^m\! \!-\!  (\sigma_{t+1|t}^{m})^{-2}\bbSig^m_{t} \bbphi^m_{\mathbf{v}}\!(\mathbf{x}_{t+1})  \bbphi^{m\top}_{\mathbf{v}}\!\!(\mathbf{x}_{t+1})\bbSig^m_{t} . 
	\end{align}
\end{subequations}
Accounting for all $M$ expert updates, our scalable IE-GP approach to GPR (see Algorithm 1) has per-iteration complexity of $\mathcal{O}(M (2\nrf)^2)$; hence, scalability is not compromised by the ensemble approach that also offers a richer model for the learning function.  The deterministic RKHS online approach (termed ``Raker" in~\cite{shen2019random}) relies on first-order gradient descent to update $\bbtheta$ at per-iteration complexity of $\mathcal{O}(M\nrf d)$, which is lower than our second-order update in \eqref{eq:posterior_update_1}. Our probabilistic IE-GP approach offers numerically improved performance that is also analytically quantifiable through the predictor variance \eqref{eq:var_f_pre} in \eqref{eq:mean_f_pre}.

\subsection{Laplace approximation for non-Gaussian likelihood}\label{sec:LA}
When the likelihood is non-Gaussian as in classification, Poisson regression or ordinal regression, the per-expert parameter posterior is no longer available in closed form. Fortunately, the so-termed Laplace approximation~\cite{williams1998bayesian} can be leveraged to carry out the prediction and correction steps with tractability in IE-GP. Specifically, each expert maintains a Gaussian approximant for the parameter vector as $p(\bbtheta^m |\mathbf{y}_{t},\mathbf{X}_{t})\approx \mathcal{N}(\bbtheta^m; \hat{\bbtheta}^m_t, \bbSig_t^m)$, which readily allows the predicted pdf of $y_{t+1}$~\eqref{eq:p_y_pre_1} to be calculated via Monte-Carlo sampling, or through probit approximation for binary classification with the logistic likelihood; see, e.g., \cite[Chapter~8.4.4.2]{murphy2012machine}. 

Given $y_{t+1}$, the EGP meta-learner updates the per-expert weight as in~\eqref{eq:w_update_1} with the per-expert loss at hand from the prediction step. Also, each expert relies on Laplace approximation to seek an updated Gaussian approximant as
 ${p}(\bbtheta^m|\mathbf{y}_{t+1},\mathbf{X}_{t+1}) \approx
	\mathcal{N}(\bbtheta^m; \hat{\bbtheta}_{t+1}^m, \bbSig^m_{t+1})$,
where $\hat{\bbtheta}^m_{t+1}$ and  $\bbSig^m_{t+1}$ are respectively the mode of the log posterior and the corresponding information matrix inverse, that can be obtained by solving the following optimization problem using Newton's iteration~\cite{murphy2012machine}
\begin{subequations}
\begin{align}
&\hspace{-0.1cm}\hat{\bbtheta}_{t+1}^m = \underset{\bbtheta^m}{\arg \max} \log {p}(\bbtheta^m |\mathbf{y}_{t},\mathbf{X}_{t}) + \log p(y_{t+1}|\bbtheta^m, \mathbf{x}_{t+1}) \nonumber\\
&\hspace{-0.1cm}(\bbSig^m_{t+1})^{-1} = (\bbSig^m_{t})^{-1}\! \!-\!\nabla^2_{\theta^m}\log p(y_{t+1}|\bbtheta^m\!\!, \mathbf{x}_{t+1})\Big\rvert_{\bbtheta^m = \hat{\bbtheta}_{t+1}^m}\nonumber.
\end{align}
\end{subequations}

%


\section{Regret analysis} \label{sec:reg_anal}
The pdf ${p}(y_{t+1}|\mathbf{y}_{t},\mathbf{X}_{t+1})$ in \eqref{eq: p_y_pred_ens} provide an online performance metric for ${\hat y}_{t+1|t}$, from which its mean and variance can be also obtained (even in closed form, cf.~\eqref{eq:var_f_pre}). These metrics however, rely on the assumption of knowing the prior pdf of $f$, and the conditional data likelihood. To guard against having \emph{imperfect knowledge} of these pdfs (the norm in adversarial settings), regret analysis is well motivated along the lines of online convex optimization~\cite{hazan2016introduction} and online learning with expert advice ~\cite{cesa2006prediction}. This is the subject of this section that aims to benchmark  performance of our IE-GP predictor relative to the best function estimator with data in hindsight when the generative assumptions are violated.

To this end, let $\mathcal{L}(f(\mathbf{x}_\tau);y_\tau) := - \log p(y_\tau|f(\mathbf{x}_\tau))$ be the per-slot negative log-likelihood (NLL). For any fixed function estimator $\hat{f}^{*}(\cdot)$, the incurred loss over $T$ slots is $\sum_{\tau = 1}^T \mathcal{L}({\hat f}^{*}(\mathbf{x}_\tau); y_\tau)$. With the EGP prior in \eqref{eq:EGP_prior}, the best function estimate (benchmark) with data $\{\mathbf{X}_T, \mathbf{y}_T\}$ available in hindsight, are obtained with the optimal weights $\{w^m\}$ in the EGP prior by maximizing the batch function posterior, $p(\mathbf{f}_T|\mathbf{y}_T, \mathbf{X}_T) \propto p(\mathbf{f}_T| \mathbf{X}_T)p(\mathbf{y}_T|\mathbf{f}_T, \mathbf{X}_T)$, as\vspace*{-0.2cm}
\begin{align}
	(\hat{\mathbf{f}}_T, \{\hat{w}^m\}) =  \underset{\substack{\mathbf{f}_T, \{w^m\}\\  \sum_{m} w^m =1} }{\arg\max} 
	p(\mathbf{y}_T|\mathbf{f}_T, \mathbf{X}_T) \!\!\sum_{m=1}^M \!w^m p(\mathbf{f}_T|i\! =\! m, \mathbf{X}_T) \nonumber
\end{align}
whose solution is $\hat{w}^{m^*} = 1$ and $\hat{w}^m = 0$ for $m\neq m^*$. This implies that only one GP expert $m^{*}$ is active in the benchmark function estimate for $\tau=1,\ldots,T$. The optimal estimate by expert $m^{*}$ are then given by\vspace{-0.1cm}
\begin{align}
	\hat{\mathbf{f}}_T = \underset{\mathbf{f}_T}{\arg\max} \ \  p(\mathbf{f}_T|i=m^{*}, \mathbf{X}_T)p(\mathbf{y}_T|\mathbf{f}_T, \mathbf{X}_T)\;. \label{eq:f_hat_fixed}
\end{align}
As every positive semidefinite kernel $\kappa^m$
is associated with a unique RHKS $\mathcal{H}^m$\cite{kakade2006worst}, the optimal function estimator $\hat{f}^{m^*} (\cdot)$ is extracted from \eqref{eq:f_hat_fixed} as \vspace*{-0.2cm}
\begin{align}
	m^*\in &\: \underset{ m\in\mathcal{M}}  {\arg \min}  \sum_{\tau = 1}^T {\mathcal L} ({\hat f}^m(\mathbf{x}_\tau); y_\tau) + \frac{1}{2}\| {\hat f}^m\|_{\mathcal{H}^m}^2 \label{eq:f_star} 
\end{align}\vspace*{-0.3cm}\\
where the optimal function estimator per expert $\hat{f}^m (\cdot)$, $m=1,\ldots,M$, is obtained as \vspace*{-0.2cm}
\begin{align}
	{\hat f}^m(\cdot) \in & \: \underset{f^m \in \mathcal{H}^m} {\arg \min}  \sum_{\tau = 1}^T {\mathcal L} (f^m(\mathbf{x}_\tau) ; y_\tau)+ \frac{1}{2}\| f^m\|_{\mathcal{H}^m}^2\:. \nonumber
\end{align}\vspace*{-0.35cm}\\
With the best fixed function estimator $\hat{f}^{m^*}(\cdot)$ at hand, the static regret over $T$ slots is then defined as \cite{kakade2005online}\vspace*{-0.1cm}
\begin{align}
	\mathcal{R}^{\rm ST}(T) := 
	\sum_{\tau = 1}^T \ell_{\tau|\tau-1} -\sum_{\tau = 1}^T \mathcal{L}({\hat f}^{m^*}(\mathbf{x}_\tau); y_\tau)
	\label{eq:static_regret}
\end{align}
where $\ell_{\tau|\tau-1}$, defined in \eqref{eq_loss_update}, captures the ensemble online Bayesian loss incurred by IE-GP. 

 \begin{figure*}[t]
	\minipage[b]{0.33\textwidth}
	\centering
	\includegraphics[width=0.99\linewidth]{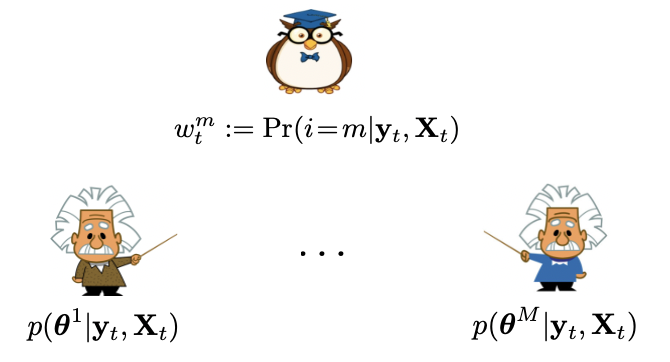}
	{(a)} \label{fig:I_EGP}
	\endminipage\qquad
	\hspace{-4mm}
	\minipage[b]{0.33\textwidth}
	\centering
	\includegraphics[width=0.99\linewidth]{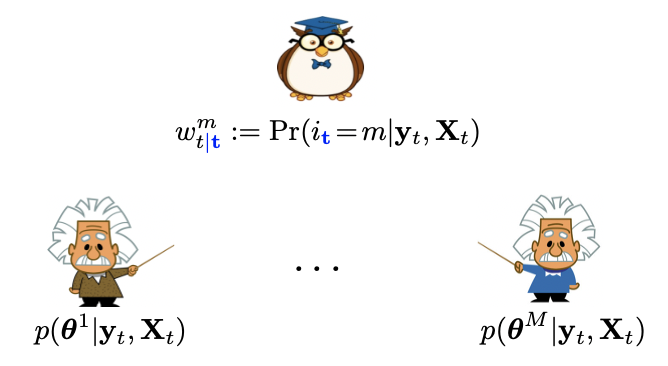}
	{(b)} \label{fig:SI_EGP}
	\endminipage\qquad
	\hspace{-4mm}
	\minipage[b]{0.33\textwidth}
	\centering
	\includegraphics[width=.99\linewidth]{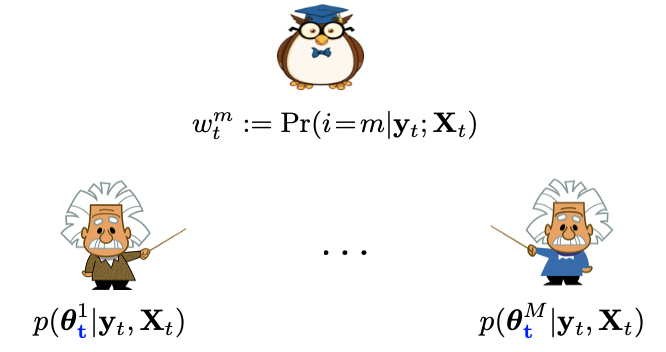}
   {(c)}  \label{fig:DI_EGP}
	\endminipage\qquad
	\caption{Schematic diagrams for (a) IE-GP, (b) SIE-GP, and (c) DIE-GP. The EGP meta-learner (Owl) assesses the contributions of GP experts (Einsteins) via the posterior probabilities of time-invariant $i$ or time-varying $i_t$. The GP experts summarize past data in the posterior of $\bbtheta^m$ or $\bbtheta^m_t$. }
\label{fig:EGP_diagrams}
	\vspace*{-0.3cm}
\end{figure*}

Although the cumulative online loss in the first sum of \eqref{eq:static_regret} has different form than that of the benchmark, they are comparable by the data likelihood, where the function is nonrandom. In other words, the online Bayesian loss is obtained by taking the expectation of the likelihood wrt the online predictive pdf of the function, thus eliminating the randomness of the function in the likelihood.

To proceed, we will need the following assumptions. 
\begin{itemize}
	\item[{\bf (as1)}] The NLL $\mathcal{L}(z_\tau; y_\tau)$  is continuously twice differentiable with $| \frac{d^2}{d z_\tau^2} \mathcal{L}(z_\tau;y_\tau) | \leq c, \forall z_\tau$;
	\item[{\bf (as2)}] The NLL $\mathcal{L}(z_t;y_t)$ is convex and has bounded derivative wrt $z_t$; that is, $| \frac{d}{d z_t} \mathcal{L}(z_t;y_t) | \leq L$;
	\item[{\bf (as3)}] Kernels $\{\bar{\kappa}^m\}_{s = 1}^M$ are shift-invariant, standardized and bounded, that is $\bar{\kappa}^m (\mathbf{x}_t, \mathbf{x}_{t'}) \leq 1, \forall \mathbf{x}_t, \mathbf{x}_{t'}$.
\end{itemize}
Differentiability and convexity of the NLL in {(as1)}-{(as2)} are satisfied by most forms of likelihood in GP-based learning, including the Gaussian likelihood in GPR, and the logistic one for classification. Conditions in {(as3)} hold for a wide class of kernels including Gaussian, Laplace and Cauchy ones~\cite{rahimi2008random}. As the derivations rely on the general form of IE-GP (cf. \eqref{eq:p_y_pre_1}-\eqref{eq:p_theta_up_1}) that corresponds to general likelihoods, the regret bound established here applies to general learning tasks.


To establish the static regret bound of IE-GP, we will need the following intermediate lemma. \\

\noindent {\bf Lemma 1.} \textit{	Under {(as1)}, and with prior of $\bbtheta^m$ given by \eqref{eq:p_theta}, the following bound holds concerning the cumulative online Bayesian loss incurred by the IE-GP and the counterpart from a single RF-based GP expert with fixed $\bbtheta^{m}_{*}$
	\begin{align}
		&\sum_{\tau = 1}^T \ell_{\tau|\tau-1} - \sum_{\tau = 1}^T \mathcal{L} (\bbphi^{m\top}_{\mathbf{v}}(\mathbf{x}_\tau)\bbtheta^{m}_{*}; y_\tau) \nonumber\\
		&\leq  \frac{\|\bbtheta^{m}_{*}\|^2}{2\sigma_{\theta^m}^2} + \nrf \log \left(1+ \frac{T \sigma_{\theta^m}^2}{2\nrf} \right) + \log M\label{eq:L_OB} \;.
	\end{align}}

\noindent {\bf Proof: }
	See Sec.~\ref{sec:proof_Lemma1}.\\

Lemma 1 bounds the cumulative online Bayesian loss of IE-GP relative to any single RF-based GP learner with a fixed strategy. Next, we will work towards the ultimate static regret by further bounding the loss of RF-based function estimator relative to the best function estimator in the original RKHS for each expert.

\noindent {\bf Theorem 1.}
\textit{Under {as(1)}-{as(3)} and with $\hat{f}^{m^*}$ belonging to the RHKS $\mathcal{H}^{m^*}$ induced by $\kappa^{m^*}$, for a fixed $\epsilon >0$, the following bound holds with probability at least $1-2^8 (\frac{\sigma_{m^*}}{\epsilon})^2 \exp\left(\frac{-\nrf\epsilon^2}{4d+8} \right)$
	\begin{align}
		\hspace*{-0.1cm}
		&\sum_{\tau = 1}^T \ell_{\tau|\tau-1} -  \sum_{\tau = 1}^T \mathcal{L}(\hat{f}^{m^{*}}(\mathbf{x}_\tau); y_\tau) \leq \\
		& \frac{(1+\epsilon)C^2}{2\sigma_{\theta^{m^*}}^2} + \nrf \log \left(1+ \frac{Tc \sigma_{\theta^{m^*}}^2 }{2\nrf} \right)  + \log M + \epsilon LTC  \nonumber
	\end{align}\\
	where $C$ is a constant, and $\sigma_{m^*}^2 : = \mathbb{E}_{\pi^{m^*}_{\bar{\kappa}}}[\| \mathbf{v}^{m^*}\|^2]$ is the second-order moment of  $\mathbf{v}^{m^*}$. 
	Setting $\epsilon = \mathcal{O}(\log T/T)$, the static regret in \eqref{eq:static_regret} boils down to 
	\begin{align}
		\mathcal{R}^{\rm ST} (T) = \mathcal{O}(\log T)\;. \label{eq:reg_full}
	\end{align}}

\noindent {\bf Proof.}
	See Sec.~\ref{sec:proof_Thm1}.

\noindent Theorem 1 asserts that IE-GP incurs no regret on average with cumulative static regret $\mathcal{O}(\log T)$ over $T$ slots, thereby demonstrating its robustness to the adversarial setting. It is also worth highlighting that this regret bound is tighter than that of the deterministic RKHS-based online multi-kernel counterpart \cite{shen2019random} with regret $\mathcal{O}(\sqrt{T})$ in the static setting.

\section{EGP for dynamic learning}\label{sec:dynamic}
In the proposed IE-GP, the EGP meta-learner relies on the posterior probability of a {\it static} random variable $i$ to assess contributions of the GP experts, each of which models the learning function via the {\it time-invariant} parameter vector $\bbtheta^m$. Such a stationary setting implies that  IE-GP  handles no dynamics in the unknown function. This is also manifested in Sec.~\ref{sec:reg_anal} that with batch data in hindsight the optimal function estimate is associated with one of the GP experts (cf. \eqref{eq:f_hat_fixed}). To further enable learning for dynamic functions, the rest of this section will explore extensions to accommodate {\it time-varying} $i_t$ and $\bbtheta^m_t$ for the EGP meta-learner and individual GP learners, respectively.

\subsection{Dynamics at EGP meta-learner}
Capitalizing on time-dependent $i_t\in \mathcal{M}$ to denote the index of the contributing expert, the EGP meta-learner models the evolution of $i_t$ via a Markov chain with prior transition probability $q_{mm'}:={\rm Pr}(i_{t+1}=m|i_t=m')$ for $m, m'\in \mathcal{M}$. Such a dynamic model allows the learning function to jump among the candidate spaces associated with the GP experts, yielding the so-termed switching (S) IE-GP hereafter. The values of $\{q_{mm'}\}_{m,m'}\in [0,1]$ are user-defined parameters. It is worth pointing out that IE-GP can be regarded as a special case of SIE-GP with $q_{mm}=1$.

The novel SIE-GP differs from IE-GP in the weight update. To illustrate this, the per-expert posterior weight in SIE-GP is first adapted as $w_{t|t}^m := {\rm Pr} (i_t=m | \mathbf{y}_t, \mathbf{X}_t)$ given time-varying $i_t$. Before propagating to $w_{t+1|t+1}^m$, the EGP meta-learner leverages the aforementioned Markov transition model to predict the weight for GP model $m$ at slot $t+1$ via  $w_{t+1|t}^m :={\rm Pr}(i_{t+1}\! =\! m|\mathbf{y}_t, \mathbf{X}_t )$, which, with $\{w_{t|t}^m\}$ available, can be obtained as \vspace*{-0.2cm}
\begin{align}
	w_{t+1|t}^m &=\!\! \sum_{m'=1}^M\! {\rm Pr}(i_{t+1}=m, i_t=m'|\mathbf{y}_t, \mathbf{X}_t )  \nonumber\\
	& =\!\! \sum_{m'=1}^M {\rm Pr}(i_{t+1}\!=\! m|i_t\! =\! m', \mathbf{y}_t, \mathbf{X}_t ){\rm Pr}(i_{t}\! =\! m'| \mathbf{y}_t, \mathbf{X}_t )\nonumber\\
	& =\!\!\sum_{m'=1}^M  q_{m,m'} w_{t|t}^{m'}  \label{eq:w_pre_s}
\end{align}
where the last equality holds since the evolution of $i_t$ is independent from $\mathbf{y}_t$ and $\mathbf{X}_t$.  With non-zero $\{q_{mm'}\}$, the prediction rule in \eqref{eq:w_pre_s} allows for activation of the previously inactive expert $m$ ($w_{t|t}^m\approx0$), thereby accommodating switching among candidate GP models or function spaces.

With the predicted weights \eqref{eq:w_pre_s}  and the per-expert predictive pdf \eqref{eq:p_y_pre_1} available, the EGP meta-learner leverages the sum-product probability rule to predict the pdf of $y_{t+1}$
\begin{align}
	&{p}(y_{t+1}|\mathbf{y}_{t},\mathbf{X}_{t+1})  = \sum_{m = 1}^M  {p}(y_{t+1}, i_{t+1}=m|\mathbf{y}_{t},\!\mathbf{X}_{t+1} ) \nonumber\\
	&= \sum_{m = 1}^M {\rm Pr}(i_{t+1}=m|\mathbf{y}_t, \mathbf{X}_t ) {p}(y_{t+1}|\mathbf{y}_{t}, i_{t+1}\!=\! m,\!\mathbf{X}_{t+1} ) \nonumber\\
	&= \sum_{m = 1}^M w_{t+1|t}^m\ {p}(y_{t+1}|\mathbf{y}_{t}, i_{t+1}\!=\! m,\mathbf{X}_{t+1} ) 
	\label{eq_pred_pdf_d}
\end{align}
where $w_{t+1|t}^m$ replaces $w_{t}^m$ in \eqref{eq: p_y_pred_ens} for IE-GP.
To facilitate the upcoming regret analysis, the aggregated online loss for SIE-GP that accounts for the per-expert loss \eqref{eq:loss_s} is defined as
\begin{align}
	\ell_{t+1|t}^{\rm SW}:= -\log \sum_{m = 1}^M w_{t+1|t}^m \exp\left(- l_{t+1|t}^m \right) \label{eq:track_loss}
\end{align}
where the superscript ``SW", denoting the switching scenario, is used to distinguish from the loss from IE-GP in \eqref{eq_loss_update}. 

Upon acquiring $y_{t+1}$, the EGP meta-leaner then updates the per-expert weight as
\begin{align}
	w_{t+1|t+1}^m&:= \frac{w_{t+1|t}^m p(y_{t+1}|\mathbf{y}_{t}, i_{t+1}\! =\! m, \mathbf{X}_{t+1})}{\sum_{m'=1}^M w_{t+1|t}^{m'} p(y_{t+1}|\mathbf{y}_{t}, i_{t+1}\! =\! m', \mathbf{X}_{t+1})}\nonumber\\
	&\ = w_{t+1|t}^m \exp(\ell_{t+1|t}^{\rm SW}- l_{t+1|t}^m )\;. \label{eq:SIEGP_w}
\end{align}
To sum up, relative to IE-GP, the meta-leaner in SIE-GP performs an additional weight prediction step \eqref{eq:w_pre_s}, which contributes to the ensemble predictive pdf \eqref{eq_pred_pdf_d} and the weight update \eqref{eq:SIEGP_w}. Next, the regret analysis of SIE-GP will be conducted in line with Sec.~\ref{sec:reg_anal} to account for the adversarial setting.

\subsubsection{Switching regret analysis}
With the underlying assumption that the active model per slot changes over time, the notion of switching or shifting regret \cite[Chapter 5.2]{cesa2006prediction} is leveraged to analyse the performance of SIE-GP in the adversarial setting where the generative assumptions are violated. Specifically, SIE-GP is compared with an arbitrary sequence of benchmark functions $\{\hat{f}^{i_\tau} \in \mathcal{H}^{i_\tau}\}_{\tau=1}^T$ with data in hindsight, yielding the switching regret defined as
\begin{align}
	\mathcal{R}^{\rm SW}(T):= \sum_{\tau = 1}^T \ell_{\tau|\tau-1}^{\rm SW} - \underset{i_1,\ldots,i_T}{\min}
	\sum_{\tau = 1}^T \mathcal{L}(\hat{f}^{i_\tau}(\mathbf{x}_\tau); y_\tau) \label{eq:regret_tr}
\end{align}
where, as in static regret \eqref{eq:static_regret}, the loss incurred by the comparator in the switching case is measured via the NLL.

Aiming at establishing an upper bound for \eqref{eq:regret_tr}, the following two additional assumptions will be entailed.
\begin{itemize}
	 \item[{\bf (as4)}] $q_{mm} = q_0$, $q_{mm'} = \frac{q_1}{M-1}$ for $m, m'\in \mathcal{M}$, $q_0+q_1=1$, and $0\leq q_1<\frac{1}{2}<q_0\leq 1$;
	\item[{\bf (as5)}] The number of switches for sequence $\{i_1, \ldots, i_T\}$ is upper bounded by $S$, i.e., $\sum_{\tau=1}^T I(i_{\tau} \neq i_{\tau+1})\leq S $, and $S\ll T$.
\end{itemize}
The transition probabilities dictated by (as4) yield a weight evolution strategy that is similar to the one given by the fixed-share forecaster in online expert-based learning \cite[Chapter 5.2]{cesa2006prediction}, where the conditions in (as5) are also leveraged in the regret analysis to bound the variation of the benchmark functions. 
Before obtaining the upper bound for \eqref{eq:regret_tr} in Theorem 2, the following two intermediate lemmas will be established first.\\

\noindent {\bf Lemma 2.}	{\it Under {as(4)}-{as(5)}, the following bound holds true concerning the cumulative ensemble switching loss and the single expert-based online counterpart for any sequence $\{i_1, \ldots, i_T\}$ 
\begin{align}
\sum_{\tau = 1}^T\! \ell^{\rm SW}_{\tau|\tau-1}\!-\!\! \sum_{\tau = 1}^T l^{i_\tau}_{\tau|\tau-1}\leq  \log M \!+\!  S \log T \!- \! S \log S \!+\! S\;. \label{eq:One_seq}
\end{align}}\\
\noindent {\bf Proof:} See Sec.~\ref{sec:proof_Lemma2}.\\

\noindent {\bf Lemma 3.} {\it Under (as1) and for any sequence $\{i_1, \ldots, i_T\}$, the following bound holds regarding the difference of the cumulative single expert-based online loss and the counterpart incurred by RF-based benchmark functions with fixed parameters $\{\bbtheta^{m}_{*}\}_{m=1}^M$
\begin{align}
	&\sum_{\tau = 1}^T l^{i_\tau}_{\tau|\tau-1} - \sum_{\tau = 1}^T \mathcal{L} (\bbphi^{i_\tau\!\top}_{\mathbf{v}}(\mathbf{x}_\tau)\bbtheta^{i_\tau}_{*}; y_\tau) \nonumber\\
	&	\leq  \sum_{m=1}^M \frac{\|\bbtheta^{m}_*\|^2}{2\sigma_{\theta^m}^2} + \nrf M\log\left(1+ \frac{Tc\sigma_{\theta^{*}}^2}{2\nrf M} \right) \label{eq:Lemma3}
	\end{align}
	where $\sigma_{\theta^{*}}^2 := \max_{m\in\mathcal{M}} \sigma_{\theta^{m}}^2$.
}

\vspace{0.2cm}
\noindent {\bf Proof:} See Sec.~\ref{sec:proof_Lemma3}.\\

\noindent {\bf Theorem 2.}
\textit{	Under {as(1)}-{as(5)} and with $\hat{f}^{m}$ belonging to the RHKS $\mathcal{H}^{m}$ induced by $\kappa^{m}$, for a fixed $\epsilon >0$, the following bound holds with probability at least $1-2^8 (\frac{\sigma_{*}}{\epsilon})^2 \exp\left(\frac{-\nrf\epsilon^2}{4d+8} \right)$
	\vspace*{-0.2cm}
	\begin{align}
		\hspace*{-0.2cm}
		&\sum_{\tau = 1}^T \ell^{\rm SW}_{\tau|\tau-1} \!\!-\!\! \sum_{\tau = 1}^T\! \mathcal{L}(\hat{f}^{i_\tau}(\mathbf{x}_\tau); y_\tau\!)  \!\leq \log \!M \!+\!  S \log T \!\!- \! S\log\! S \!+\! S\nonumber \\
		& + \epsilon LTC'\!+\!\!\sum_{m=1}^M\!\!\frac{(1+\epsilon){C'}^2}{2\sigma_{\theta^{m^*}}^2}\! +\!  \nrf M \log \left(1+ \frac{Tc \sigma_{\theta^{*}}^2 }{2\nrf M} \right)  \label{eq:Thm_2}
	\end{align}\\
	where $C'$ is some constant, and $\sigma_{*}^2 : =\max_{m\in\mathcal{M}} \sigma_{m}^2= \max_{m\in\mathcal{M}}\  \mathbb{E}_{\pi^{m^*}_{\bar{\kappa}}}[\| \mathbf{v}^{m^*}\|^2]$. 
	Setting $\epsilon = \mathcal{O}(\log T/T)$, the switching regret in \eqref{eq:regret_tr} boils down to 
	\begin{align}
		\mathcal{R}^{\rm SW}(T) = \mathcal{O}(\log T)\;. \label{eq:reg_trk}
	\end{align}}

\noindent {\bf Proof:}
See Sec.~\ref{sec:proof_Thm2}.\\

\noindent Even in the presence of model switching, the advocated SIE-GP suffers from diminishing average regret by explicitly accounting for such switching dynamics. 

\subsection{Dynamics within each GP expert}
The aforementioned SIE-GP accounts for dynamics of expert switching at the EGP meta-leaner. To further handle a dynamic learning function within each expert $m$, a time-varying parameter vector $\bbtheta^m_t$ will be considered instead of time-invariant $\bbtheta^m$ in IE-GP, yielding the dynamic (D) IE-GP approach. Specifically, DIE-GP captures dynamics in  $\bbtheta^m_t$ via the random walk model 
\begin{align}
	\bbtheta^m_{t+1} = \bbtheta^m_{t} + \bbepsilon_{t+1}^m \label{eq:theta_s} \vspace{-0.5cm}
\end{align}\vspace*{-0.5cm}\\
where the noise $\bbepsilon_{t+1}^m$ is white and Gaussian distributed with mean zero and covariance matrix $\sigma_{\epsilon^m}^2\mathbf{I}_{2D}$. 

Rather than updating $p(\bbtheta^m|\mathbf{y}_t, \mathbf{X}_t)$ as in IE-GP, expert $m$ in DIE-GP propagates $p(\bbtheta^m_t|\mathbf{y}_t, \mathbf{X}_t)$ across slots. Taking into account \eqref{eq:theta_s}, expert $m$ first predicts the pdf of $\bbtheta_{t+1}^m$ at the beginning of slot $t+1$ as
\begin{align}
	p(\!\bbtheta_{t+1}^m|\mathbf{y}_t, \mathbf{X}_{t+1}) =\int p(\bbtheta_{t+1}^m|\bbtheta_{t}^m)p(\bbtheta_{t}^m|\mathbf{y}_t, \mathbf{X}_{t}) d\bbtheta_t^m \!\!\label{eq:p_theta_pre}
\end{align}\vspace*{-0.4cm}\\
which replaces $p(\bbtheta^m|\mathbf{y}_t,\mathbf{X}_t)$ in \eqref{eq:p_y_pre_1} and \eqref{eq:p_theta_up_1} to obtain the predictive pdf $p(y_{t+1}|\mathbf{y}_t, \mathbf{X}_{t+1})$, and the posterior $p(\bbtheta^m_{t+1}|\mathbf{y}_{t+1},\mathbf{X}_{t+1})$ in the dynamic setting. Specifically for GPR with per-expert Gaussian posterior $p(\bbtheta^m_t|\mathbf{y}_t,\mathbf{X}_{t}) = \mathcal{N}(\bbtheta_t^m; \hat{\bbtheta}^m_t, \!\bbSig^m_{t})$ , the predictive pdf in \eqref{eq:p_theta_pre} is $p(\bbtheta_{t+1}^m|\mathbf{y}_t,\mathbf{X}_{t+1}) = \mathcal{N}(\bbtheta_{t+1}^m; \!\hat{\bbtheta}_{t}^m,  \bbSig_t^m\!+\!\sigma_{\epsilon^m}^2\mathbf{I}_{2D})$. 

\noindent 
{\bf Remark 1}.  The dynamics in EGP meta-learner and GP learners can be readily combined to yield the DSIE-GP generalization. See Fig.~\ref{fig:EGP_diagrams} for the schematic diagrams of (D,S)IE-GP.


\myspa
\myspace
\begin{algorithm}[t]
	\caption{IE-GPLVM}\label{Alg:o_rf_gplvm}
	\begin{algorithmic}[1]
	\begin{spacing}{1}
	\vspace*{1.7mm}
		\State\textbf{Initialization:} 
		\For{$m = 1, \ldots, M$}
	\State	Draw  vectors $\{\mathbf{v}_j^m\}_{j = 1}^{\nrf} \sim  \pi_{\bar{\kappa}^m}(\mathbf{v})$; %
	\State Embed $\mathbf{Y}_{t_0}\!\! \rightarrow \hat{\mathbf{X}}_{t_0}^m$ and obtain hyperparameters (cf. \eqref{eq:init_opt});
	\State {{$\mathbf{B}^m_{t_0}={\hat{\bbPhi}_{t_0}}^{m\top} \mathbf{Y}_{t_0}$} with {${\hat{\bbPhi}_{t_0}}^{m}=[\bbphi^m_{\mathbf{v}}(\hat{\mathbf{x}}_{1}^m) \ldots \bbphi_{\mathbf{v}}(\hat{\mathbf{x}}_{t_0}^m)]^{\top}$}};
	\State {{$\mathbf{R}_{t_0}^m=$} \verb|CholeskyFactor|$({\hat{\bbPhi}_{t_0}}^{m\top}{\hat{\bbPhi}_{t_0}}^{m} + \noisevar \mathbf{I}_{2\nrf})$};
	\EndFor
	\newline
		\For{$t = t_0+1, t_0+2, \ldots$}
		\State Receive datum $\mathbf{y}_{\ysub{t}}$;
		\For{$m = 1, \ldots, M$}
		\State Obtain embedding $\hat{\mathbf{x}}_t^m$ based on \eqref{eq:test_point_prb}

		\State \textcolor{black}{ $\mathbf{B}_t^m=\mathbf{B}_{t-1}^m+\bbphi^m_{\mathbf{v}}({\hat{\mathbf{x}}_t^m})\mathbf{y}^{\top}_{\ysub{t}}$}
		\State $\mathbf{R}_t^m=\verb|CholeskyUpdate| (\mathbf{R}_{t-1}^m,\bbphi^m_{\mathbf{v}}({\hat{\mathbf{x}}_t^m}))$
		\EndFor
		\State Obtain $\hat{\mathbf{x}}_t = \hat{\mathbf{x}}_t^{m*}$ based on~\eqref{eq:m_star};
	\State Update $w_{t+1}^m$ based on~\eqref{eq:w_up_LVM};
		\EndFor
	\end{spacing}
\end{algorithmic}
\end{algorithm}

\section{EGPs for online unsupervised learning}
Rather than supervised learning, this section deals with EGP-based latent variable model (LVM) for unsupervised dimensionality reduction. Consider first the GPLVM context, where the $D\times 1$ observation $\mathbf{y}_{\ysub{\tau}}:=[y_{\tau,1}\ldots y_{\tau,D}]^\top$\footnote{Notice the introduction of the colon in the subscript, indicating a multivariate observation at time $\tau$; not to be confused with $\mb{y}_\tau$ in previous sections. } is linked with the {\it unobserved} low-dimensional input $\mathbf{x}_\tau \in \mathbb{R}^d$ ($d<D$) via~\cite{Lawrence05}
\begin{align}
y_{\tau j}=f_j(\mb{x}_\tau)+n_{\tau j},\quad  j = 1,\ldots, D \label{eq:main}
\end{align}
where $f_j(\cdot)\sim \mathcal{GP}(0,\kappa)$, and $\{n_{tj}\}$  are assumed to be drawn i.i.d. from $\mc{N}(0,\noisevar)$.

Given $t$ observations $\mb{Y}_t:=[\mb{y}_{\ysub{1}} \ldots \mb{y}_{\ysub{t}}]^{\top}\equiv[\mb{y}_{t}^{(1)}\ldots \mb{y}_{t}^{(D)}]$, where $\mb{y}_{t}^{(j)}:=[y_{1,j}\ldots y_{t,j}]^\top$, the estimate of the low-dimensional embedding $\mb{X}_t:=[\mb{x}_1 \ldots \mb{x}_t]^{\top}$ is sought together with the kernel hyperparameters $\boldsymbol{\alpha}$ by solving~\cite{Lawrence05}
\begin{align}
(\hat{\mb{X}}_t, \hat{\boldsymbol{\alpha}} )=\underset{\mb{X}_t,\boldsymbol{\alpha}}{\arg\max}\ \log p(\mb{Y}_t |\mb{X}_t;\boldsymbol{\alpha}) + \log p(\mathbf{X}_t) \label{eq:likelihood_gplvm}
\end{align}
where $p(\mb{Y}_t |\mb{X}_t; \boldsymbol{\alpha}) = \prod_{j=1}^{D} \mc{N}(\mb{y}_{t}^{(j)};\mb{0},\mb{K}_t+\noisevar\mb{I}_t)$ is the so-termed marginal likelihood (ML), and a standard choice for the prior of $\mb{X}_t$ is $p(\mb{X}_t)=\prod_{\tau=1}^{t}\mc{N}(\mb{x}_\tau;\mb{0},\sigma_x^2\mb{I}_d)$.

The routine to solve \eqref{eq:likelihood_gplvm} entails inverting the $t\times t$ kernel matrix, thus incurring unaffordable complexity as in the supervised setting~\cite{Lawrence05}. 
To effect scalablility via the RF approximation and accommodate online kernel adaptation, an IE-GP based LVM is well motivated in accordance with the preceding discussion. 

\subsection{IE-GPLVM}
In the novel IE-GPLVM, the EGP meta-learner employs an ensmeble of GP experts to independently seek low-dimensional embeddings of the observations. As before, each expert $m$ will leverage the $\kappa^m$-induced RF mapping $\bbphi_{\mb{v}}^{\esym}(\cdot)$~\eqref{eq:phi_x} to yield the per-datum conditional likelihood 
\begin{align}
p(\mathbf{y}_{\ysub{\tau}}|\bbTheta^m,\mb{x}_\tau) &=  \prod_{j=1}^{D} \mc{N}(y_{\tau j};\boldsymbol{\phi}_{\mathbf{v}}^{m\top}(\mb{x}_\tau)\bbtheta_j^{m},\noisevar)\label{eq:IEGPLVM_LF}
\end{align}
where the RF-based parameter vectors over $D$ output channels are collected in $\bbTheta^m:=[\bbtheta_1^m, \ldots, \bbtheta_D^m]^{\top}$, whose prior pdf is given by
\begin{align}
p(\bbTheta^m)&=\prod_{j=1}^{D}\mc{N}(\bbtheta_j^m;\mb{0},\sigma_{\theta^m}^2\mb{I}_{2\nrf})\;.\label{eq:IEGPLVM_prior}
\end{align}
It is worth mentioning that such an RF-based GPLVM can be regarded as the nonlinear dual form of probabilistic PCA~\cite{Lawrence05}.

To incrementally project subsequent observations to the low-dimensional embeddings,
each expert $m$ in IE-GPLVM relies on the generative model \eqref{eq:IEGPLVM_LF}--\eqref{eq:IEGPLVM_prior}
to summarize past outputs $\mathbf{Y}_t$ and the estimated inputs $\hat{\mathbf{X}}_t^m$
in the posterior pdf
\begin{align}
p(\bbTheta^m|\mathbf{Y}_t,\hat{\mathbf{X}}_t^m) = \prod_{j=1}^D \mathcal{N}(\bbtheta^m_j; \hat{\bbtheta}^m_{t,j}, \bbSig_{t}^m)\label{eq:lvm_post}
\end{align}
where the parameter vectors associated with different output channels share the same covariance matrix $\bbSig_{t}^m$. Further, the per-expert weight assessed by the EGP meta-learner in IE-GPLVM is adapted with estimated inputs as
$w_t^m:={\rm Pr}(i=m|\mb{Y}_{t},\{\hat{\mb{X}}_{t}^{\nu}\}_{\nu=1}^{\Esym}))$.
In the same spirit as IE-GP, each iteration  in IE-GPLVM alternates between estimation of $\mb{x}_{\tau}$ from $\mb{y}_{\ysub{\tau}}$, and correction of the per-expert weight $w_t^m$ and posterior pdf~\eqref{eq:lvm_post}. Note that instead of the moments of~\eqref{eq:lvm_post}, matrices $\mathbf{A}_t^m:=(\bbSig_{t}^m)^{-1}$ and $\mathbf{B}_t^m:=(\bbSig_{t}^m)^{-1}\hat{\bbTheta}_{t}^m$ will be equivalently updated with Chelosky decomposition performed via $\mathbf{A}_t^m=\mathbf{R}_t^{m\top}\mathbf{R}_t^{m}$ for numerical stability (cf. Alg.~2). However, for consistency with previous sections, the following discussion still uses the moments $\hat{\bbTheta}_{t}^m$ and $\bbSig_{t}^m$.\\

\noindent {\bf Estimation.}
To estimate $\mb{x}_{t+1}$ based on $\mb{y}_{\ysub{t+1}}$, expert $m$ capitalizes on the parameter posterior~\eqref{eq:lvm_post} to obtain the conditional likelihood (similar to \eqref{eq:p_y_pre_GPR}) as
 $p(\mb{y}_{\ysub{t+1}}|\mb{Y}_t,i\!=\! m, \hat{\mathbf{X}}_t^m,\mb{x}_{t+1})=\mc{N}(\mb{y}_{\ysub{t+1}};\hat{\mathbf{y}}_{t+1:}^{m}(\mb{x}_{t+1}),(\sigma_{t+1}^m(\mb{x}_{t+1}))^2 \mb{I}_{D}) $, 
 where the first two moments are functions of $\mathbf{x}_{t+1}$ as
\begin{subequations} \label{eq:subeqns}
\begin{align}
 \hat{\mathbf{y}}_{t+1:}^m(\mb{x}_{t+1}) &= \hat{\bbTheta}_{t}^m \bbphi^m_{\mathbf{v}} (\mathbf{x}_{t+1}) \label{eq:gplvm_test_mean}    \\
(\sigma_{t+1}^m(\mb{x}_{t+1}))^2 &=  \bbphi^{m\top}_{\mathbf{v}}  (\mathbf{x}_{t+1})\bbSig_{t}^m\bbphi^{m}_{\mathbf{v}}  (\mathbf{x}_{t+1})+\sigma_n^2\;. \label{eq:gplvm_test_var}
\end{align}
\end{subequations}
Further imposing a prior on $\mb{x}_{t+1}$, the maximum-a-posteriori (MAP) estimate of $\mb{x}_{t+1}$ is given by
\begin{align}
\hat{\mb{x}}_{t+1}^m\!=\!
\underset{\mb{x}_{t+1}}{\arg\max}\  \log p(\mb{y}_{\ysub{t+1}}|\mb{Y}_t,\! i\!=\! m, \hat{\mathbf{X}}_t^m\!\!,\mb{x}_{t+1})\!+\!\log p(\mb{x}_{t+1}\!). \label{eq:test_point_prb}
\end{align}
With the per-expert embeddings $\{\hat{\mb{x}}_{t+1}^m\}_{\esym=1}^{\Esym}$ at hand, the EGP meta-learner seeks the final estimate as $\hat{\mb{x}}_{t+1}=\hat{\mb{x}}_{t+1}^{m^*}$, where
\begin{align}
\hspace*{-0.2cm}
m^*\! =\! \underset{\esym\in {\cal M}}{\arg\max}\ w_t^m  {p}(\mb{y}_{\ysub{t+1}}|\mathbf{Y}_{t},\!i\!=\!m,\hat{\mb{X}}_{t}^m\!\!, \hat{\mb{x}}_{t+1}^m) \; p(\hat{\mb{x}}_{t+1}^m\!)\;. \label{eq:m_star}
\end{align}
It can be readily verified that $(m^*, \hat{\mb{x}}_{t+1}^{m^*})$ corresponds to the MAP solution of $p(\mathbf{x}_{t+1},i=\esym|\mathbf{y}_{\ysub{t+1}},\mathbf{Y}_{t},\{\hat{\mb{X}}_{t}^{\nu}\}_{\nu=1}^{\Esym}) $.\footnote{\vspace{-1mm} Observe that \begin{align}
&\max_{\mathbf{x}_{t+1},\esym} \; p(\mathbf{x}_{t+1},i=\esym|\mathbf{y}_{\ysub{t+1}},\mathbf{Y}_{t},\{\hat{\mb{X}}_{t}^{\nu}\}_{\nu=1}^{\Esym})  \notag \\
&\equiv \max_{\mathbf{x}_{t+1},\esym} p(\mathbf{y}_{\ysub{t+1}},\mathbf{x}_{t+1},i\!=\!\esym|\mathbf{Y}_{t}, \{\hat{\mb{X}}_{t}^{\nu}\}_{\nu=1}^{\Esym}) \notag \\ &\equiv \max_{\mathbf{x}_{t+1},\esym}  p(i\!=\!\esym|\mathbf{Y}_{t},\{\hat{\mb{X}}_{t}^{\nu}\}_{\nu=1}^{\Esym}) {p}(\mathbf{y}_{\ysub{t+1}}|\mathbf{Y}_{t},i\!=\!\esym,\hat{\mb{X}}_{t}^{m}\!\!,\mathbf{x}_{t+1}) p(\mathbf{x}_{t+1}) \notag
\end{align}}\\ %

\noindent {\bf Correction.} Upon obtaining the estimate $\hat{\mb{x}}_{t+1}^m$, the EGP meta-learner updates the per-expert weight
\begin{align}
w_{t+1}^m&:={\rm Pr}(i=m|\mb{Y}_{t+1}, \{\hat{\mathbf{X}}_{t+1}^\nu\}_{\nu=1}^{\Esym})\notag\\ 
&\ \propto w_{t}^m\;{p}(\mb{y}_{\ysub{t+1}}|\mathbf{Y}_{t},i=\esym,\hat{\mathbf{X}}_t^m,\hat{\mathbf{x}}_{t+1}^m)  \label{eq:w_up_LVM}
\end{align}
which is quite intuitive in that it favors experts which assign higher likelihood to the observation $\mb{y}_{\ysub{t+1}}$. In the long run, we expect the probability mass to concentrate at the expert(s) whose embeddings best describe the observed data. IE-GPLVM is thus effectively performing online kernel selection, adapting to the data as they become available.

Meanwhile, the pair $\{\hat{\mb{x}}_{t+1}^m, \mb{y}_{\ysub{t+1}}\}$ allows expert $m$ to update the posterior pdf of $\bbTheta^m$ as
$p(\bbTheta^m|\mathbf{Y}_{t+1}, \hat{\mathbf{X}}_{t+1}^m) = \prod_{j=1}^D \mathcal{N}(\bbtheta^m_j; \hat{\bbtheta}^m_{t+1,j}, \bbSig_{t+1}^m)$.
As previously mentioned, the update of the moments is implemented by propagation of $\mathbf{B}_t^m$ and $\mathbf{R}_t^m$ as shown in
Alg.~2, where \texttt{CholeskyFactor} computes the Cholesky factor (CF) of its argument and \texttt{CholeskyUpdate} performs a rank-one CF update.\\

\noindent A few remarks are in order.

\noindent {\bf Remark 2.}
To obtain the kernel hyperparameters, expert $m$ will leverage the first $t_0$ samples $\mathbf{Y}_{t_0}$ to solve the optimisation problem
\begin{align}
(\hat{\mb{X}}_{t_0}^m, \hat{\boldsymbol{\alpha}}^m )=\underset{\mb{X}_t,\boldsymbol{\alpha}}{\arg\max}\ \log p(\mb{Y}_{t_0} |\mb{X}_{t_0},i=m;\boldsymbol{\alpha}) + \log p(\mathbf{X}_{t_0}) \label{eq:init_opt}
\end{align}
where the RF-based likelihood for expert $m$ is 
$p(\mb{Y}_{t_0}|i\!=\!m, \mb{X}_{t_0};\boldsymbol{\alpha}) = \prod_{j=1}^{D} \mc{N}(\mb{y}_{t_0}^{(j)};\mb{0},\sigma_{\theta^m}^2\mb{\Phi}_{t_0}^m\mb{\Phi}^{m\top}_{t_0}+\noisevar\mb{I}_{t_0})$
with $\mb{\Phi}_{t_0}^m:=[\bbphi_{\mb{v}}^{m}(\mb{x}_1) \ldots \bbphi_{\mb{v}}^{m}(\mb{x}_{t_0})]^{\top}$. With $\hat{\mb{X}}_{t_0}^m$ at hand, expert $m$ further relies on the generative model~\eqref{eq:IEGPLVM_LF}--\eqref{eq:IEGPLVM_prior} with inputs replaced by the estimates, to obtain $p(\bbTheta^m|\mathbf{Y}_{t_0}, \hat{\mathbf{X}}_{t_0}^m) = \prod_{j=1}^D \mathcal{N}(\bbtheta^m_j; \hat{\bbtheta}^m_{{t_0},j}, \bbSig_{{t_0}}^m)$, based on which the upcoming data will be processed incrementally (cf. Alg.~2).

\noindent {\bf Remark 3.} In practice, when solving \eqref{eq:test_point_prb}, $\mb{x}_{t+1}$ is initialized at the embedding corresponding to the point in $\mb{Y}_t$ that is the nearest neighbor (NN) of $\mb{y}_{t+1:}$~\cite{yao2011learning}. Note, however, that since the per incoming observation complexity for obtaining the NN, at slot $t$ scales linearly with $t$, it can significantly surpass the (constant wrt $t$) likelihood evaluation complexity of $\mc{O}(\nrf^2)$. To obtain a scalable algorithm, we will rely on an approximate kNN search scheme that relies on a hierarchical graph construct~\cite{malkov2018efficient}. {This approach can be shown to (approximately, at the limit) achieve} $\mc{O}(\log t)$ complexity; see~\cite{malkov2018efficient} for a detailed description.

\noindent {\bf Remark 4.} Dynamic variants of IE-GPLVM can also been pursued along the lines of Sec.~\ref{sec:dynamic}.

\section{Numerical tests}
To assess performance, real-data tests are presented here for regression, classification as well as dimensionality reduction tasks. The supplementary file also contains synthetic tests that validate the regret bounds in Theorems 1 and 2.
 \begin{figure*}[t]
	\minipage[b]{0.32\textwidth}
	\centering
	\includegraphics[width=\linewidth]{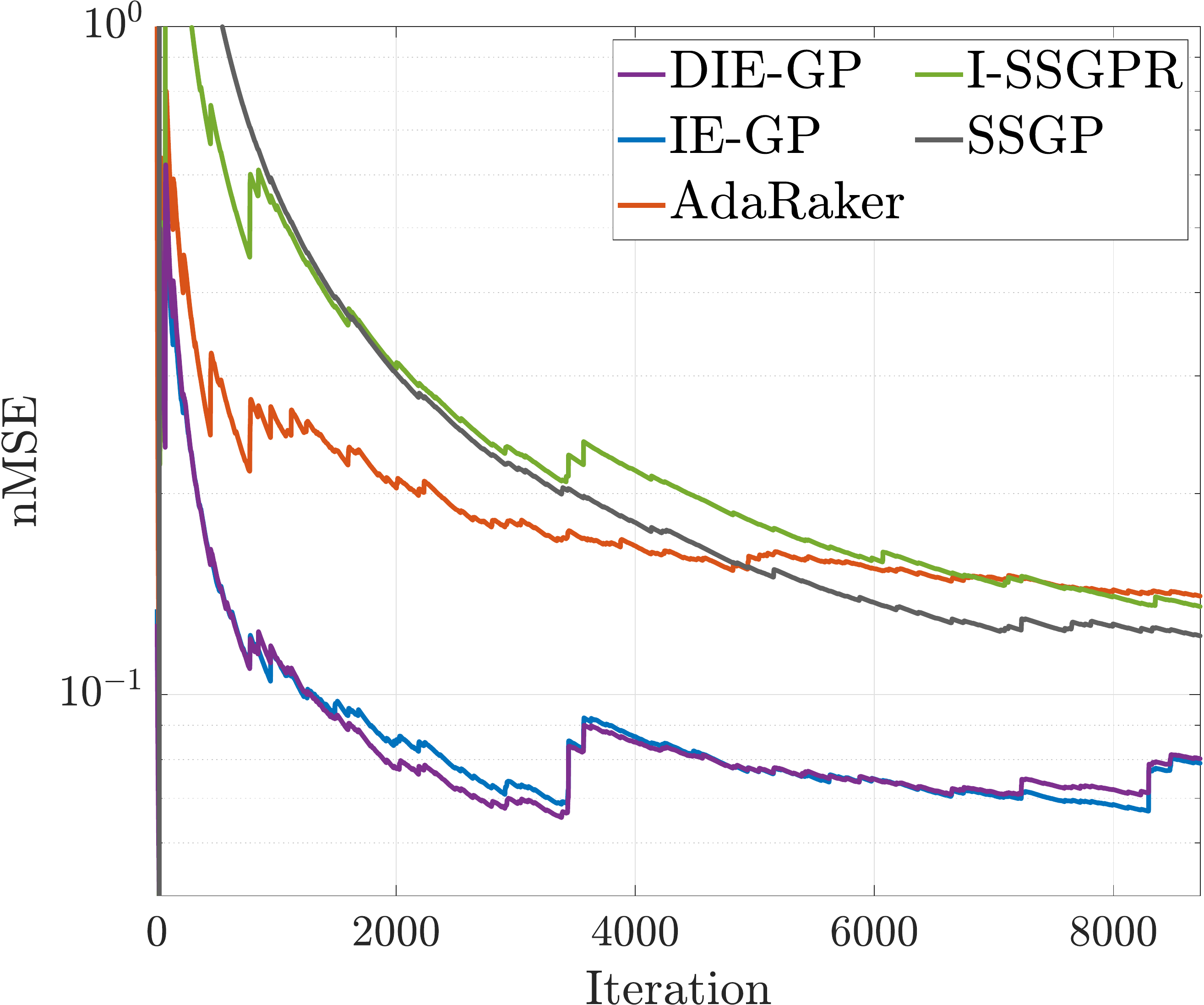}
	{(a)}\label{fig:tom}
	\endminipage\qquad
	\hspace{-4mm}
	\minipage[b]{0.32\textwidth}
	\centering
	\includegraphics[width=0.987\linewidth]{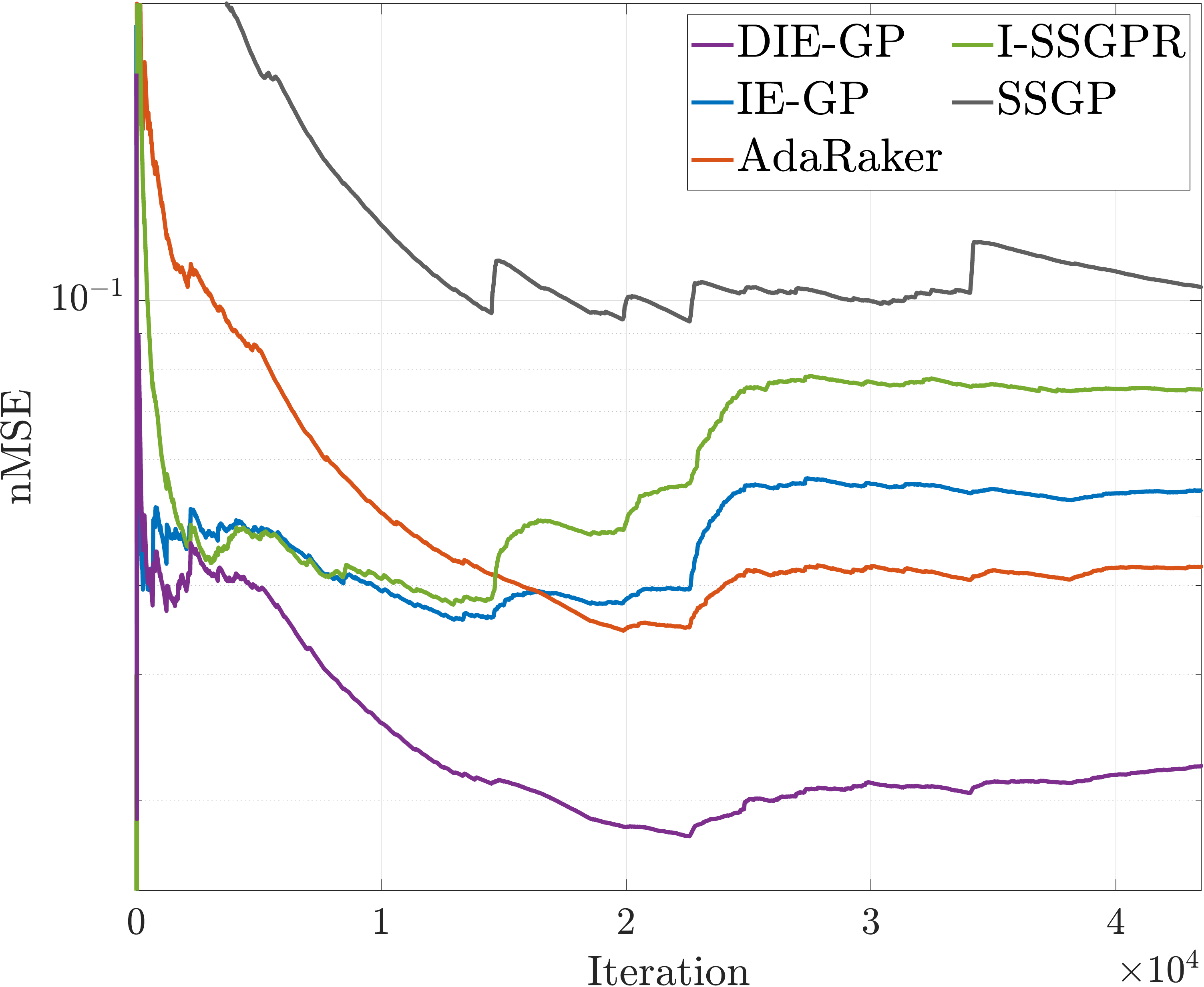}
	{(b)}\label{fig:sarcos}
	\endminipage\qquad
	\hspace{-4mm}
	\minipage[b]{0.32\textwidth}
	\centering
	\includegraphics[width=0.997\linewidth]{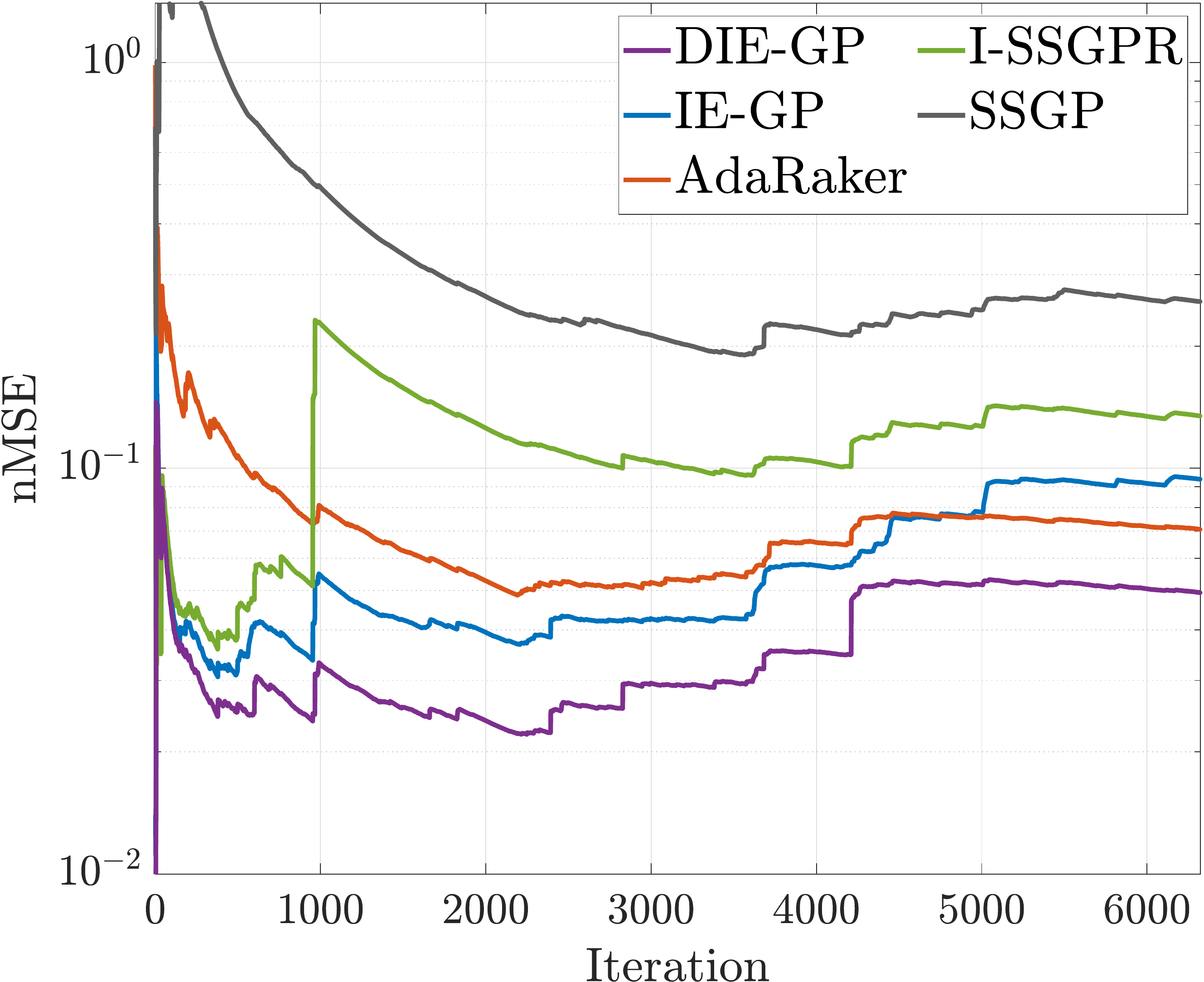}
	{(c)}\label{fig:air}
	\endminipage\qquad
	\caption{Log scale nMSE plots on (a) \texttt{Tom's hardware}; (b) \texttt{SARCOS}; and, (c) \texttt{Air quality} datasets.} \label{fig:common}
	\vspace*{-0.3cm}
\end{figure*}

\begin{table}[h]
	\caption{Statistics of the datasets} \label{table:stat}
	\begin{center}
		\begin{tabular}{c|c|c|c|c}
			\hline
			\hline
		Task & 	\textbf{Datasets}  &  $T$ &  $d$  & $D$ \\
			\hline
			\hline 
		Regression &	Tom's hardware &  9725 & 96 & 1  \\
			\hline 		
		 Regression &	SARCOS~\cite{williams2006gaussian}           & 44484 & 21 & 1   \\
			\hline 
		Regression &	Air Quality~\cite{de2008field}   &  7322 & 12 & 1  \\
			\hline
	     Classification &   Banana~\cite{Rae98}  & 5300 & 2& 1 \\
	       	\hline
	        Classification &   Musk~\cite{dietterich1994comparison}  & 6598  & 166 &1 \\
	        \hline
	       Classification &   Ionosphere~\cite{sigillito1989classification} & 351 & 34 &1 \\
	        \hline
	        GPLVMs & USPS~\cite{hull1994database}  &  5474 & NA &  256\\
	        \hline
	       GPLVMs & Oil flow~\cite{oilData}	 & 1000  & NA & 12\\
	        \hline
	       GPLVMs & MNIST	& 70000 &  NA  &784\\
	        \hline
		\end{tabular}
	\end{center}
\end{table}


\begin{figure*}[t]
	\minipage[b]{0.32\textwidth}
	\centering
	\includegraphics[width=0.98\linewidth]{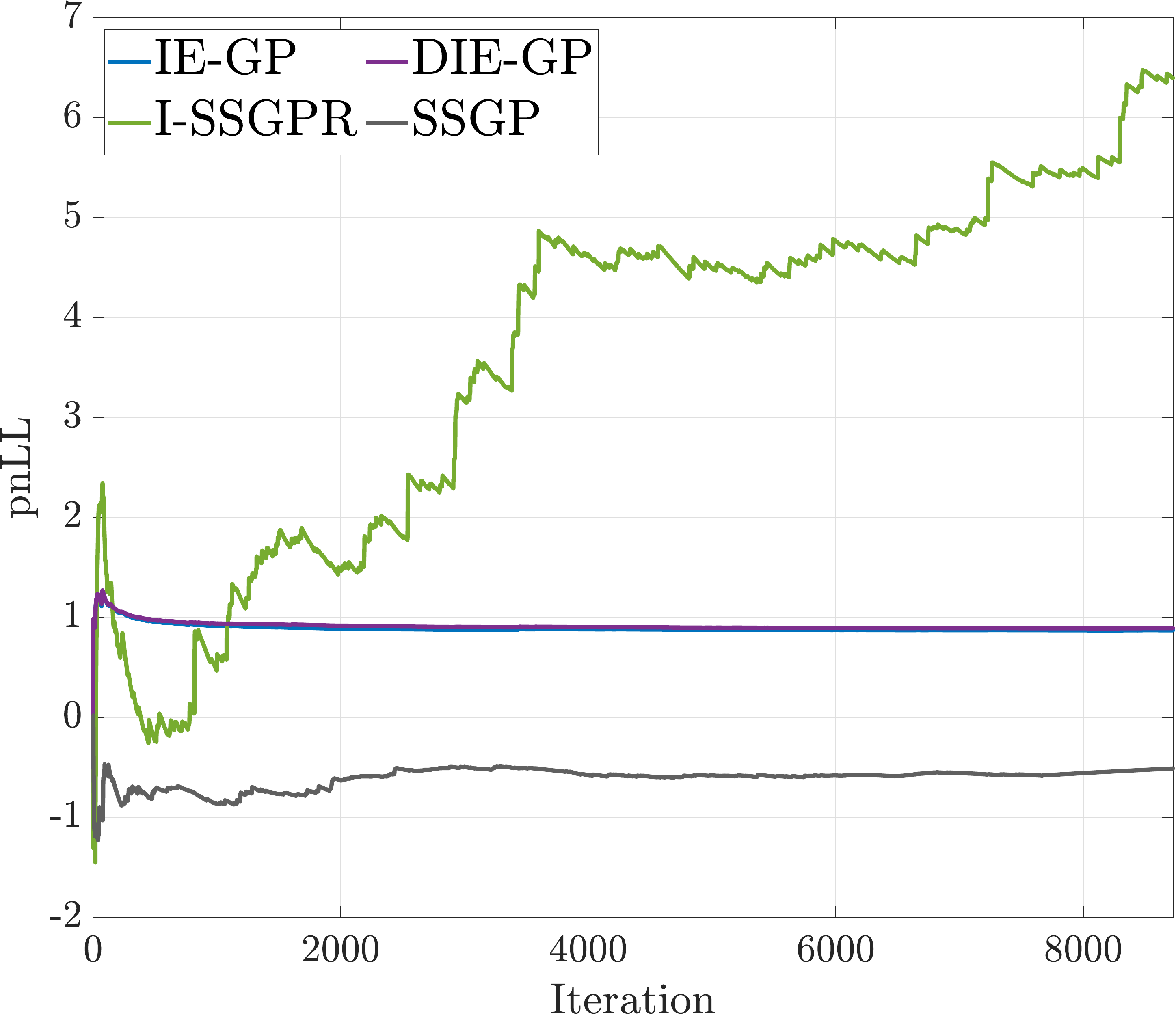}
	{(a)}\label{fig:tom_npll}
	\endminipage\qquad
	\hspace{-4mm}
	\minipage[b]{0.32\textwidth}
	\centering
	\includegraphics[width=\linewidth]{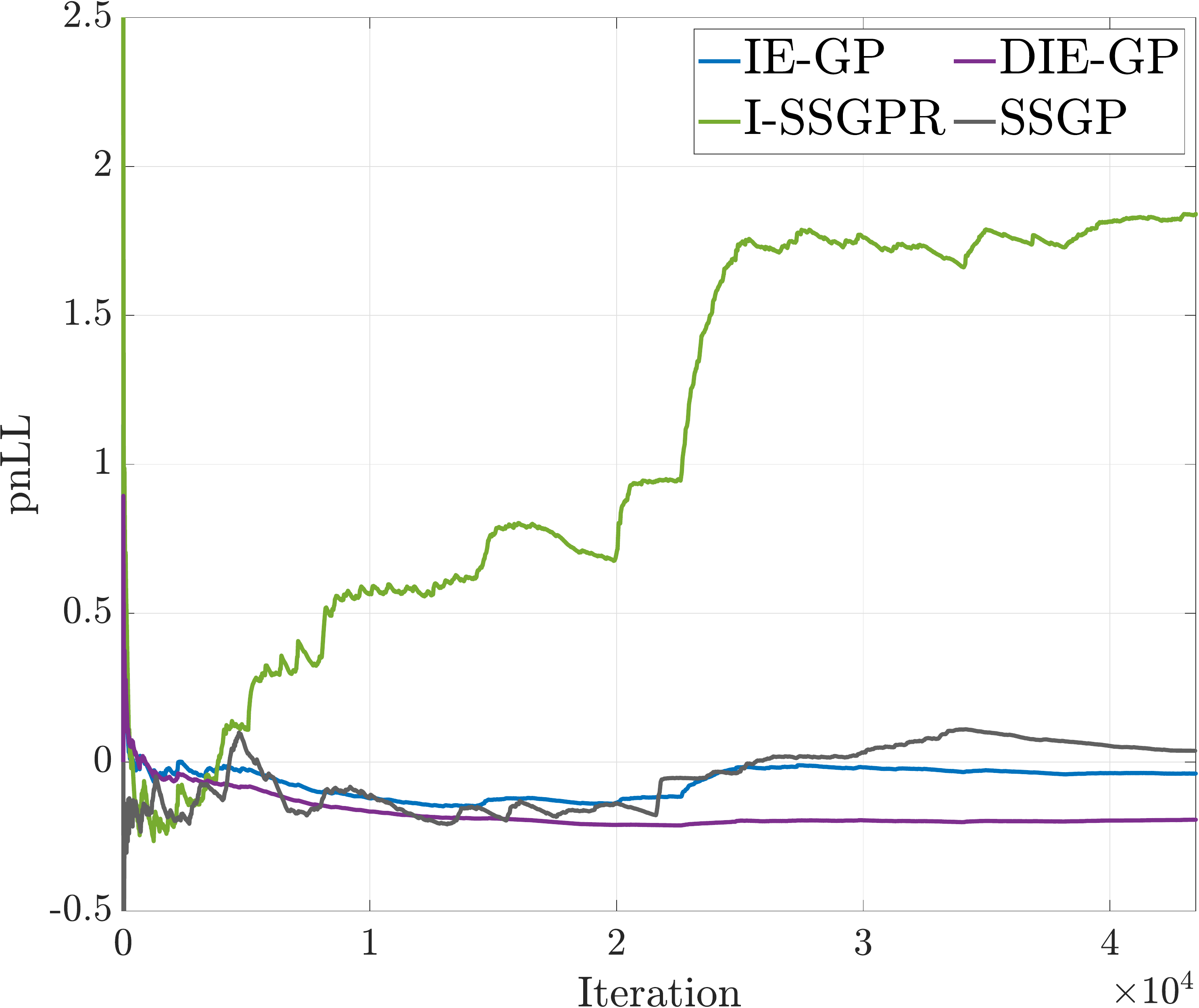}
	{(b)}\label{fig:sarcos_npll}
	\endminipage\qquad
	\hspace{-4mm}
	\minipage[b]{0.32\textwidth}
	\centering
	\includegraphics[width=\linewidth]{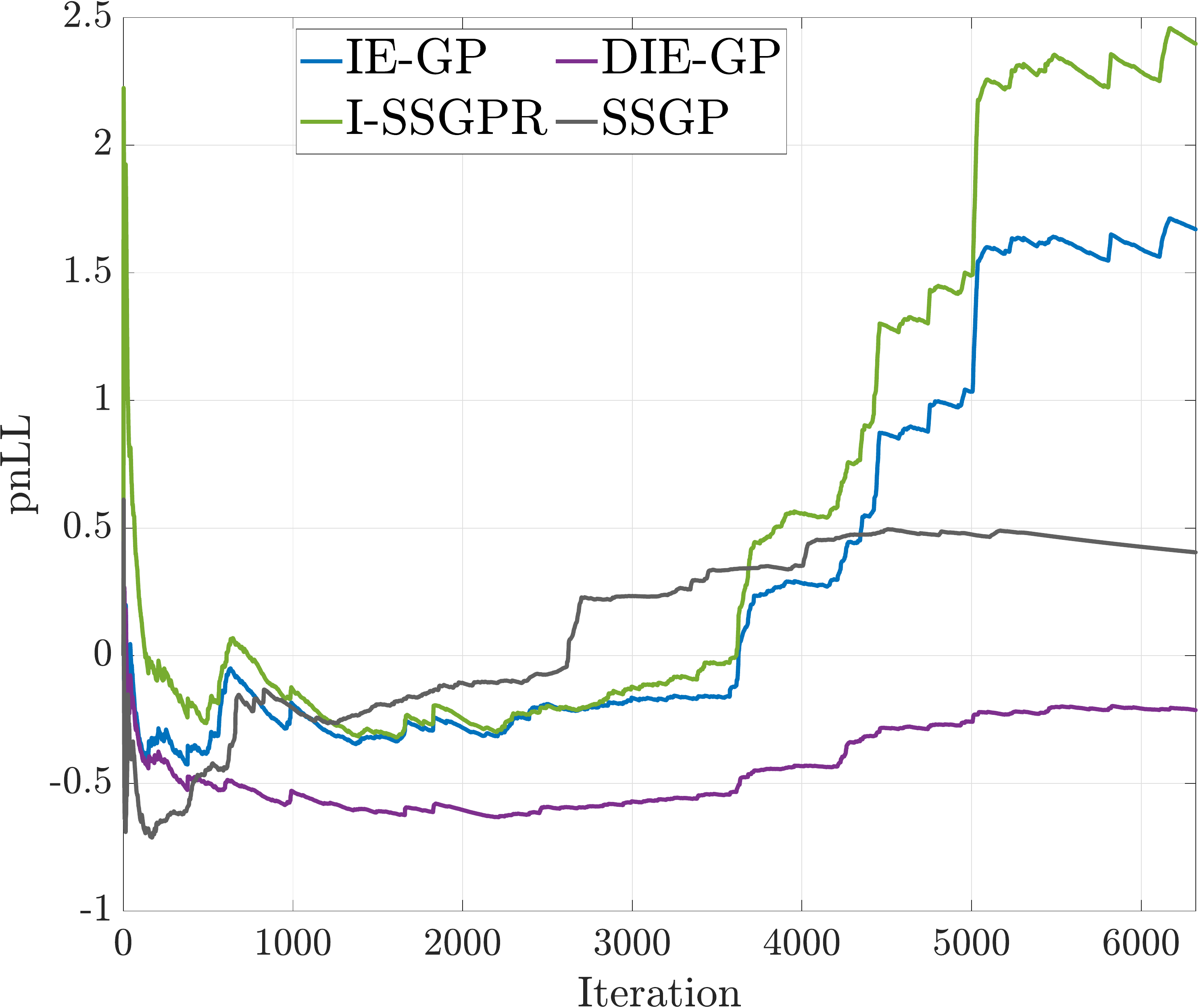}
	{(c)}\label{fig:air_npll}
	\endminipage
	\hspace*{-4cm}
	\caption{Predictive negative log-likelihood on  (a) \texttt{Tom's hardware}; (b) \texttt{SARCOS}; and, (c) \texttt{Air quality} datasets.} \label{fig:common_npll}
\end{figure*}

\vspace*{-0.1cm}
\subsection{Regression}
Regression tests were performed on the \texttt{SARCOS} dataset \cite{williams2006gaussian}, widely used for evaluating GP-based approaches, as well as on the \texttt{Air quality} \cite{de2008field}, \texttt{Tom's hardware} and \texttt{Twitter} datasets \cite{data_tom_and_twitter} from the UCI repository \cite{data_uci}. The statistics of the datasets are summarized in Table~\ref{table:stat}.
We compared the proposed (D)IE-GP approaches with AdaRaker \cite{shen2019random}, Incremental Sparse Spectrum Gaussian Process Regression (I-SSGPR)  \cite{gijsberts2013real}, and the Streaming Sparse Gaussian Process (SSGP) approach \cite{bui2017streaming}, in terms of normalized mean-square error (nMSE) and running time. SIE-GP is not included for comparison since the datasets exhibit no switching behavior among the candidate GP models. With $s_y^2$ denoting the sample variance of $\mathbf{y}_T$, the nMSE is defined as 
$\text{nMSE}_t:= t^{-1} \sum_{t'=1}^{t} (y_{t'} -\hat{y}_{t'|t'-1})^2/{s_y^2}$.

For all RF-based  approaches (namely (D)IE-GP, AdaRaker and I-SSGPR) we used $2\nrf=100$ and the reported results correspond to the run which resulted in the median nMSE among $101$ runs for the corresponding method. Finally, all reported runtimes include hyperparameter learning/model initialization computations performed on the first $1,000$ samples. 
If for some expert $m$ and time instance $t$ we have that $w_t^m=0$, it follows that $w_{t'}^m=0$ for all $t'>t$ (cf.~\eqref{eq:w_update_1}). Experts with $w_t^m<10^{-16}$ were deemed inactive for $t'> t$; thus, we set $w_{t'}^m=0$ for $t'> t$, and avoided unnecessary prediction/correction steps.

\begin{figure*}[t]
	\hspace{-2mm}
	\minipage[b]{0.32\textwidth}%
	\centering
	\hspace{-2mm}
	\includegraphics[width=0.85\linewidth,height =1.05\linewidth ]{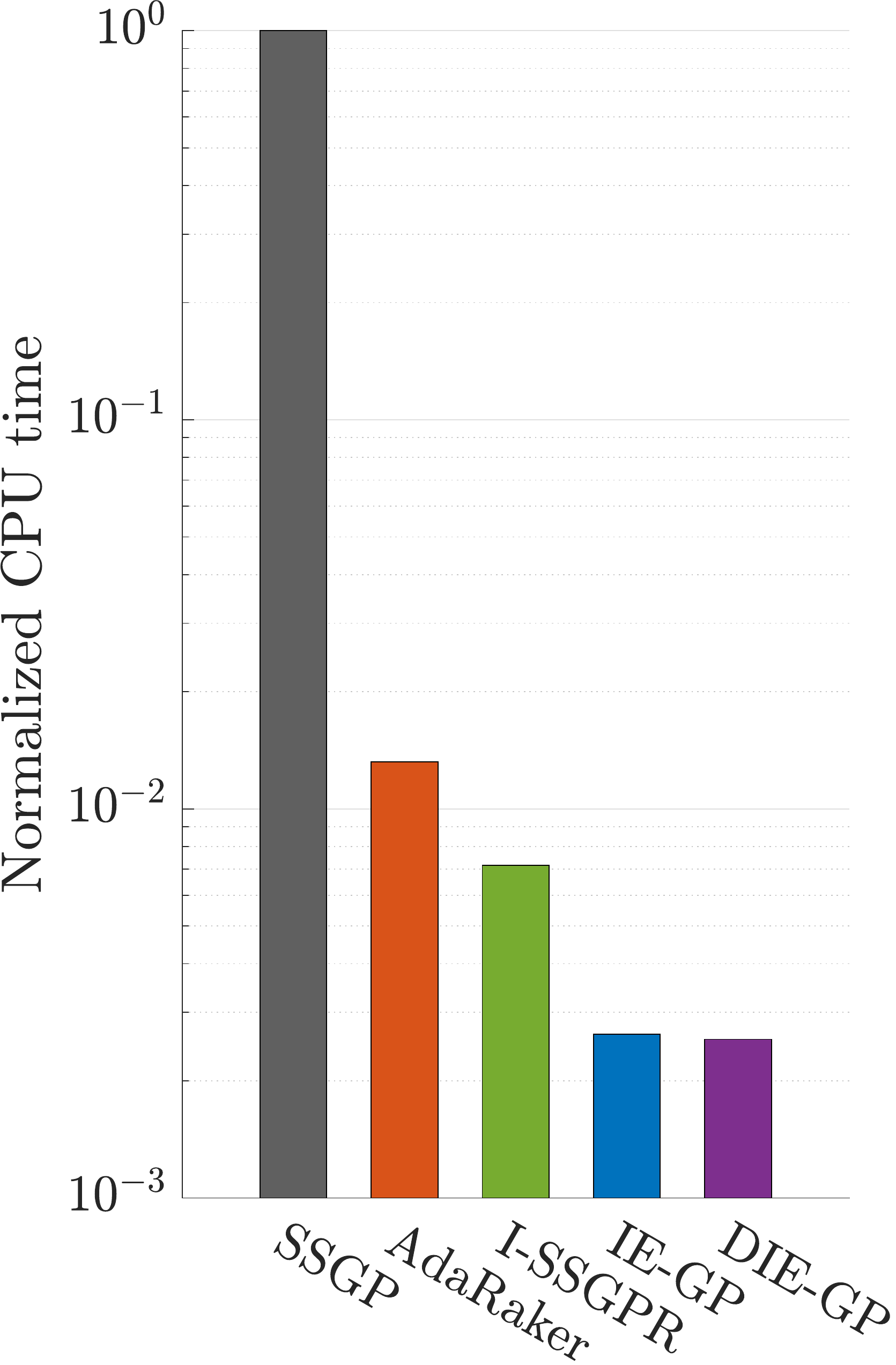}\\
	{(a)}\label{fig:tom_run}
	\endminipage
	\minipage[b]{0.32\textwidth}%
	\centering
	\hspace{-2mm}
	\includegraphics[width=0.8\linewidth,height =1.05\linewidth]{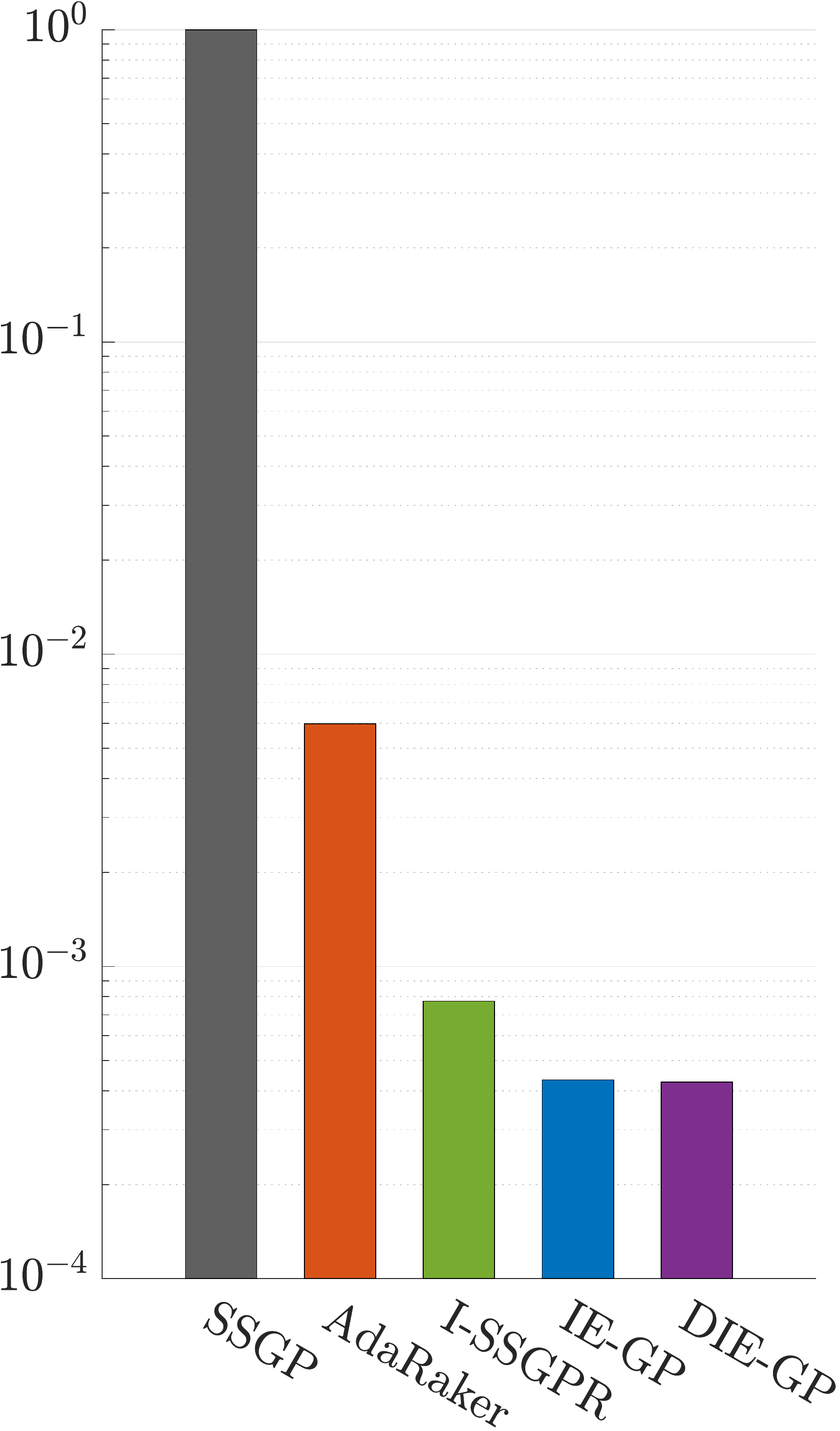}\\
	{(b)}\label{fig:sarcos_run}
	\endminipage
	\minipage[b]{0.32\textwidth}%
	\centering
	\hspace{-2mm}
	\includegraphics[width=0.8\linewidth,height =1.05\linewidth]{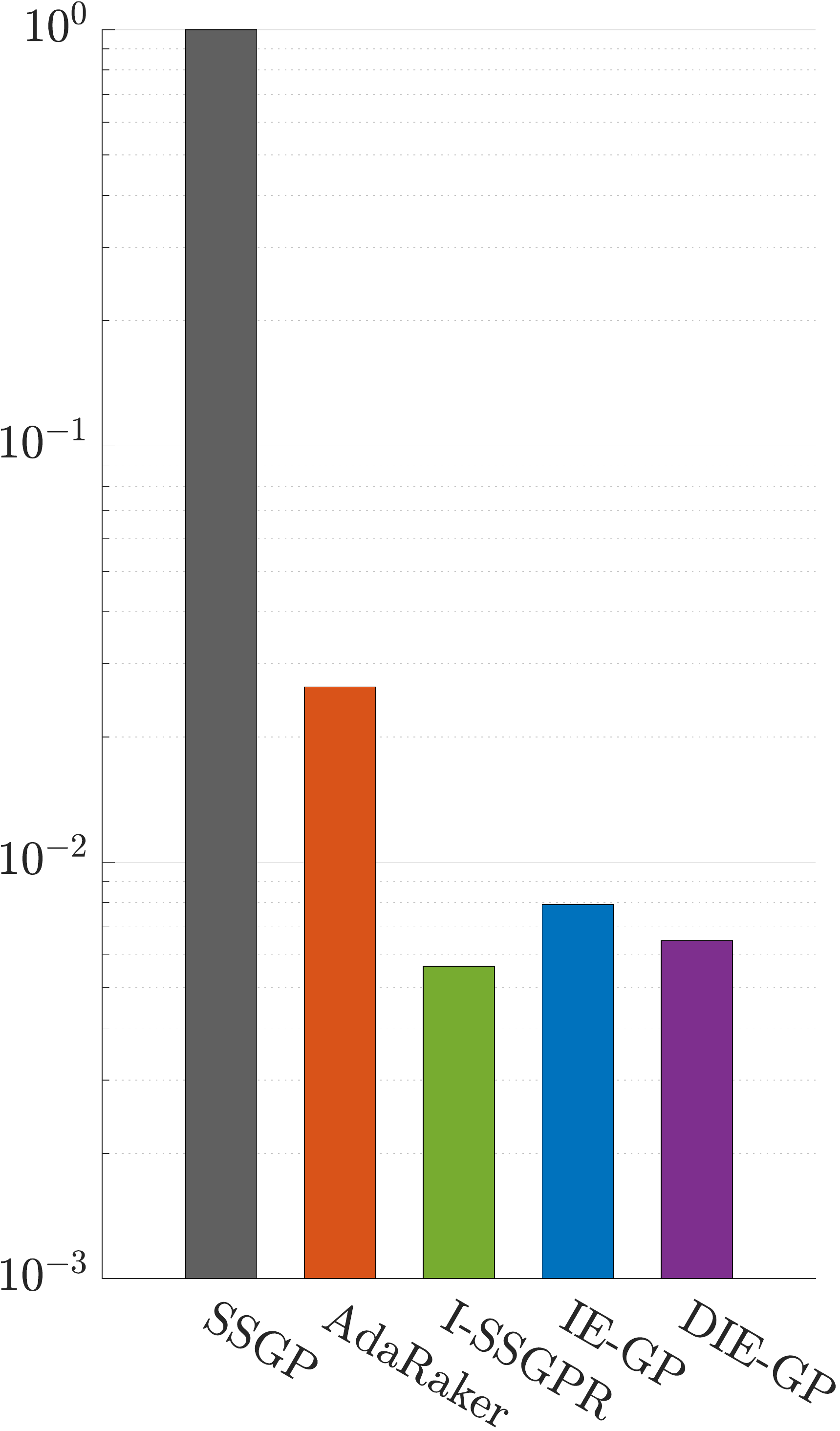}
	\\{(c)}\label{fig:air_run}
	\endminipage
	\vspace{-0.1cm}
	\caption{Normalized running times for regression on  (a) \texttt{Tom's hardware}; (b) \texttt{SARCOS}; and, (c) \texttt{Air quality} datasets.} \label{fig:run_all}
\end{figure*}

The kernel dictionary for (D)IE-GP and AdaRaker comprised radial basis functions (RBFs) with variances from the set $\{10^k\}_{k=-4}^6$. The automatic relevance determination (ARD) kernel was used for I-SSGPR, as in~\cite{gijsberts2013real}. The per kernel noise and prior variances (as well as ARD length scales for I-SSGPR),  were estimated by maximizing the marginal likelihood of the first $1,000$ samples using the \verb|minimize| function from the GPML toolbox \cite{GPML}. The aforementioned samples were not used in the deployment phase. 
In DIE-GP, $\sigma_{\epsilon^m}^2=0.001$  was used for all $m$ and in all experiments. Regarding SSGP, the ARD kernel was used, the batch size was set to $300$, the number of inducing points was $100$ and the first $1,000$ samples were used for obtaining an initial model, all as per the original work \cite{bui2017streaming}. 

\begin{figure*}[t]
	\minipage[b]{0.32\textwidth}
	\centering
	\includegraphics[width=1.1\linewidth]{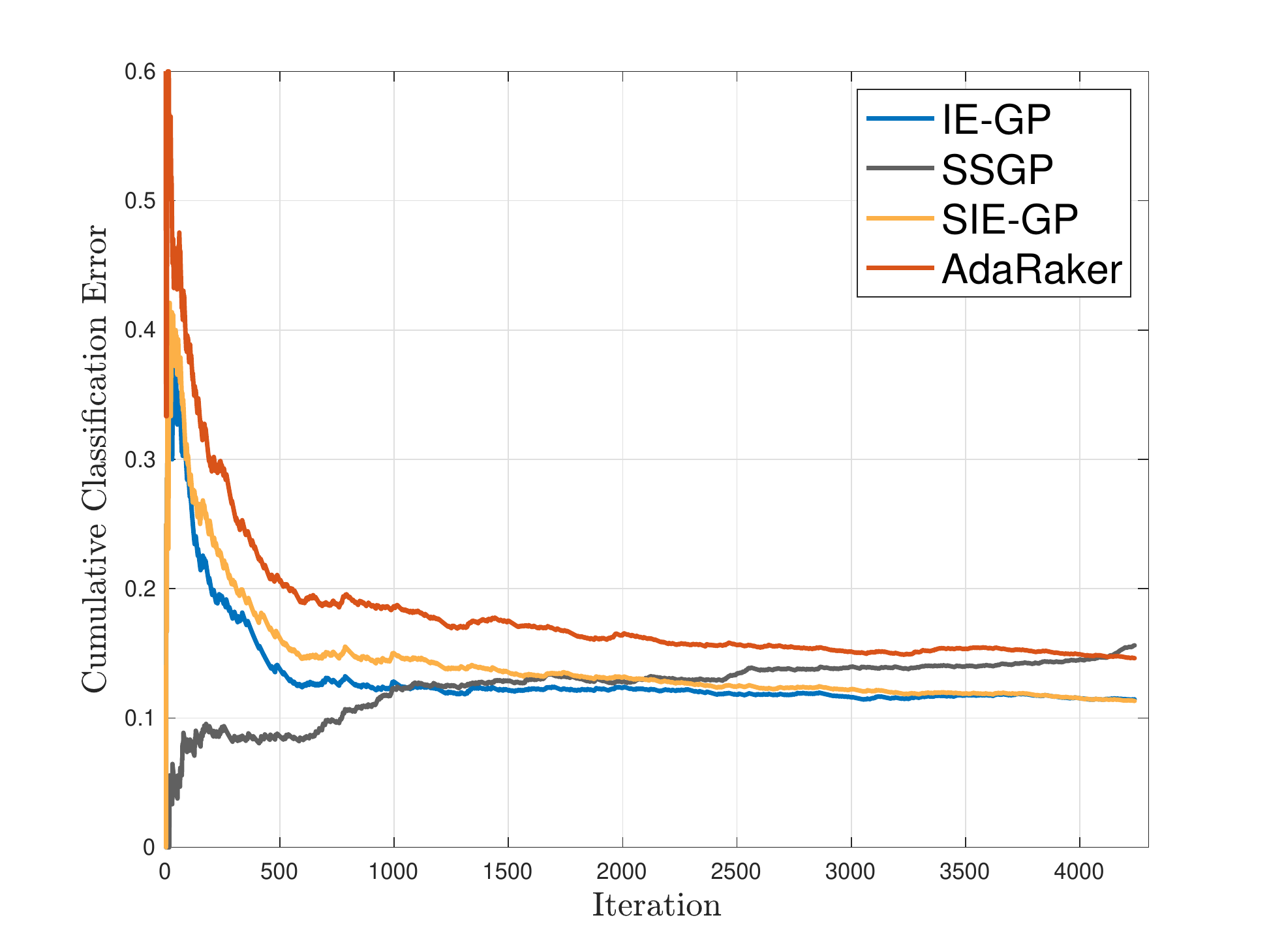}
	{(a)}\label{fig:class_banana}
	\endminipage\qquad
	\hspace{-4mm}
	\minipage[b]{0.32\textwidth}
	\centering
	\includegraphics[width=1.1\linewidth]{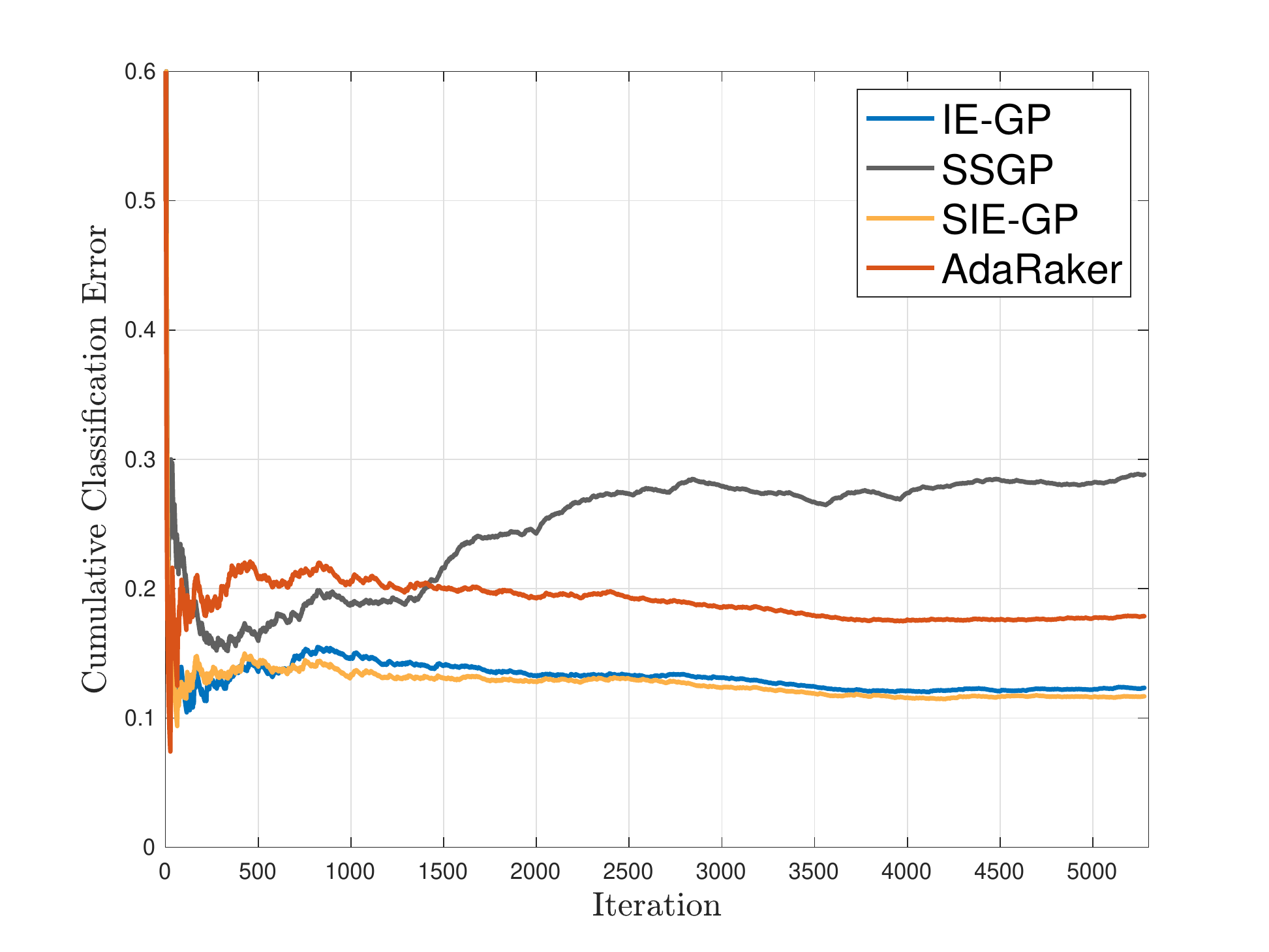}
	{(b)}\label{fig:class_musk}
	\endminipage\qquad
	\hspace{-4mm}
	\minipage[b]{0.32\textwidth}
	\centering
		\hspace{-4mm}
	\includegraphics[width=1.1\linewidth]{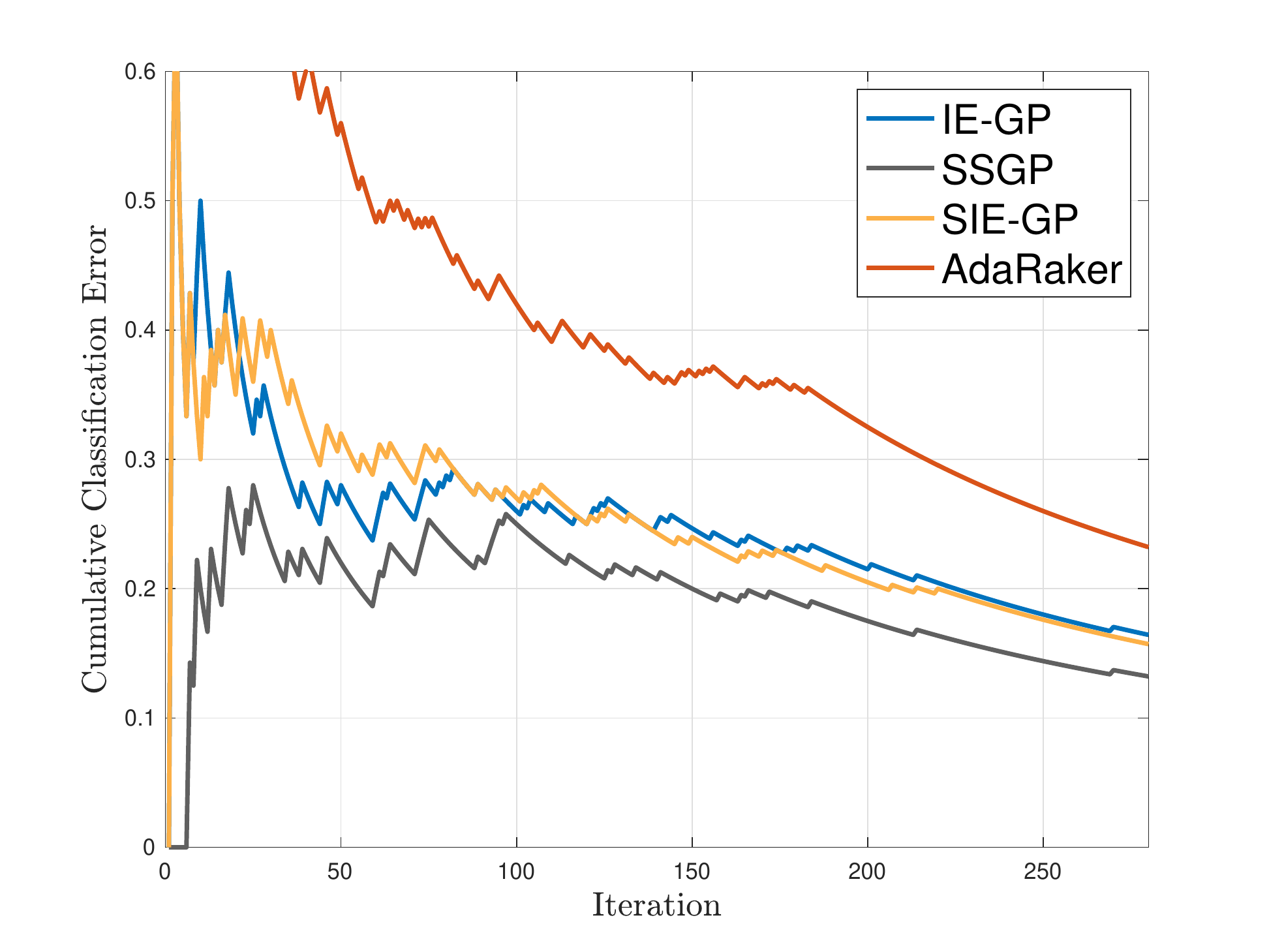}
	{(c)}\label{fig:class_iono}
	\endminipage
	\hspace*{-4cm}
\caption{Cumulative classification errors on  (a) \texttt{Banana}, (b) \texttt{Musk}, and (c) \texttt{Ionosphere} datasets.}\label{fig:Pe_d}
\end{figure*}

The nMSE performance of the tested approaches on the \texttt{Tom's hardware} dataset is plotted in Fig.~\ref{fig:common}(a). The proposed (D)IE-GP approaches outperform the competing alternatives in terms of nMSE while also featuring the lowest running time, which corresponds to less than $0.3\%$ of that of the most closely competing (in terms of nMSE) alternative (cf. Fig.~\ref{fig:run_all}(a)). The results on the \texttt{SARCOS} dataset are depicted in Fig.~\ref{fig:common}(b). Our IE-GP remains competitive whereas the proposed dynamic variant (DIE-GP) features the lowest nMSE, while also achieving both faster convergence as well as a runtime that is an order of magnitude lower than that of the second best (in terms of nMSE) approach (cf. Fig.~\ref{fig:run_all}(b)). These results further highlight the computational efficiency of the proposed approaches. Similar observations can be made on the \texttt{Air quality} (cf. Figs.~\ref{fig:common},\ref{fig:run_all}(c)).

To further demonstrate (D)IE-GP's uncertainty quantification performance, tests were conducted among GP-based approaches regarding the predictive negative log-likelihood (pnLL)  as ${\rm pnLL}_{t}:=-\log p(y_t|\mathbf{y}_{t-1}, \mathbf{X}_t)$, which is computable from \eqref{eq: p_y_pred_ens}.
As illustrated in Fig.~\ref{fig:common_npll}, (D)IE-GP always outperform I-SSGPR; while they outperform SSGP in \texttt{SARCOS} and \texttt{air-quality} datasets; they are comparable in the \texttt{Twitter} dataset; and perform inferior to SSGP on the \texttt{Tom’s hardware} dataset, even though SSGP is two orders of magnitude slower than (D)IE-GP.

\subsection{Classification}
Coupled with the logistic likelihood, our IE-GP and the switching variant were tested for binary classification using Laplace approximation (cf. Sec.~\ref{sec:LA}). The performance of (S)IE-GP were also compared with AdaRaker \cite{shen2019random} and SSGP \cite{bui2017streaming} in terms of classification error and running time on the \texttt{Banana}, \texttt{Musk}, and \texttt{Ionosphere} datasets whose statistics are provided in Table~\ref{table:stat}. DIE-GP are not included for comparison since it achieves similar performance relative to IE-GP. For (S)IE-GP and AdaRaker, the value of $\nrf$ was set to $15$, and the kernel dictionary is the same as in the regression test. Regarding SSGP, the number of inducing points was $30$, the batch size was chosen to be $40$, and the first $20\%$ of the samples were used for model initialization. For (S)IE-GP, the kernel magnitude $\sigma_{\theta^m}^2$, the only hyperparameter per expert, was obtained by maximizing the marginal likelihood using Laplace apprximation~\cite[Chapter~5.5.1]{williams2006gaussian}.

\begin{figure*}
\hspace{-2mm}
	\minipage[b]{0.32\linewidth}%
	\centering
	\hspace{-2mm}
	\includegraphics[width=1\linewidth,height =1.0\linewidth ]{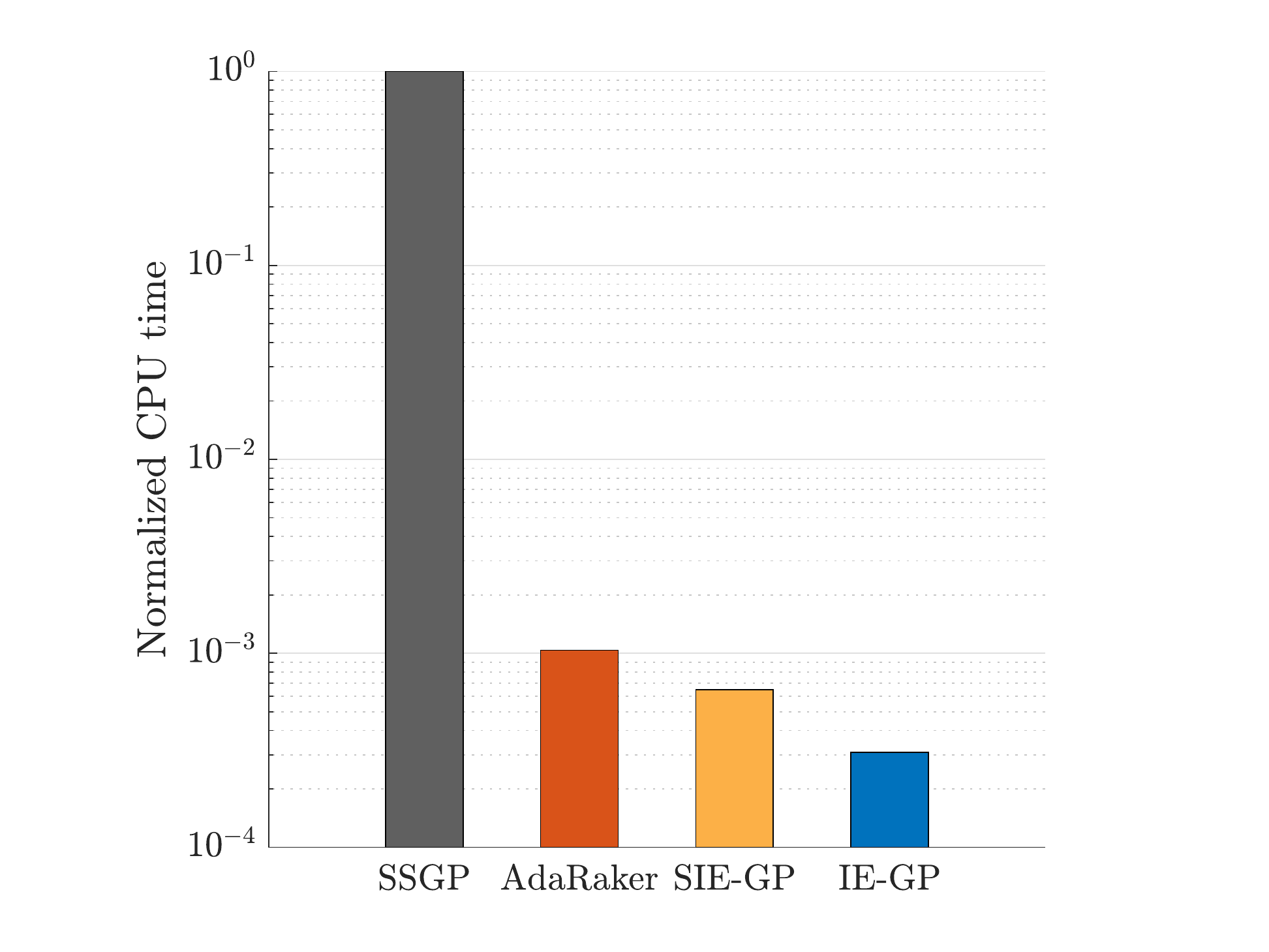}\\
	{(a)}\label{fig:rt_banana}
	\endminipage
	\minipage[b]{0.32\linewidth}%
	\centering
	\hspace{-2mm}
	\includegraphics[width=1\linewidth,height =1\linewidth]{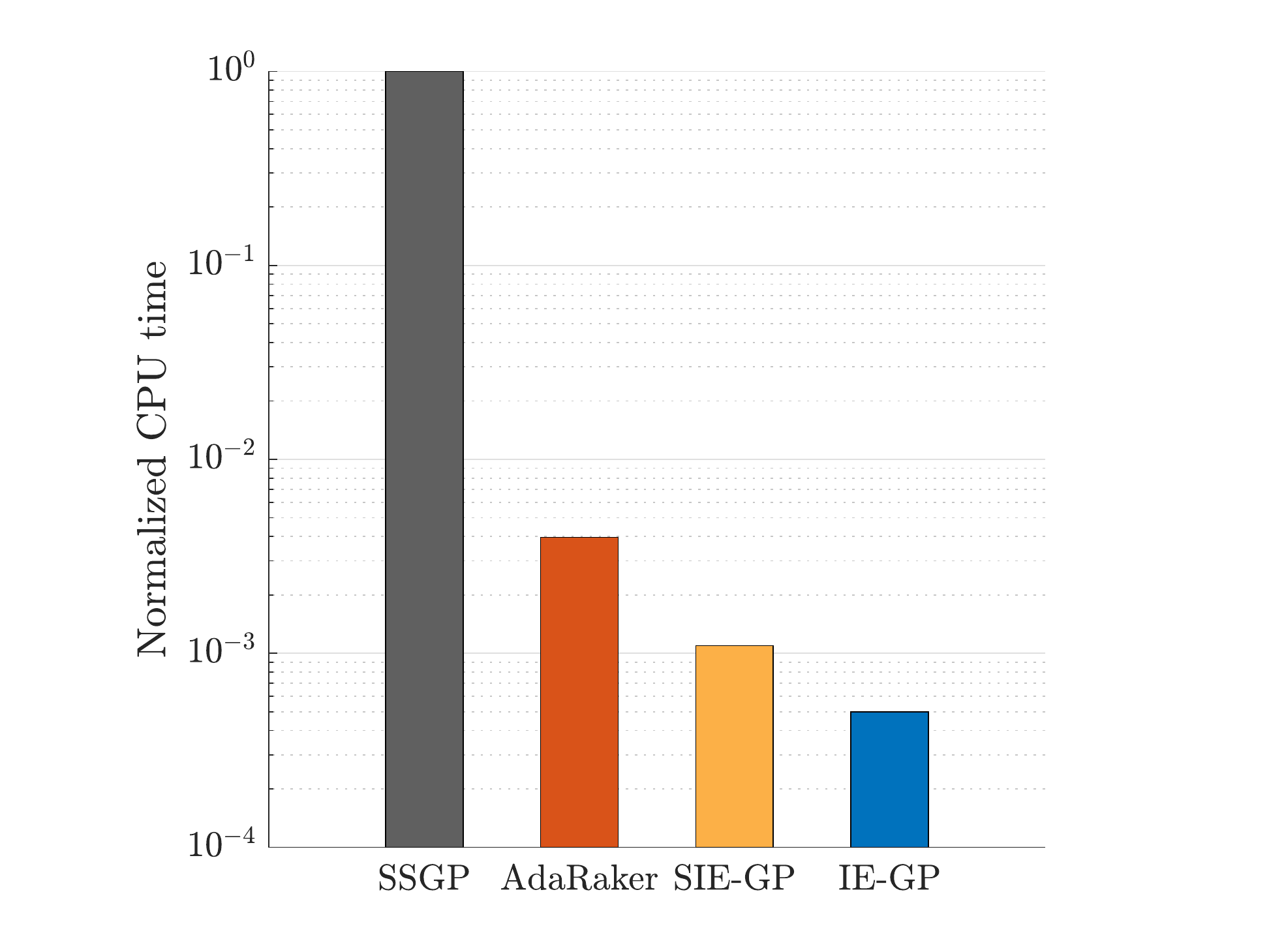}\\
	{(b)}\label{fig:rt_musk}
	\endminipage
	\minipage[b]{0.32\linewidth}%
	\centering
	\hspace{-2mm}
	\includegraphics[width=1\linewidth,height =1\linewidth]{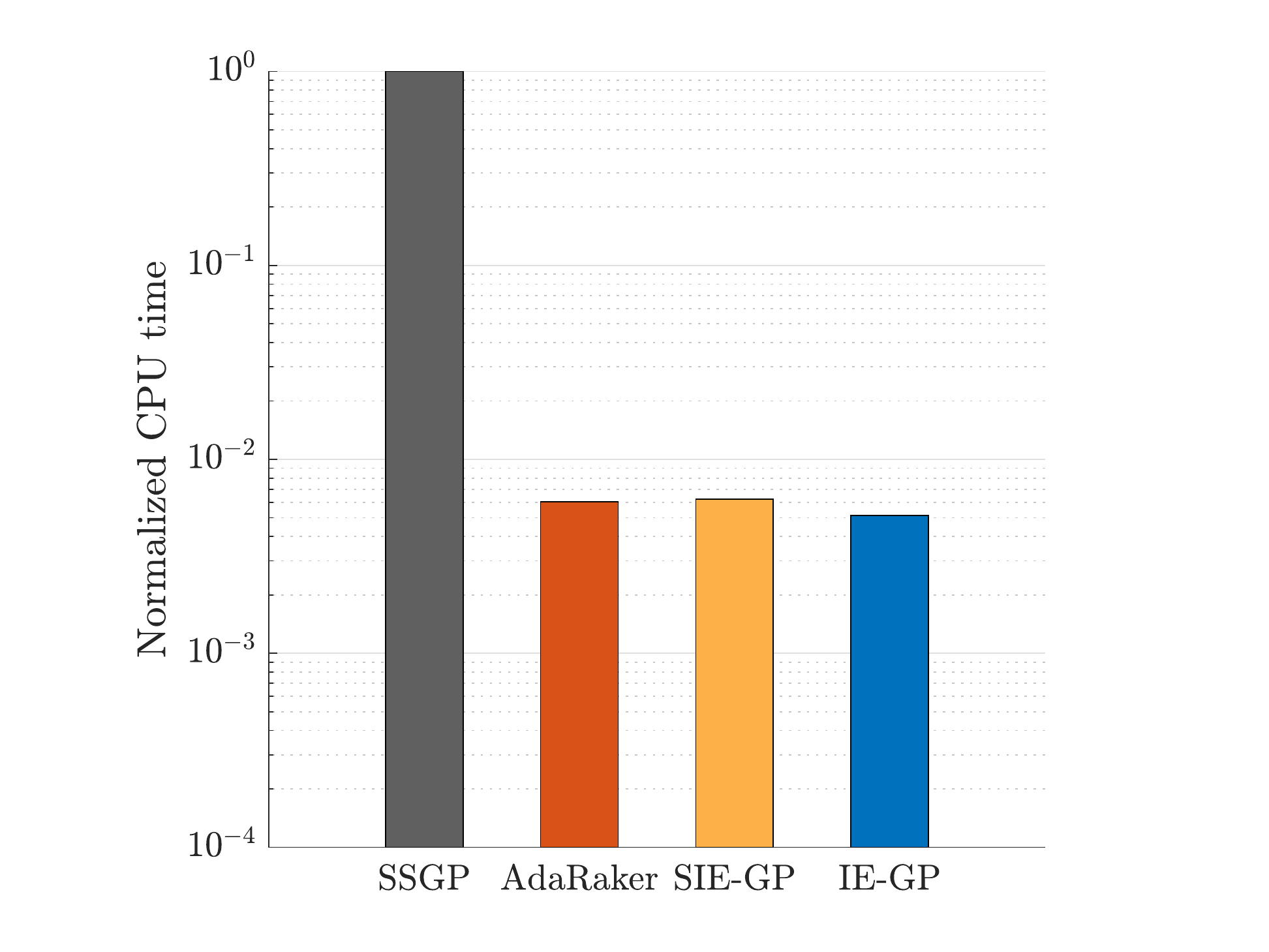}
	\\{(c)}\label{fig:rt_iono}
\endminipage
\vspace{-1mm}
\caption{Normalized running times for classification on  (a) \texttt{Banana}, (b) \texttt{Musk}, and (c) \texttt{Ionosphere} datasets.}
		\label{fig:time_d}
\end{figure*}

The cumulative classification error and running time of the four competing approaches are plotted in Figs.~\ref{fig:Pe_d}--\ref{fig:time_d}. Clearly, (S)IE-GP outperforms SSGP and AdaRaker in both classification accuracy and computational efficiency on the \texttt{Banana} and \texttt{Musk} datasets. Although achieving lower classification error than (S)IE-GP on the \texttt{Ionosphere} dataset, SSGP runs more than two orders of magnitude slower. Also, it is worth mentioning that the performance of SSGP depends on the batch size. Decreasing its value from $40$ to $20$ yields SSGP performing inferior to (S)IE-GP in classification accuracy. Regarding the two proposed approaches, SIE-GP achieves comparable classification accuracy relative to the static IE-GP since the tested datasets exhibit negligible switching dynamics among the candidate models. In addition, SIE-GP's higher running time is explained by the fact that all the GP experts update at all slots to detect possible model switching, whereas IE-GP implements expert shutdown for computational efficiency.

\begin{figure*}[t]
\hspace{-2mm}
\minipage[b]{0.32\linewidth}%
  \centering
  \includegraphics[width=0.83\linewidth]{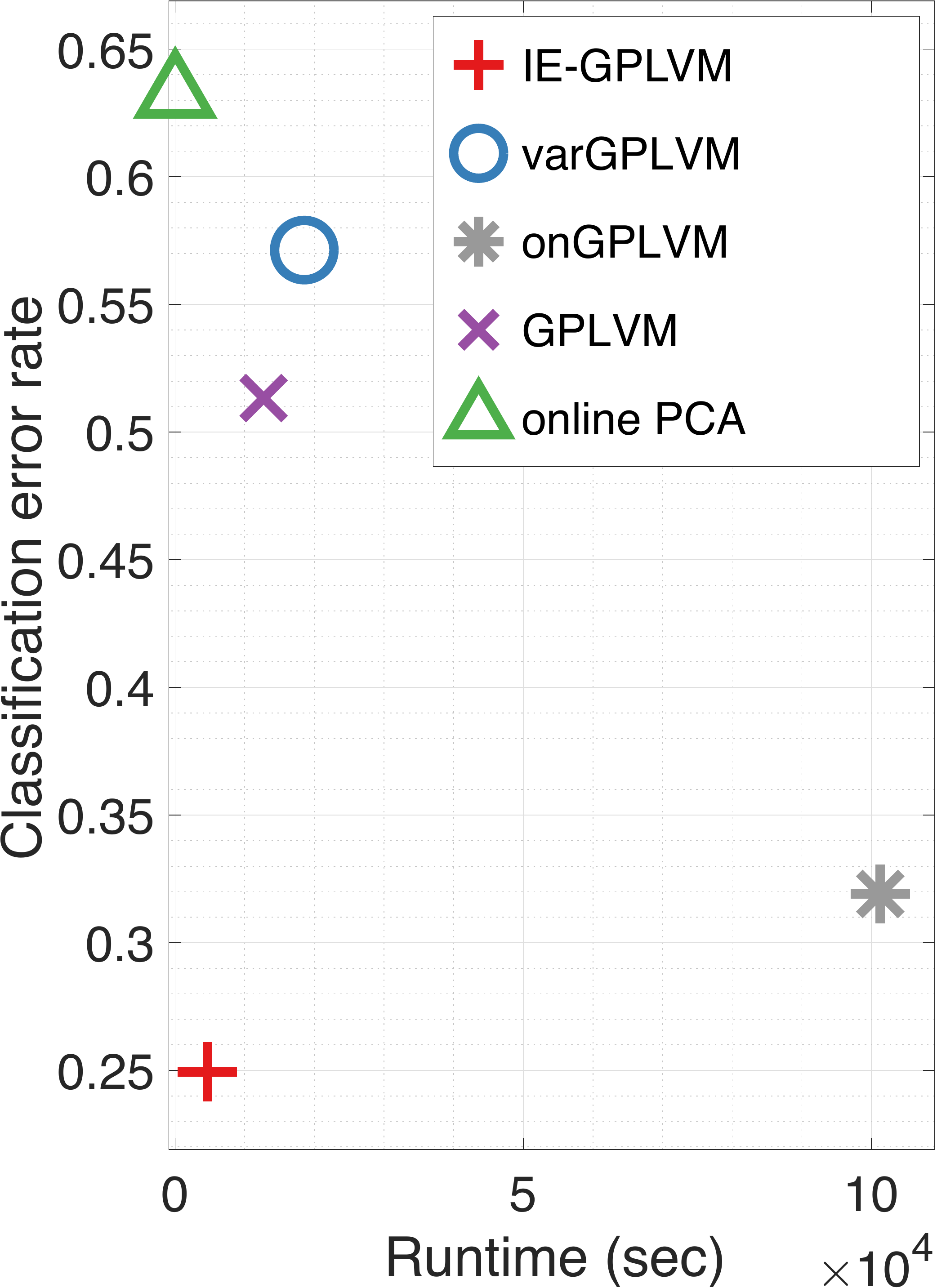}\\
  {(a)}
  \label{fig:mnist1}
\endminipage
\minipage[b]{0.32\linewidth}%
  \centering
  \includegraphics[width=.81\linewidth]{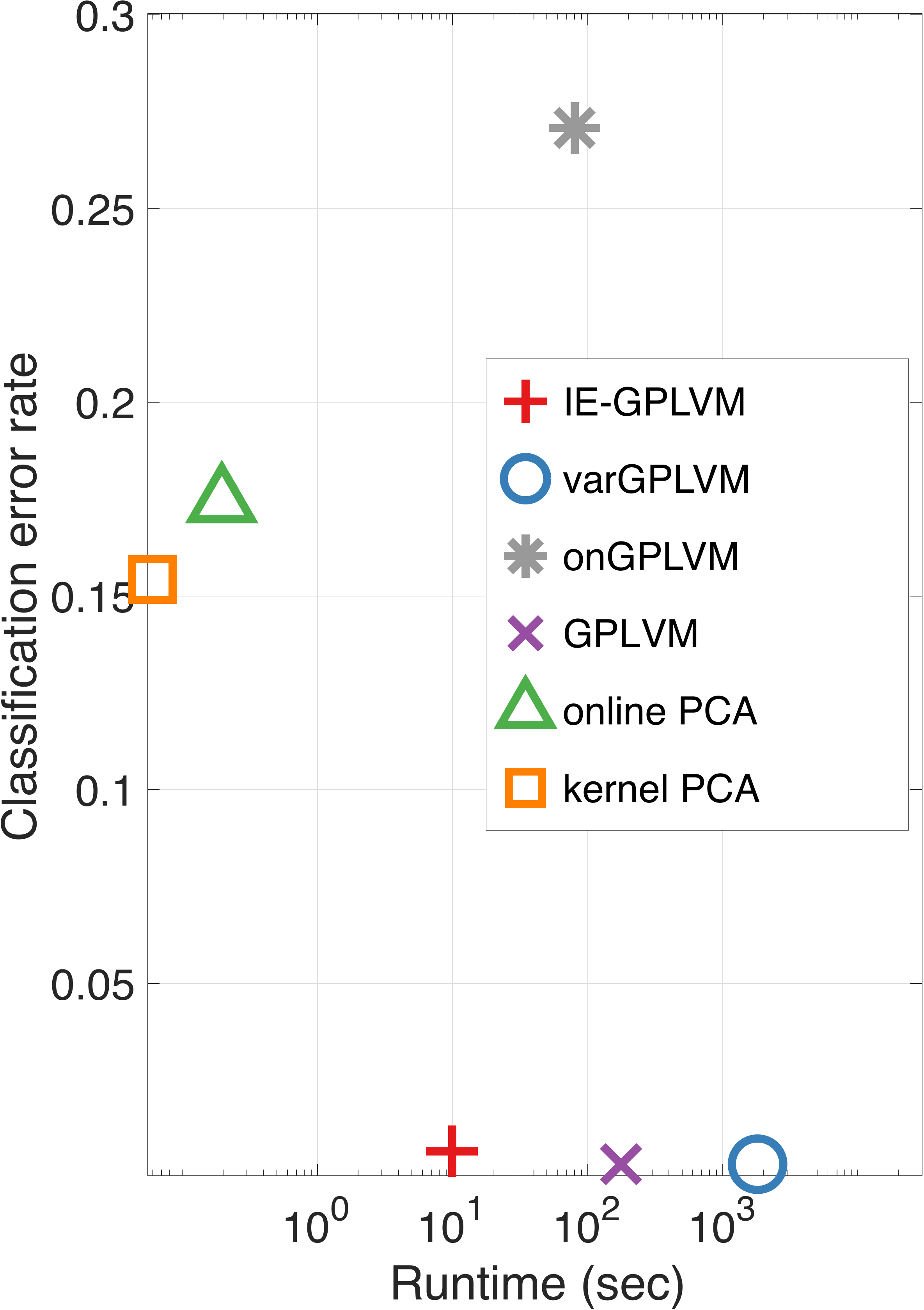}\\
    {(b)}
  \label{fig:oil}
\endminipage
\minipage[b]{0.32\linewidth}%
  \centering
  \includegraphics[width=.85\linewidth]{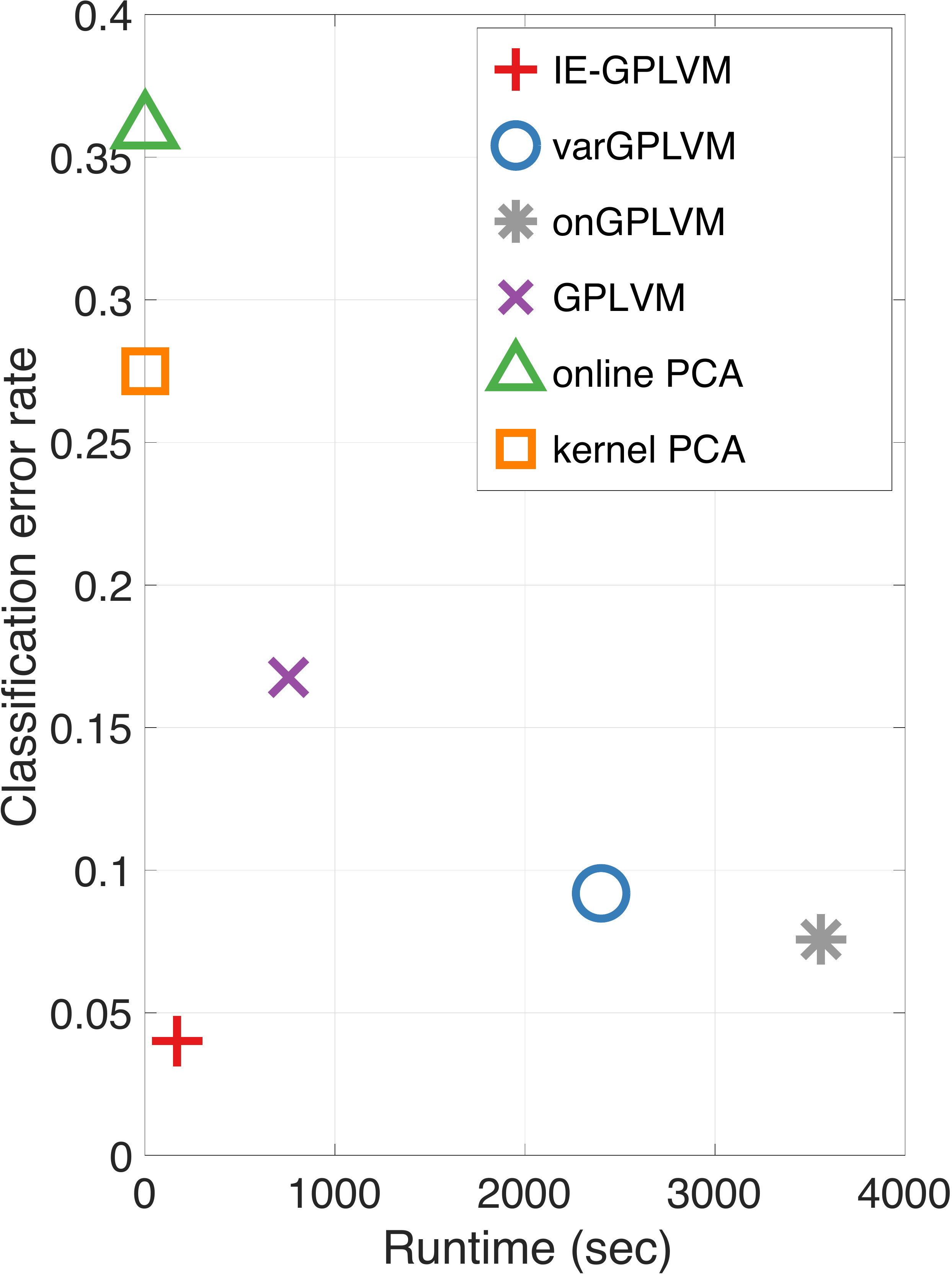}\\
    {(c)}
  \label{fig:usps}
\endminipage
\vspace{-1mm}
\caption{Classification error versus runtime plots for dimensionality reduction on the (a) \texttt{MNIST}, (b) \texttt{oil}, and (c) \texttt{USPS} datasets.}
\label{fig:err_time}
\end{figure*}

\begin{figure}[]
\vspace{-8mm}
\centering
  \includegraphics[width=1.08\linewidth]{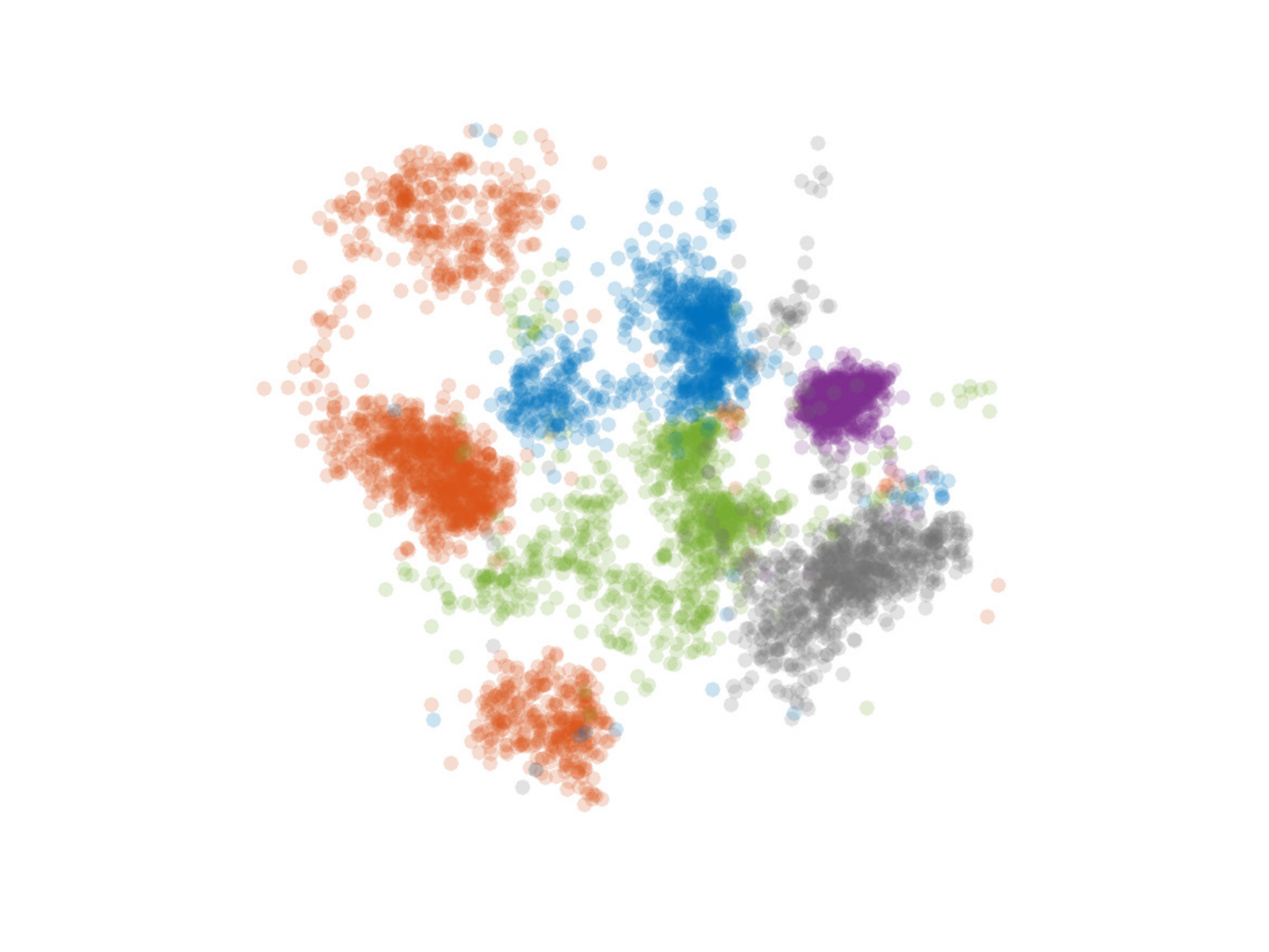}
  \vspace{-1cm}
\caption{Visualization of the embedding attained by IE-GPLVM on the \texttt{USPS} dataset. Colors represent different digits.}
\label{fig:viz}
\end{figure}

\subsection{Dimensionality reduction}
Tests for dimensionality reduction were performed on several benchmark datasets, including \texttt{MNIST} ($D=784$) as well as the \texttt{oil} flow data ($D=12$) ~\cite{oilData}, and the subset of the \texttt{USPS} handwritten digits set ($D=256$) comprising digits $0-4$. . 
{{The latter two} datasets were also used in the original GPLVM paper}~\cite{Lawrence05}. Several competing alternatives were considered. GPLVM based methods comprise the original GPLVM~\cite{Lawrence05,Lawrence07},
a variational inference based scheme 
(varGPLVM)~\cite{damianou2016variational}, as well as an online GPLVM variant (onGPLVM); see Alg. 2 in~\cite{yao2011learning}. PCA based alternatives encompass online PCA~\cite{on1,on2}, and (batch) kernel PCA~\cite{scholkopf1998nonlinear}.
The embedding dimensionality was set to $d=2$, and the results presented correspond to the median across 11 trials. Regarding the proposed IE-GPLVM scheme, $\nrf=50$
random features were used, each  expert relied on a RBF kernel with variance taken from the set  $\{2^k\}_{k=-3}^3$, $t_0$ {was set to} $5\%$ of the number of samples for \texttt{MNIST} and to $10\%$ for the remaining (smaller) datasets, and \eqref{eq:init_opt} was additionally optimized over $\noisevar$.  
For the GPLVM based methods, the RBF kernel was used, $100$ inducing points were utilized, and the maximum number of iterations was set to $1,000$. Initializations were provided by means of  PPCA embeddings. Finally, for kernel PCA (kPCA), a grid search was performed over RBF kernels with variances in $\{2^k\}_{k=-10}^{10}$ and the lowest error rate achieved is reported. Note that for kPCA the reported runtime does not include the computational time required for the grid search. 

The error rate of the nearest neighbor classification rule was used as the performance metric,  when applied to the resultant embeddings; see e.g.~\cite{Lawrence05}. The results for the three tested datasets are summarized in Fig. \ref{fig:err_time}, in the form of error rate versus runtime plots. In the \verb|MNIST| dataset, the proposed \verb|IE-GPLVM| approach achieves both the lowest overall error rate, as well as runtime among GPLVM schemes. The only method achieving a somewhat similar error rate, namely \texttt{onGPLVM}, has a runtime that is more than $20$ times higher relative to our approach. In the \verb|oil| dataset, our IE-GPLVM achieves similar error rate (less than $1 \%$) to \verb|GPLVM| and \verb|varGPLVM|, while being more than one and two orders of magnitude faster, respectively. In the \verb|USPS| set, our approach achieves the lowest overall error rate. The only schemes that achieve error rates in the same order of magnitude, namely \verb|varGPLVM| and \verb|onGPLVM|,  have runtimes that are $14$ and $21$ times higher, respectively. Finally, in both experiments, PCA based schemes, although computationally efficient, yield high error rates. A visualization of the embedding attained is provided for the proposed IE-GPLVM on the \verb|USPS| dataset (Fig.~\ref{fig:viz}). We can observe that good separation between clusters of different digits is achieved, in line with the low classification error rate in Fig. \ref{fig:err_time}.

\section{Conclusions}
This paper put forth an incremental scheme that leverages an ensemble of scalable RF-based parametric GP learners to jointly infer the unknown function along with its performance, and a data-driven kernel combination. Dynamic function learning was enabled through modeling structured dynamics for the EGP meta-learner and individual GP learners. On the theoretical aspect, regret analysis was conducted to benchmark even in adversarial settings the novel IE-GP and its dynamic variant relative to benchmark strategies with data in hindsight. Further, EGP-based latent variable model is devised for online kernel-adaptive dimensionality reduction. Extensive experimental results are provided to illustrate the superior performance of the novel IE-GP schemes.

\section{Proofs} 
\subsection{Proof of Lemma 1}\label{sec:proof_Lemma1}
To prove Lemma 1, we will first upper bound the cumulative online Bayesian loss associated with IE-GP, $\sum_{\tau = 1}^T \ell_{\tau|\tau-1}$, relative to that incurred by any RF-based GP expert $m$, namely $\sum_{\tau = 1}^T l_{\tau|\tau-1}^m$. Reorganizing \eqref{eq:w_update_1}, we have
$\exp(-\ell_{\tau|\tau-1})/\exp(-l_{\tau|\tau-1}^m) = w_{\tau-1}^m /w_{\tau}^m$, multiplying which \eqref{eq:w_update_1} from $\tau = 1$ to $T$, it follows that
$\exp(-\sum_{\tau = 1}^T \ell_{\tau|\tau-1}+\sum_{\tau = 1}^T l_{\tau|\tau-1}^m) = 1/(M w_{T}^m)$,
yielding
\begin{align}
	\sum_{\tau = 1}^T \ell_{\tau|\tau-1} \! -\!\sum_{\tau = 1}^T l_{\tau|\tau-1}^m \!= \log M \!+\! \log  w_{T}^m \! \overset{(a)}{\leq}  \! \log M \label{eq:L_mix_single}
\end{align}
where $(a)$ holds because $w_{T}^m \in [0, 1]$. 

Next, we will bound the difference between $\sum_{\tau = 1}^T l_{\tau|\tau-1}^m$ 
and the cumulative loss incurred by a fixed strategy $\bbtheta^{m}_*$, for any expert $m \in \mathcal{M}$. For notational brevity, we will drop expert index $m$ in the remaining of the proof.

 Upon defining the cumulative loss over $T$ slots with a time-invariant $\bbtheta$ as $\mathcal{L}_\theta := \sum_{\tau = 1}^T \mathcal{L} ( \bbphi_\mathbf{v}^\top(\mathbf{x}_\tau)\bbtheta; y_\tau )= - \log p(\mathbf{y}_T | \bbtheta,\mathbf{X}_T)$, the expected cumulative loss over any pdf  $q(\bbtheta)$ is~\cite{kakade2005online}
\begin{align}
	\bar{\mathcal{L}}_{q_\theta} := \mathbb{E}_q [\mathcal{L}_\theta] = \int_{\bbtheta} q(\bbtheta) \mathcal{L}_\theta d\bbtheta \;. \nonumber
\end{align} 
%
On the other hand, the following equality holds for the cumulative online Bayesian loss based on Bayes' rule
	\begin{align}
		\sum_{\tau = 1}^{T} \ell_{\tau|\tau-1} &= \sum_{\tau = 1}^{T} -\log p(y_\tau| \mathbf{y}_{\tau-1},\mathbf{X}_\tau)  = -\log p(\mathbf{y}_T|\mathbf{X}_T)   \nonumber\;.
	\end{align}
Let $q(\bbtheta) = \mathcal{N}(\bbtheta; \bbtheta_{*}, \xi^2 \mathbf{I}_{2\nrf})$ with the variational parameter $\xi$ to be tuned later, and $p(\bbtheta) = \mathcal{N}(\bbtheta; \mathbf{0}, \sigma_{\theta}^2 \mathbf{I}_{2\nrf})$. It then follows that
	\begin{align}
		&\sum_{\tau = 1}^{T} l_{\tau|\tau-1}- \bar{\mathcal{L}}_{q_\theta} = \int q(\bbtheta) \log \frac{p(\mathbf{y}_T |\bbtheta,\mathbf{X}_T)}{p(\mathbf{y}_T| \mathbf{X}_T)} d\bbtheta \nonumber\\
		& \overset{(a)}{=} \int q(\bbtheta) \log \frac{p(\bbtheta| \mathbf{y}_T,\mathbf{X}_T)}{p(\bbtheta)} d\bbtheta \nonumber \\
		&  = \int q(\bbtheta) \log \frac{q(\bbtheta)}{p(\bbtheta)} d\bbtheta 
		- \int q(\bbtheta) \log \frac{q(\bbtheta)}{p(\bbtheta| \mathbf{y}_T, \mathbf{X}_T)} d\bbtheta \ \ \nonumber\\
		& = \!{\rm KL}(q(\!\bbtheta)\|p(\!\bbtheta)\!) \!-\! {\rm KL}(q(\!\bbtheta)\|p(\!\bbtheta| \mathbf{y}_T,\mathbf{X}_T)\!) \!\overset{(b)}{\leq} \! {\rm KL}(q(\!\bbtheta)\|p(\!\bbtheta)\!)\nonumber	\\
		&= 2\nrf \log \sigma_{\theta} + \frac{\|\bbtheta_{*}\|^2+2\nrf \xi^2}{2\sigma_{\theta}^2}\!-\!\nrf\!-\!2\nrf\log\xi    \label{eq:KL(q||p_0)}
	\end{align}
where $(a)$ holds since Bayes' rule yields $ p(\mathbf{y}_T| \bbtheta, \mathbf{X}_T)p(\bbtheta)  = p(\mathbf{y}_T| \mathbf{X}_T)p(\bbtheta| \mathbf{y}_T, \mathbf{X}_T)$; and, $(b)$ comes from the fact that ${\rm KL}(q(\!\bbtheta)\|p(\!\bbtheta| \mathbf{y}_T, \!\mathbf{X}_T)\!)\geq 0$.


The last step towards bounding $\sum_{\tau = 1}^{T} l_{\tau|\tau-1}-\mathcal{L}_{\theta_{*}}$ is to establish an upper bound for $\bar{\mathcal{L}}_{q_\theta} -\mathcal{L}_{\theta_{*}}$. To this end,
let $z_\tau = \bbphi_{\mathbf{v}}^\top (\mathbf{x}_\tau)\bbtheta$ and $z_\tau^{*} =  \bbphi^\top_{\mathbf{v}} (\mathbf{x}_\tau) \bbtheta_{*} $. Taking the Taylor's expansion of $\mathcal{L}(z_\tau; y_\tau)$ around $z_\tau^{*}$, yields
\begin{align}
	\mathcal{L}(z_\tau;y_\tau) &= \mathcal{L}(z_\tau^{*};y_\tau) +  \frac{d \mathcal{L}(z_\tau^{*};y_\tau)}{dz_\tau}  (z_\tau - z_\tau^{*}) \nonumber\\
	&\ \ \ \ +     \frac{d^2}{dz_\tau^2}\mathcal{L}(h(z_\tau);y_\tau)  \frac{(z_\tau-z_\tau^{*})^2}{2} \label{eq:L(z,y)}
\end{align}
where $h(z_\tau)$ is some function lying between $z_\tau$ and $z_\tau^*$. Taking the expectation of \eqref{eq:L(z,y)} wrt $q(\bbtheta)$, leads to
\begin{align}
	\mathbb{E}_{q} [\mathcal{L}(z_\tau;y_\tau)] \! -\! \mathcal{L}(z_\tau^{*};y_\tau)&=  \mathbb{E}_q \left[\frac{d^2}{dz_\tau^2}\mathcal{L}(h(z_\tau);y_\tau)  \frac{(z_\tau-z_\tau^{*})^2}{2} \right] \nonumber\\
	& \overset{(a)}{\leq}  c \mathbb{E} \left[ \frac{(z_\tau-z_\tau^{*})^2}{2} \right] \overset{(b)}{\leq}  \frac{c  \xi^2}{2}
	\label{eq:E(g_y(z))}
\end{align}
where $(a)$ makes use of (as1) that $\left| \frac{d^2}{dz^2}\mathcal{L}(z;y) \right| \leq c\  \forall z$, and $(b)$ relies on the equality $ \| \bbphi_{\mathbf{v}} (\mathbf{x}_\tau)\|^2 =1$. 

Summing \eqref{eq:E(g_y(z))} from $\tau = 1$ to $T$, we have
\begin{align}
	\bar{\mathcal{L}}_{q_\theta}  \leq \mathcal{L}_{\theta_{*}} + \frac{Tc \xi^2}{2}  \label{eq: 56}
\end{align}
which, in conjunction with \eqref{eq:KL(q||p_0)}, yields the inequality 
\begin{align}
	&\sum_{\tau = 1}^{T} l_{\tau|\tau-1} -\mathcal{L}_{\theta_{*}} \label{eq:Lq_Lstar_1}\\
	&\leq  \frac{Tc \xi^2}{2} \!+\! 2\nrf\log \sigma_{\theta} \!+\! \frac{\|\bbtheta_{*}\|^2+2\nrf \xi^2}{2\sigma_{\theta}^2} \!-\! \nrf\!-\! 2\nrf\log\xi    \nonumber \;.
\end{align}
Replacing the RHS of \eqref{eq:Lq_Lstar_1}, a convex function of $\xi$,  with the minimal value taken at $\xi^2 = \frac{2\nrf \sigma_{\theta}^2}{2\nrf + Tc\sigma_{\theta}^2}$, simplifies \eqref{eq:Lq_Lstar_1} for any expert $m\in \mathcal{M}$ to 
\begin{align}
	\sum_{\tau = 1}^T l_\tau^m \!-\!\! \sum_{\tau = 1}^T\mathcal{L}\!\left(\! \bbphi^{m\top}_{\mathbf{v}}\! (\mathbf{x}_\tau\!)\bbtheta^{m}_*; y_\tau \!\right) \! \leq\! \frac{\|\bbtheta^{m}_*\|^2}{2\sigma_{\theta^m}^2} \!+\! \nrf \log \!\left(\! 1\! +\! \frac{Tc \sigma_{\theta^m}^2}{2\nrf} \!\right) \nonumber 
\end{align}
which, together with \eqref{eq:L_mix_single}, readily prove Lemma 1.

\subsection{Proof of Theorem 1} \label{sec:proof_Thm1}
For a given shift-invariant standardized kernel $\bar{\kappa}^m$, the maximum point-wise error of the RF kernel approximant is uniformly bounded with probability at least $1-2^8  (\frac{\sigma_m}{\epsilon})^2 \exp\left(\frac{-\nrf\epsilon^2}{4d+8} \right)$ \cite{rahimi2008random}
\begin{align}
	\sup_{\mathbf{x}_i, \mathbf{x}_j \in \mathcal{X}} \left|\bbphi^{m\top}_{\mathbf{v}}(\mathbf{x}_i)\bbphi^m_{\mathbf{v}}(\mathbf{x}_j) - \bar{\kappa}^m (\mathbf{x}_i, \mathbf{x}_j)\right| < \epsilon \label{eq:kernel_approx}
\end{align}
where $\epsilon$ is a given constant, $\nrf$ is the number of spectral feature vectors, $d$ is the dimension of $\mathbf{x}$, and $\sigma_m^2 : = \mathbb{E}_{\pi_{\bar{\kappa}}^m}[\| \mathbf{v}^m\|^2]$ is the second-order moment of the RF vector $\mathbf{v}^m$.

The optimal function estimator in $\mathcal{H}^m$ incurred by $\kappa^m$ is
\begin{align}
	\hat f^{m} (\mathbf{x}):= \sum_{\tau = 1}^T \hat \alpha^{m}_{\tau} \kappa^m (\mathbf{x}, \mathbf{x}_\tau) =  \sigma_{\theta^m}^2\sum_{\tau = 1}^T \hat \alpha^{m}_{\tau} \bar{\kappa}^m (\mathbf{x}, \mathbf{x}_\tau)
\end{align}
and its RF-based approximant is $\check{f}^{m}_* (\mathbf{x}):= \bbphi^{m\top}_{\mathbf{v}}(\mathbf{x})\bbtheta^m_*$ with $\bbtheta^{m}_*:=\sigma_{\theta^m}^2\sum_{\tau = 1}^{T} \hat \alpha_{\tau}^{m} \bbphi^{m}_{\mathbf{v}} (\mathbf{x}_\tau)$. We then have that 
\begin{align}
	&\left|\sum_{\tau = 1}^T\mathcal{L} \left(\check{f}^{m}_* (\mathbf{x}_\tau); y_\tau\right) - \sum_{\tau = 1}^T \mathcal{L} \left(\hat f^{m} (\mathbf{x}_\tau); y_\tau \right)   \right|\label{eq:L_ff} \\
	& \overset{(a)}{\leq} 
	\sum_{\tau = 1}^{T} \left|\mathcal{L} \left(\check{f}^{m}_* (\mathbf{x}_\tau); y_\tau\right) -  \mathcal{L}\left(\hat f^{m} (\mathbf{x}_\tau); y_\tau \right) \right|   \nonumber \\
	& \overset{(b)}{\leq}   \sum_{\tau = 1}^{T} L\sigma_{\theta^m}^2\! \left|\sum_{\tau'=1}^{T}\! \hat\alpha_{\tau'}^{m} \bbphi_{\mathbf{v}}^{m\top}\! (\mathbf{x}_\tau) \bbphi^m_{\mathbf{v}}(\mathbf{x}_{\tau'}) \!-\!\!\! \sum_{\tau'=1}^{T}\!\! \hat\alpha_{\tau'}^{m} \bar{\kappa}^m (\mathbf{x}_\tau, \mathbf{x}_{\tau'})   \right| \nonumber \\
	&\overset{(c)}{\leq}  \sum_{\tau = 1}^{T} L\sigma_{\theta^m}^2 \sum_{\tau'=1}^{T} \left|\hat\alpha_{\tau'}^{m}\right| \left|\bbphi_{\mathbf{v}}^{m\top} (\mathbf{x}_\tau) \bbphi_{\mathbf{v}}^m(\mathbf{x}_{\tau'}) -   \bar{\kappa}^m (\mathbf{x}_\tau, \mathbf{x}_{\tau'})   \right| \nonumber
\end{align}
where $(a)$ follows from the triangle inequality; $(b)$ makes use of (as2), which states the convexity and bounded derivative of $\mathcal{L}(z;y)$ wrt $z$, and $(c)$ results from the Cauchy-Schwarz inequality. Leveraging \eqref{eq:kernel_approx} to upper bound the RHS of \eqref{eq:L_ff}, we find
\begin{align}
	&\left|\sum_{\tau = 1}^T\mathcal{L} \left(\check{f}^{m}_* (\mathbf{x}_\tau); y_\tau\right) - \sum_{\tau = 1}^T \mathcal{L} \left(\hat f^{m} (\mathbf{x}_\tau); y_\tau \right)   \right| \nonumber\\
	& \leq
	\sum_{\tau = 1}^T L \sigma_{\theta^m}^2\epsilon \sum_{\tau = 1}^T \left|\hat\alpha_{\tau}^{m}\right| \leq \epsilon LTC, \ {\rm w.h.p.} \label{eq:lambda_RF_RKHS_1}
\end{align}
where $C:=\underset{m \in \mathcal{M}}{\max} \sum_{\tau = 1}^T \sigma_{\theta^m}^2 | \hat\alpha_{\tau}^{m}|$. It thus holds w.h.p. that
\begin{align}
\hspace*{-0.2cm}
	\sum_{\tau = 1}^T\!\mathcal{L}\! \left(\!\bbphi^{m\top}_{\mathbf{v}}(\mathbf{x}_\tau){\bbtheta^m_*} ; y_\tau\!\right) \!-\! \sum_{\tau = 1}^T\! \mathcal{L}\! \left(\!\hat f^{m} (\mathbf{x}_\tau); y_\tau \!\right)  \leq \epsilon LTC. \label{eq:lambda_RF_RKHS_2}
\end{align}
On the other hand, the uniform convergence bound in \eqref{eq:kernel_approx} and (as3) imply w.h.p. that
\begin{align}
	\underset{\mathbf{x}_\tau, \mathbf{x}_{\tau'} \in \mathcal{X}}{\sup }\!\!\bbphi^{m\top}_{\mathbf{v}}\!\! (\mathbf{x}_\tau) \bbphi^m_{\mathbf{v}}(\mathbf{x}_{\tau'})\! \leq\! 1\!+\! \epsilon
\end{align}
based on which
\begin{align}
	&\|\bbtheta^{m}_{*} \|^2:=  \left\|\sigma_{\theta^m}^2\sum_{\tau = 1}^T \hat\alpha^{m}_{\tau} \bbphi^m_{\mathbf{v}} (\mathbf{x}_\tau)  \right\|^2 \label{eq:theta_bound}\\
	&=  \sigma_{\theta^m}^4\sum_{\tau = 1}^T\sum_{\tau' = 1}^T \hat\alpha^{m}_{\tau} \hat\alpha^{m}_{\tau'}\bbphi^{m\top}_{\mathbf{v}} (\mathbf{x}_\tau)\bbphi^m_{\mathbf{v}} (\mathbf{x}_{\tau'})  \leq (1+\epsilon)C^2   \; . \nonumber
\end{align}
Hence, in conjunction with \eqref{eq:theta_bound}, \eqref{eq:lambda_RF_RKHS_2} and Lemma 1, it follows for any $m\in \mathcal{M}$ that
\begin{align}
	&\sum_{\tau = 1}^T \ell_{\tau|\tau-1} -  \sum_{\tau = 1}^T \mathcal{L}(\hat f^{m}(\mathbf{x}_\tau); y_\tau)  \\
	&\leq  \nrf \log \left(1+ \frac{Tc \sigma_{\theta^m}^2}{2\nrf} \right) + \log M + \frac{(1+\epsilon)C^2}{2\sigma_{\theta^m}^2}+\epsilon LTC  \nonumber
\end{align}
thus completing the proof of Theorem 1 with $m = m^{*}$.

\subsection{Proof of Lemma 2}\label{sec:proof_Lemma2}
Inspired by \cite{cesa2006prediction}, the proof of Lemma 2 will be conducted relying on the notion of compound experts, each associated with a sequence of contributing models over $T$ slots, denoted as $\mathbf{i}_T = [i_1, \ldots, i_T]^{\top}$. Let $\bar{w}_{t} (\mathbf{i}_T)$ represent the posterior weight of compound expert $\mathbf{i}_T$ at slot $t$ as $\bar{w}_{t} (\mathbf{i}_T):={\rm Pr}(\mathbf{i}_T| \mathbf{y}_t, \mathbf{X}_t)$, which is updated at slot $t+1$ as
	\begin{align}
	\bar{w}_{t+1} (\mathbf{i}_T) &= 
	\frac{\bar{w}_{t} (\mathbf{i}_T) p(y_{t+1}|\mathbf{y}_{t}, \mathbf{i}_T,\mathbf{X}_{t+1}) }{ p(y_{t+1}|\mathbf{y}_{t}, \mathbf{X}_{t+1})} \nonumber\\
 & \overset{(a)}{=} \bar{w}_{t} (\mathbf{i}_T)\exp\left(\bar{\ell}_{t+1|t}-l_{t+1|t}^{i_{t+1}} \right) \label{eq:w_comp}
\end{align}
where equality $(a)$ holds since $ p(y_{t+1}|\mathbf{y}_{t}, \mathbf{i}_T,\mathbf{X}_{t+1}) =p(y_{t+1}|\mathbf{y}_{t}, i_{t+1},\mathbf{X}_{t+1})$, and $\bar{\ell}_{t+1|t}$ signifies the ensemble online loss accounting for all the compound experts as
\begin{align}
\bar{\ell}_{t+1|t}&:= -\log \sum_{i_1,\ldots,i_T} \bar{w}_{t} (\mathbf{i}_T) p(y_{t+1}|\mathbf{y}_{t}, \mathbf{i}_T,\mathbf{X}_{t+1}) \nonumber\\
&\ = -\log \sum_{i_{t+1}} \bar{w}_{t+1} ^{i_{t+1}} \exp(-l_{t+1|t}^{i_{t+1}})
\end{align}
with marginal weight at slot $t+1$ given by
\begin{align}
\bar{w}_{t+1} ^{i_{t+1}}:=\sum_{i_1,\ldots,i_{t}, i_{t+2},\ldots, i_T} \bar{w}_{t}  (\mathbf{i}_T)\;. \label{eq:w_marg}
\end{align}
Next, the equivalence of $\bar{\ell}_{t+1|t}$ and ${\ell}_{t+1|t}^{\rm SW}$ \eqref{eq:track_loss} will be established via proving the equality $\bar{w}_{t +1}^{i_{t+1}}=w_{t+1|t}^{i_{t+1}}$ by induction. Towards this, we will first leverage \eqref{eq:w_comp} to obtain
\begin{align}
\bar{w}_{t} (\mathbf{i}_T) \propto \bar{w}_{0}  (\mathbf{i}_T)  \exp\left(\sum_{\tau=1}^{t} -l_{\tau|\tau-1}^{i_{\tau}}\right) \nonumber
\end{align}
based on which, \eqref{eq:w_marg} can be rewritten as
\begin{align}
 \bar{w}_{t+1}^{i_{t+1}} 
 & \propto \sum_{i_1,\ldots,i_{t}}  \bar{w}_{0}  (\mathbf{i}_{t+1})  \exp\left(\sum_{\tau=1}^{t} -l_{\tau|\tau-1}^{i_{\tau}}\right)\nonumber\\
 & = \sum_{i_{t}} \frac{\bar{w}_{0}  (\mathbf{i}_{t+1}) }{\bar{w}_{0}  (\mathbf{i}_{t}) } e^{-l_{t|t-1}^{i_{t}}} \hspace*{-0.4cm}\sum_{i_1,\ldots,i_{t-1}} \hspace*{-0.3cm}\bar{w}_{0} (\mathbf{i}_{t}) \exp\left(\sum_{\tau=1}^{t-1} -l_{\tau|\tau-1}^{i_\tau}\right)  \nonumber\\
 & \propto  \sum_{i_t}  {\rm Pr} (i_{t+1}|i_t) \exp \left(-l_{t|t-1}^{i_{t}}\right)\bar{w}_{t}^{i_{t}} \nonumber\\
 &\overset{(a)}{=} \sum_{i_t}  {\rm Pr} (i_{t+1}|i_t) \exp \left(-l_{t|t-1}^{i_{t}}\right) {w}_{t|t-1}^{i_{t}} \nonumber\\
& \overset{(b)}{\propto} \sum_{i_t}  {\rm Pr} (i_{t+1}|i_t)  w_{t|t}^{i_t} \overset{(c)}{=} w_{t+1|t}^{i_{t+1}} \label{eq: w_eq}
\end{align}
where $(a)$ follows due to the induction assumption that $\bar{w}_{t}^{i_{t}} =w_{t|t-1}^{i_{t}}$; and, $(b)$ and $(c)$ are based on \eqref{eq:SIEGP_w} and \eqref{eq:w_pre_s}, respectively.
	
With $\bar{\ell}_{t+1|t}={\ell}_{t+1|t}^{\rm SW}$ being established according to \eqref{eq: w_eq}, multiplying \eqref{eq:w_comp} from $t=1$ to $T$ yields
\begin{align}
		\sum_{\tau=1}^T \ell_{\tau+1|\tau}^{\rm SW}- \sum_{\tau=1}^T l_{\tau+1|\tau}^{i_{\tau+1}} &=\log \bar{w}_{T} (\mathbf{i}_T) -\log \bar{w}_{0} (\mathbf{i}_T) \nonumber\\
		&\leq -\log \bar{w}_{0} (\mathbf{i}_T)\;.  \label{eq:id_1}
	\end{align}
Since $\bar{w}_{0} (\mathbf{i}_T)\!  =\!  {\rm Pr} (i_1) \prod_{\tau=2}^{T} {\rm Pr}(i_\tau|i_{\tau-1})\! =\! \frac{1}{M} q_0^{ T-s} q_1^{ s} $ with $s$ denoting the number of switches in $\mathbf{i}_T$, \eqref{eq:id_1} can be rewritten as
\begin{align}
\hspace*{-0.2cm}	\sum_{\tau=1}^T \ell_{\tau+1|\tau}^{\rm SW}\!\!-\!\! \sum_{\tau=1}^T l_{\tau+1|\tau}^{i_{\tau+1}} &\leq \log\! M\!-\! (T\!-\! s) \log\! q_0\! -\! s \log q_1 \nonumber\\
		&\overset{(b)}{\leq} \log\! M\! -\! T \log\! q_0 \!+\! S\! \log \frac{q_0}{1-q_0} \label{eq:id_2}
\end{align}
where the inequality in $(a)$ results from $s\leq S$ and $q_0\geq q_1=1-q_0$ based on (as4)-(as5). As the RHS of \eqref{eq:id_2} is a convex function of $q_0$, the following holds true upon setting it to its minimal value taken at $q_0^* = (T-S)/T$
	\begin{align}
		\sum_{\tau=1}^T \ell_{\tau+1|\tau}^{\rm SW}\!-\! \sum_{\tau=1}^T l_{\tau+1|\tau}^{i_{\tau+1}} &\leq \log M \!-\! S \log \frac{S}{T}\!+\! (T\!-\!S) \log  \frac{T}{T-S} \nonumber 
	\end{align}
which, upon leveraging $(T-S) \log \frac{T}{T-S} \leq (T-S) \frac{S}{T-S}  =S$ for $S\ll T$, yields Lemma 2.\\

\subsection{Proof of Lemma 3}\label{sec:proof_Lemma3}
 For any sequence $\mathbf{i}_T$, the cumulative loss over $T$ slots measured by negative log-likelihood for fixed parameter set $\bbTheta:=\{\bbTheta^m\}_{m=1}^M$ is given by
$\mathcal{L}_\Theta^{\mathbf{i}_T} := - \log p(\mathbf{y}_T | \bbTheta, \mathbf{i}_T, \mathbf{X}_T)  = \sum_{\tau = 1}^T \mathcal{L} (\bbphi^{i_\tau\top}_{\mathbf{v}}(\mathbf{x}_\tau)\bbtheta^{i_\tau}; y_\tau)$,
whose expected value over factorized pdf $q(\bbTheta) = \prod_{m=1}^{M}q(\bbtheta^m) =\prod_{m=1}^{M} \mathcal{N}(\bbtheta^m; \bbtheta^m_*, \xi_m^2\mathbf{I}_{2D},)$ is
\begin{align}
\bar{\mathcal{L}}_{q_\Theta}^{\mathbf{i}_T}  &:= \!\!\int \!\!	\mathcal{L}_\Theta^{\mathbf{i}_T}  q(\bbTheta)d\bbTheta \!=\! \sum_{\tau = 1}^T \!\int\! \!\mathcal{L} (\bbphi^{i_\tau\top}_{\mathbf{v}}(\mathbf{x}_\tau)\bbtheta^{i_\tau}; y_\tau) q(\bbtheta^{i_\tau}) d\bbtheta^{i_\tau} \nonumber
\end{align}
which can be re-expressed as $\bar{\mathcal{L}}_{q_\Theta}^{\mathbf{i}_T}  = \sum_{m=1}^M \bar{\mathcal{L}}_{q_\Theta}^{m}$, where 
\begin{align}
\bar{\mathcal{L}}_{q_\Theta}^{m}  := \sum_{\tau\in\mathcal{T}_m} \int
\mathcal{L} (\bbphi^{m\top}_{\mathbf{v}}(\mathbf{x}_\tau)\bbtheta^{m}; y_\tau) q(\bbtheta^m) d\bbtheta^m     \nonumber
\end{align}	
with $\mathcal{T}_m$ collecting the $T_m$ slot indices when GP model from expert $m$ is in action, that is, $\mathcal{T}_m:=\{\tau|i_\tau\!=\!m, 1\!\leq\!\tau\!\leq\! T\}$.

On the other hand, the cumulative online loss for any $\mathbf{i}_T$ is 
\begin{align}
\sum_{\tau = 1}^T l^{i_\tau}_{\tau|\tau-1}\!\!=\!\sum_{\tau = 1}^{T}\! -\!\log p(y_\tau| \mathbf{y}_{\tau-1},i_\tau,\! \mathbf{X}_\tau)\!  =\! -\log p(\mathbf{y}_T|\mathbf{i}_T,\!\mathbf{X}_T). \nonumber
\end{align}
With $p(\bbTheta) \!=\! \prod_{m=1}^{M}p(\bbtheta^m)\!=\!\prod_{m=1}^{M} \mathcal{N}(\bbtheta^m; \mathbf{0}_{2\nrf}, \sigma_{\theta^m}^2\mathbf{I}_{2\nrf})$, the following inequality can be proved in accordance with \eqref{eq:KL(q||p_0)}
\begin{align}
&\sum_{\tau = 1}^{T} l_{\tau|\tau-1}^{i_\tau}- \bar{\mathcal{L}}_{q_\Theta}^{\mathbf{i}_T}  = \int q(\bbTheta) \log \frac{p(\mathbf{y}_T |\bbTheta,\mathbf{i}_T, \mathbf{X}_T)}{p(\mathbf{y}_T|\mathbf{i}_T, \mathbf{X}_T)} d\bbTheta \nonumber\\
&	= \int q(\bbTheta) \log \frac{p(\bbTheta| \mathbf{y}_T,\mathbf{i}_T,\mathbf{X}_T)}{p(\bbTheta)} d\bbTheta \nonumber\\ 
&  = \sum_{m=1}^M \int q(\bbtheta^m) \left(\log \frac{q(\bbtheta^m)}{p(\bbtheta^m)}-\log \frac{q(\bbtheta^m)}{p(\bbtheta^m| \mathbf{y}^m,\mathbf{X}^m)}\right) d\bbtheta^m\nonumber\\
&\leq \sum_{m=1}^M  KL(q(\bbtheta^m)\|p(\bbtheta^m)) \nonumber\\
&= \sum_{m=1}^M \!\!\left(\! 2\nrf\! \log\frac{\sigma_{\theta^m}}{\xi_m} \!+\! \frac{1}{2\sigma_{\theta}^2} \left(\|\bbtheta_{*}^m\|^2\! +\! 2\nrf \xi_m^2 \right)\!-\! \nrf\!\right). \label{eq:74}
\end{align}
Next, following the derivations in \eqref{eq:L(z,y)}--\eqref{eq: 56} yields for any $m$
\begin{align}
\bar{\mathcal{L}}_{q_\Theta}^{m} \leq \sum_{t\in \mathcal{T}_m}  \mathcal{L} (\bbphi^{m\top}_{\mathbf{v}}(\mathbf{x}_t)\bbtheta^m_{*}; y_t)  +\frac{T_m c \xi_m^2}{2}	\nonumber
\end{align}
the sum of which over $m$ in conjunction with \eqref{eq:74} results in
\begin{align}
&\sum_{\tau = 1}^{T} l_{\tau|\tau-1}^{i_\tau}-\sum_{\tau=1}^T  \mathcal{L} (\bbphi^{i_\tau\top}_{\mathbf{v}}(\mathbf{x}_\tau)\bbtheta^{i_\tau}_{*}; y_\tau) \leq \label{eq:75}\\
& \sum_{m=1}^M \!\!\!\left(\! 2\nrf\! \log\frac{\sigma_{\theta^m}}{\xi_m} \!+\! \frac{1}{2\sigma_{\theta}^2} \left(\|\bbtheta_{*}^m\|^2\! +\! 2\nrf \xi_m^2 \right)\!-\! \nrf+\!\frac{T_m c \xi_m^2}{2}\!\right) . \nonumber
\end{align}
It is evident that the RHS of \eqref{eq:75} is a sum  of $M$ convex functions of $\xi_m$, each of which takes minimal value at $\xi_m^* = \sqrt{(2\nrf \sigma_{\theta^m}^2)/(2\nrf + T_mc\sigma_{\theta^m}^2)}$. Setting the RHS of \eqref{eq:75}  to its minimal value yields
	\begin{align}
	&\sum_{\tau = 1}^T l^{i_\tau}_{\tau|\tau-1} - \sum_{\tau = 1}^T \mathcal{L} (\bbphi^{i_\tau\!\top}_{\mathbf{v}}(\mathbf{x}_\tau)\bbtheta^{i_\tau}_{*}; y_\tau) \nonumber\\
	&	\leq  \sum_{m=1}^M\left( \frac{\|\bbtheta^{m}_*\|^2}{2\sigma_{\theta^m}^2} + \nrf\log\left(1+ \frac{T_m c\sigma_{\theta^{m}}^2}{2\nrf} \right)\right)\nonumber\\
	& \leq \sum_{m=1}^M\left( \frac{\|\bbtheta^{m}_*\|^2}{2\sigma_{\theta^m}^2} + \nrf\log\left(1+ \frac{T_m c\sigma_{\theta^{*}}^2}{2\nrf} \right)\right) \label{eq:Lemma3_last}
	\end{align}
	where $\sigma_{\theta^{*}}^2 = \max_{m\in\mathcal{M}} \sigma_{\theta^{m}}^2$. Since $\log (\cdot)$ is a concave function, it follows with $T=\sum_{m=1}^M T_m$ that
	\begin{align}
	\sum_{m=1}^M \log\left(1+ \frac{T_m c\sigma_{\theta^{*}}^2}{2\nrf} \right)\leq M \log\left(1+ \frac{Tc\sigma_{\theta^{*}}^2}{2\nrf} \right) \nonumber
	\end{align}
which, upon plugging into \eqref{eq:Lemma3_last}, finalizes the proof of Lemma~3.

\subsection{Proof of Theorem 2.} \label{sec:proof_Thm2}
For a given $\mathbf{i}_T$, expert $m$ relies on batch data $\{\mathbf{x}_\tau, y_\tau, \tau\in\mathcal{T}_m\}$ in hindsight to learn the benchmark function $\hat{f}^m (\mathbf{x})$ in the RKHS and the RF-based one $\check{f}_{*}^m (\mathbf{x})$ with parameter vector $\bbtheta^m_*$.
Following \eqref{eq:kernel_approx}--\eqref{eq:lambda_RF_RKHS_2} in Sec.~\ref{sec:proof_Thm1}, it holds true
with probability at least $1-2^8  (\frac{\sigma_m}{\epsilon})^2 \exp\left(\frac{-\nrf\epsilon^2}{4d+8} \right)$ that 	
\begin{align}
\sum_{\tau  \in \mathcal{T}_m}\!\!\!\mathcal{L} \!\left(\bbphi^{m\top}_{\mathbf{v}}(\mathbf{x}_\tau){\bbtheta^m_*} ; y_\tau\!\right) \!-\!\!\!\! \sum_{\tau \in \mathcal{T}_m}\!\!\! \mathcal{L} \!\left(\hat f^{m} (\mathbf{x}_\tau); y_\tau \!\right)  \!\leq\! \epsilon LT_m C' \nonumber
\end{align}
where $C':=\underset{m \in \mathcal{M}}{\max} \sum_{\tau\in \mathcal{T}_m } \sigma_{\theta^m}^2 | \hat\alpha_{\tau}^{m}|$. Summing the above inequality over $m\in \mathcal{M}$ leads to 
\begin{align}
  \sum_{\tau=1}^T\!\mathcal{L} \!\left(\bbphi^{m\top}_{\mathbf{v}}(\mathbf{x}_\tau){\bbtheta^m_*} ; y_\tau\!\right) \!-\!\!\! \sum_{\tau=1}^T\! \mathcal{L} \!\left(\hat f^{m} (\mathbf{x}_\tau); y_\tau \!\right)  \!\leq\! \epsilon LT C'\;\label{eq:lambda_RF_RKHS_new}  
\end{align}
holding true with probability at least $1-2^8 (\frac{\sigma_{*}}{\epsilon})^2 \exp\left(\frac{-\nrf\epsilon^2}{4d+8} \right)$, where $\sigma_{*}^2 : =\max_{m\in\mathcal{M}} \sigma_{m}^2$.

Meanwhile, adapting the result in \eqref{eq:theta_bound} to expert $m$ possessing data $\mathcal{D}_m$, yields the ensuing inequality concerning $\bbtheta^{m}_{*}$
\begin{align}
	&\|\bbtheta^{m}_{*} \|^2:=  \left\|\sigma_{\theta^m}^2\sum_{\tau  \in \mathcal{T}_m} \hat\alpha^{m}_{\tau} \bbphi^m_{\mathbf{v}} (\mathbf{x}_\tau)  \right\|^2 \label{eq:theta_bound_new}\\
	&=  \sigma_{\theta^m}^4\sum_{\tau \in \mathcal{T}_m}\sum_{\tau'\in \mathcal{T}_m} \hat\alpha^{m}_{\tau} \hat\alpha^{m}_{\tau'}\bbphi^{m\top}_{\mathbf{v}} (\mathbf{x}_\tau)\bbphi^m_{\mathbf{v}} (\mathbf{x}_{\tau'})  \leq (1+\epsilon){C'}^2   \nonumber
\end{align}
which, in conjunction with Lemmas 2-3 and \eqref{eq:lambda_RF_RKHS_new}, readily proves Theorem~2.

\ifCLASSOPTIONcompsoc
  \section*{Acknowledgments}
\else
  \section*{Acknowledgment}
\fi

The authors would like to thank the anonymous reviewers for their constructive feedback. We also gratefully acknowledge the support from NSF grant 1901134.

\ifCLASSOPTIONcaptionsoff
  \newpage
\fi



%

\bibliographystyle{IEEEtran}
\bibliography{SSGP}

\appendix
\begin{figure}
	\centering
	\includegraphics[width=0.75\linewidth]{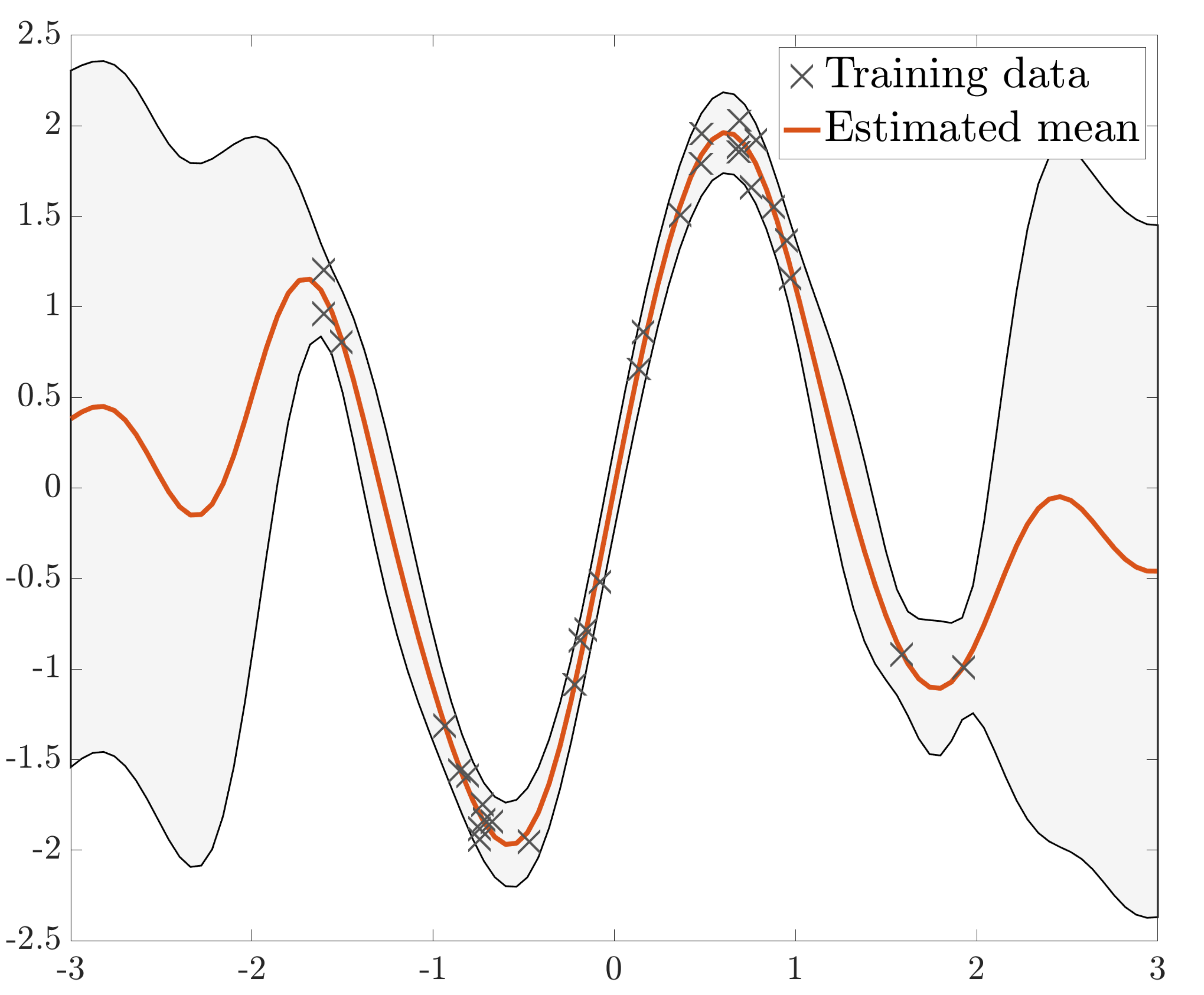}
	\caption{IE-GP inference on a synthetic dataset. Shaded regions indicate $95\%$ confidence intervals.}
	\label{fig:synth}
\end{figure}
\begin{figure}
	\centering
	\includegraphics[width=0.75\linewidth]{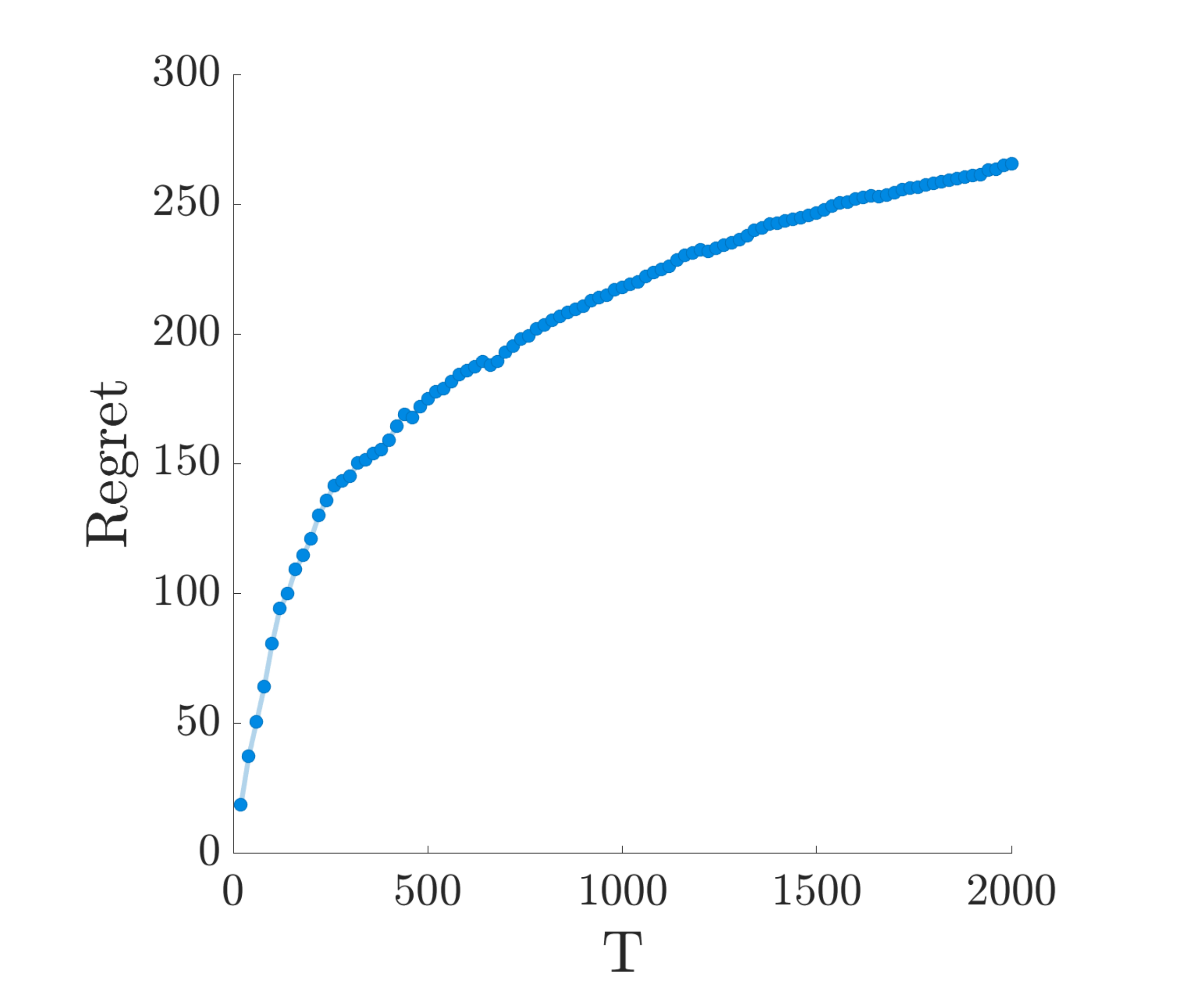}
		\caption{IE-GP cumulative static regret \eqref{eq:reg_full} on a synthetic dataset.}
		\label{fig:reg}
\end{figure}

\begin{figure}
    \centering
    \includegraphics[width=0.75\linewidth]{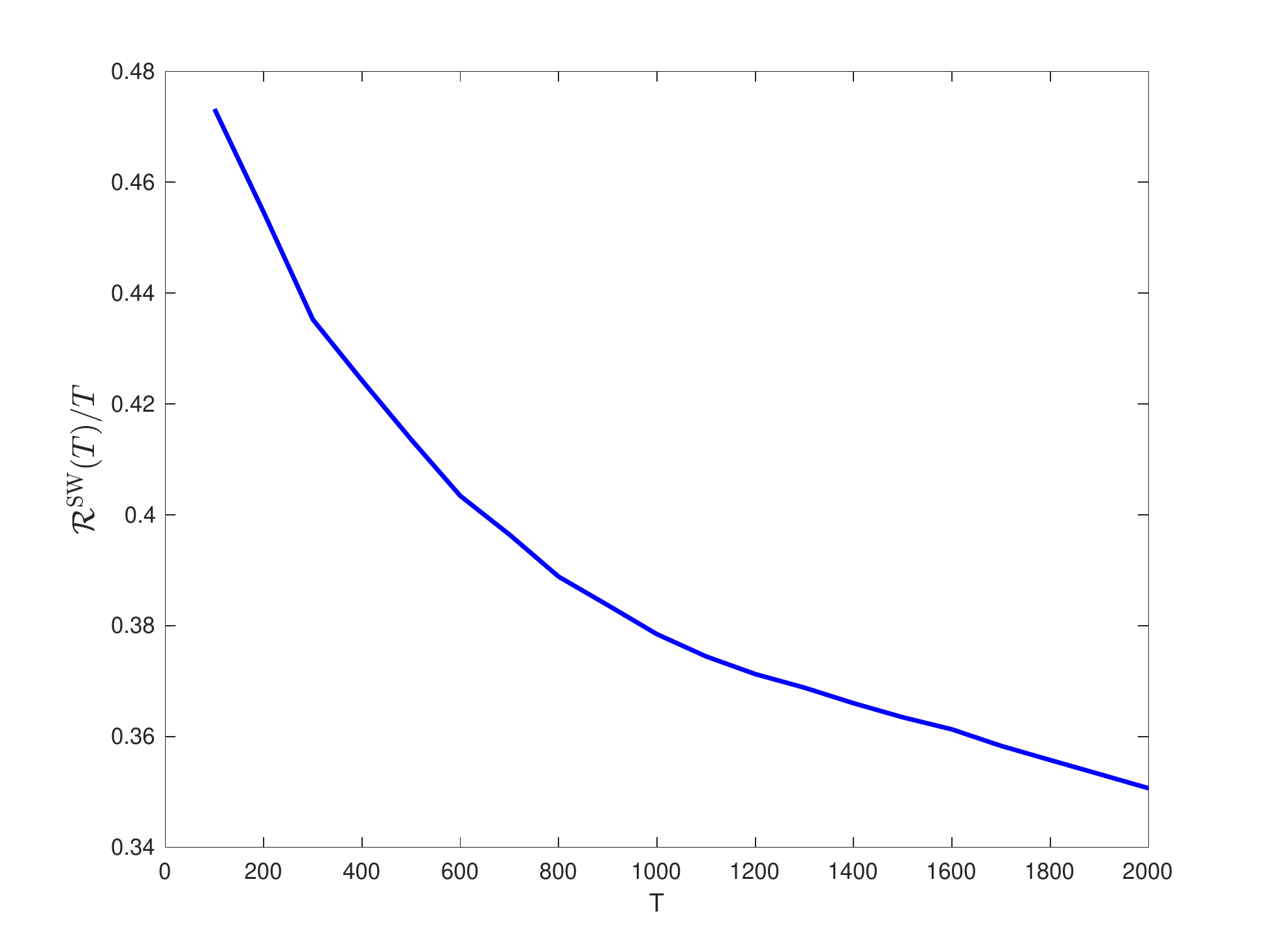}
    \caption{SIE-GP average switching regret, $\frac{\mathcal{R}^{\rm SW}(T)}{T}$, versus $T$.}
    \label{fig:SIEGP_Regret}
\end{figure}
\section{Synthetic tests} \label{app:synth_tests}
This section contains tests with synthetic data to validate the expected uncertainty quantification performance of IE-GP and the regret bounds of (S)IE-GP. To start with, scalar input data $\{x_t\}_{t=1}^{30}$ were randomly drawn from a standardized normal distribution, and outputs were generated as $y_t = \sin(2x_t)+\sin(3x_t)+\epsilon_t$, where  $\epsilon_t \sim \mathcal{N}(0,0.01)$. The inferred mean function as well as (approximate) $95 \%$ confidence intervals are shown in Fig. \ref{fig:synth}. As expected, regions populated with training examples correspond to tighter confidence bands relative to unpopulated ones. 

To validate the regret bound for IE-GP (cf. \eqref{eq:reg_full}), datasets of increasing size $T$ were generated from the aforementioned synthetic model, albeit with $x_t \sim \mathcal{N}(0,100)$. It is evident from Fig. \ref{fig:reg} that the regret can be upper bounded by $\mathcal{O}(\log T)$, as predicted by the Theorem 1. To further corroborate the regret bound of SIE-GP, synthetic switching datasets with growing $T$ were generated, where the first $T/2$ noiseless outputs were drawn from $\mathcal{GP}(0,\kappa_1)$ and the second half from $\mathcal{GP}(0,\kappa_2)$. Here, $\kappa_1$ and $\kappa_2$ were two RBF kernels whose magnitudes are $1$ and characteristic lengthscales are $0.01$ and $100$, respectively. The observed outputs were obtained by adding noises sampled from zero-mean Gaussian distribution with variance $1$. After running SIE-GP with these two GP models, the average switching regret \eqref{eq:regret_tr} was plotted as a function of $T$ in Fig.~\ref{fig:SIEGP_Regret}, where the convergence behavior of the curve agrees well with Theorem~2.
\end{document}